\documentclass[]{acmart}

\makeatletter
\newif\ifanonymousmode
\newif\ifnotanonymousmode
\if@ACM@anonymous
  \anonymousmodetrue
  \notanonymousmodefalse
\else
  \anonymousmodefalse
  \notanonymousmodetrue
\fi
\makeatother

\usepackage{color}
\usepackage{etoolbox}
\usepackage{xfp}
\usepackage{graphicx}
\usepackage{subcaption}
\usepackage{enumitem}
\usepackage{multirow}
\usepackage[utf8]{inputenc} 
\usepackage[T1]{fontenc}    
\usepackage{hyperref}       
\usepackage{url}            
\usepackage{booktabs}       
\usepackage{amsfonts}       
\usepackage{nicefrac}       
\usepackage{microtype}      
\usepackage{xcolor} 
\usepackage{adjustbox}
\usepackage{datatool}
\usepackage{algorithm}
\usepackage{algpseudocode}
\usepackage{siunitx}
\usepackage{pgffor} 
\usepackage{arydshln} 
\usepackage{comment}
\usepackage{tikz}
\usetikzlibrary{shapes, positioning, arrows.meta, decorations.pathmorphing, backgrounds, shadows} 
\usetikzlibrary{shadows}

\usepackage{fontawesome}
\usepackage[skip=0.5em]{caption}
\usepackage{xcolor}
\definecolor{Goldenrod}{rgb}{0.85, 0.65, 0.13}
\definecolor{Green}{rgb}{0.0, 0.8, 0.0}  
\usepackage[utf8]{inputenc}
\DeclareSIUnit\year{yr}

\newtoggle{cameraReady}

\ifnotanonymousmode
  \toggletrue{cameraReady}  
\fi

\newtoggle{CommentsMode}
\iftoggle{cameraReady}{\togglefalse{CommentsMode}}{\toggletrue{CommentsMode}}

\usepackage{ulem}
\newcommand{\rmText}[1]{\iftoggle{cameraReady}{\relax}{\textcolor{blue}{\sout{#1}}}}
\newcommand{\newSubsection}[1]{\iftoggle{cameraReady}{\subsection{#1}}{\subsection{\textcolor{blue}{#1}}}}  

\newcommand{\newText}[1]{{\iftoggle{cameraReady}{#1}{\textcolor{blue}{#1}}}}

\DTLloaddb{datasetCounts}{data/dataset_counts.csv}

\newcommand{\bestModel}{yolo11m-obb}

\newcommand{\DTLfetchsave}[5]{%
  \edtlgetrowforvalue{#2}{\dtlcolumnindex{#2}{#3}}{#4}%
  \dtlgetentryfromcurrentrow{\dtlcurrentvalue}{\dtlcolumnindex{#2}{#5}}%
  \let#1\dtlcurrentvalue
}
\newcommand{\UPSurveyCounts}{19671}
\newcommand{\UPMZeroTP}{15627}
\newcommand{\UPMZeroPP}{26978}
\DTLdiv{\UPMZeroPrecisionF}{\UPMZeroTP}{\UPMZeroPP}
\DTLmul{\UPMZeroPrecisionF}{\UPMZeroPrecisionF}{100}
\newcommand{\UPMZeroPrecision}{\num[round-mode=places, round-precision=0]{\UPMZeroPrecisionF}}

\newcommand{\UPFinalP}{82}
\newcommand{\BiharFinalP}{76}
\newcommand{\WBFinalP}{66}
\newcommand{\HaryanaFinalP}{72}
\newcommand{\PunjabFinalP}{71}

\DTLfetchsave\UPOurCounts{datasetCounts}{State}{Uttar Pradesh}{Total}
\DTLconverttodecimal{\UPOurCounts}{\NumUPOurCounts}

\DTLmul{\UPOurCountsMul}{\UPOurCounts}{100}
\DTLdiv{\UPDetectionPercentFull}{\UPOurCountsMul}{\UPSurveyCounts}

\AtBeginDocument{%
  }

\setcopyright{acmlicensed}
\copyrightyear{2018}
\acmYear{2018}
\acmDOI{XXXXXXX.XXXXXXX}

\acmConference[Conference acronym 'XX]{Make sure to enter the correct
  conference title from your rights confirmation email}{June 03--05,
  2018}{Woodstock, NY}
\acmISBN{978-1-4503-XXXX-X/18/06}

\begin{document}
\begin{filecontents*}{data/dataset_counts.csv}
State,CFCBK,FCBK,Zigzag,Total
Uttar Pradesh,1450,9933,5952,17335
Bihar,40,1584,4424,6048
West Bengal,33,1105,1967,3105
Haryana,1,130,1948,2079
Punjab,0,305,1766,2071
Total,1524,13057,16057,30638
\end{filecontents*}

\title[Space to Policy]{Space to Policy: Scalable Brick Kiln Detection and Automatic Compliance Monitoring with Geospatial Data}

\author{Zeel B Patel}
\email{patel_zeel@iitgn.ac.in}
\orcid{0000-0002-1981-3912}
\author{Rishabh Mondal}
\email{rishabh.mondal@iitgn.ac.in}
\author{Shataxi Dubey}
\email{shataxi.dubey@iitgn.ac.in}
\author{Suraj Jaiswal}
\email{jaiswalsuraj@iitgn.ac.in}
\affiliation{%
  \institution{IIT Gandhinagar}
  \city{Gandhinagar}
  \state{Gujarat}
  \country{India}
}
\author{Sarath Guttikunda}
\email{simair@urbanemissions.info}
\affiliation{%
  \institution{UrbanEmissions.Info}
  \country{India}
}
\author{Nipun Batra}
\email{nipun.batra@iitgn.ac.in}
\affiliation{%
  \institution{IIT Gandhinagar}
  \city{Gandhinagar}
  \state{Gujarat}
  \country{India}
}








\renewcommand{\shortauthors}{Trovato et al.}

\begin{abstract}
Air pollution kills 7 million people annually. The brick kiln sector significantly contributes to economic development but also accounts for 8-14\% of air pollution in India. Policymakers have implemented compliance measures to regulate brick kilns. Emission inventories are critical for air quality modeling and source apportionment studies. However, the largely unorganized nature of the brick kiln sector necessitates labor-intensive survey efforts for monitoring. Recent efforts by air quality researchers have relied on manual annotation of brick kilns using satellite imagery to build emission inventories, but this approach lacks scalability. Machine-learning-based object detection methods have shown promise for detecting brick kilns; however, previous studies often rely on costly high-resolution imagery and fail to integrate with governmental policies. In this work, we developed a scalable machine-learning pipeline that detected and classified 30638 brick kilns across five states in the Indo-Gangetic Plain using free, moderate-resolution satellite imagery from Planet Labs. Our detections have a high correlation with on-ground surveys. We performed automated compliance analysis based on government policies. In the Delhi airshed, stricter policy enforcement has led to the adoption of efficient brick kiln technologies. This study highlights the need for inclusive policies that balance environmental sustainability with the livelihoods of workers.
\end{abstract}





\maketitle

\section*{Scope of Applicability}
This study is conducted for academic and research purposes to contribute to scientific understanding and inform policy development on air quality and sustainability issues. While the findings highlight opportunities for improving compliance and technology adoption in the brick kiln sector, they are not intended for direct enforcement or punitive actions against individual entities or groups.
We acknowledge the socio-economic importance of brick kilns and recommend that any policy interventions based on this work be developed in consultation with all stakeholders, including kiln operators, workers, policymakers, and researchers. 

\section{Introduction}

Air pollution causes seven million deaths annually
with 22\% occurring in India~\cite{unep19}. In 2019, 99\% of the world’s population lived in areas that did not meet WHO air quality guideline levels~\cite{who_air_pollution}.
In India, brick kilns contribute 8–14\% of air pollution~\cite{worldbank2020} and employ 15 million workers, including children~\cite{rajarathnam2014, boyd2018slavery}, directly impacting UN Sustainable Development Goal 8.7~\cite{boyd2018slavery}. 

Initially, many brick kilns operated with Fixed Chimney Bull's Trench Kiln (FCBK), an older technology. Studies show that converting to Zigzag technology can reduce CO and particulate matter emissions by 70\%~\cite{tibrewal2023reconciliation}. FCBKs and Zigzag kilns account for 75\% of the brick production in India~\cite{rajarathnam2014}. 
The National Clean Air Program (NCAP), launched in 2019, introduced guidelines to mitigate air pollution and identified 130 cities as ``non-attainment cities'' due to non-compliance with air quality guidelines. Indian central government and state governments have established compliance rules for brick kilns to regulate their operations~\cite{BrickKilnsRules2022}. 
Source apportionment studies and air quality modeling rely on accurate emission inventory~\cite{guttikunda2019air}. Developing emission inventories and monitoring compliance require accurate geo-mapping of brick kilns and their technologies. However, brick kilns are small-scale, unorganized units, making geo-mapping through field surveys resource-intensive.

Air quality researchers often use tools like `Google Earth' to manually annotate brick kilns in satellite imagery for emission inventory development. Manually annotating an area of size 3600 km$^2$ takes around 6-8 hours. Scaling this for a country like India would require more than 7000 hours of manual effort. Periodically updating the inventory would again require a similar effort.
Recent studies~\cite{redmon2016you,szegedy2017inception,boyd2018slavery,nazir2020kiln,lee2021scalable,haque2022impact,imaduddin2023detection} have demonstrated the potential of object detection methods to scale brick kiln detection from satellite imagery. In our work, we employ state-of-the-art object detection models to detect brick kilns and identify their technologies. To the best of our knowledge, ours is the first large-scale study that accurately detects brick kilns using freely available moderate-resolution imagery. Ours is also the first study that combines brick kiln detection with compliance monitoring in close alignment with government policies and systematically analyzes the growth of brick kilns over time.

In this study, we utilize quarterly satellite imagery with a resolution of
4.77 meters per pixel from Planet Labs, which is freely available under a `Personal Research License' through their Education and Research Program~\cite{planetEducationResearch}. For creating the initial training data for object detection models, we annotated brick kilns in four strategically selected regions. The selection criteria included high pollution levels, population density, recommendations from air quality experts, and designation as non-attainment cities.
These regions were deliberately chosen due to their significant environmental and health concerns, ensuring a representative dataset.
The annotation process covered an area of 15,018 km$^2$, requiring approximately 125 hours of effort. We trained variants of the YOLO model~\cite{ultralytics_yolo_2023} to detect brick kilns and classify their technologies into three categories: i) Circular Fixed Chimney Bull's Trench Kiln (CFCBK); ii) Fixed Chimney Bull's Trench Kiln (FCBK); and iii) Zigzag kilns. The dataset was divided into training, validation, and test sets for model selection, with \bestModel\ achieving the best performance on the test set. \newText{Additionally, we conducted an `out-of-region' experiment to evaluate the best model's performance on areas far from the training data. This experiment revealed a significant performance drop, underscoring the importance of using a model trained on all four initial regions for further detection.}
The best model was used to predict kilns in Uttar Pradesh, the largest state in India by population,
followed by manual validation, yielding \UPMZeroTP\ newly identified kilns.
We further fine-tuned the model by incorporating these newly validated kilns into the training data. The updated model was then applied across five Indo-Gangetic states: Uttar Pradesh, Bihar, West Bengal, Haryana, and Punjab, covering an area of 520,000 km$^2$ with a population of 448 million.
After manual validation, we obtained \DTLfetch{datasetCounts}{State}{Total}{Total} geo-located kilns, categorized by technology and annotated with bounding boxes. Our analysis revealed that CFCBKs, the oldest technology, are predominantly found in Uttar Pradesh. FCBKs are distributed across all five states, while Zigzag kilns are concentrated in Bihar and areas near Delhi. We validated our results against a \newText{state-wide} survey conducted by the Uttar Pradesh Pollution Control Board (UPPCB)~\cite{UPPCB2023}, achieving a strong correlation of 0.94. \newText{We also validated our results against surveys conducted in Delhi-NCR, Bihar, and West-bengal, suggesting strong alignment with our detections.} 

Using the geo-located brick kiln dataset, we applied various big data resources for automated compliance monitoring, focusing on two types of compliance rules: distance-based and technology-based. \textbf{Our analysis of distance-based compliance, using OpenStreetMap data~\cite{geofabrik2024} across 13 categories, revealed that 70\% of brick kilns violate at least one rule}. For technology-based compliance, we examined trends in brick kiln technologies over 12 years in two non-attainment city airsheds, Delhi (the capital city of India) and Lucknow (the capital city of Uttar Pradesh)
using Esri's historical high-resolution imagery~\cite{esriWaybackServer}. Between 2017 and 2021, brick kilns around Delhi significantly shifted toward advanced technologies, reflecting stricter policy enforcement. In contrast, no such shift was observed in Lucknow, indicating weaker implementation of regulations.
We further estimated daily emissions for each state using experimentally established emission rates~\cite{rajarathnam2014}. These emissions, combined with other sources, were incorporated into an emission inventory to perform a source apportionment study for Delhi March and April 2024 using WRF-CAMx~\cite{camx}. \textbf{Our findings indicate that brick kilns contribute approximately 8\% to PM$_{2.5}$ concentrations in Delhi. Additionally, using LandScan population data~\cite{lebakula2024landscan}, we determined that 30.66 million people live within 800 meters of brick kilns, violating central government regulations~\cite{BrickKilnsRules2022}.}

Our work demonstrates a scalable approach to accelerating policy validation processes by leveraging advancements in machine learning, object detection, satellite imagery, and publicly available big data. Notably, aside from computational resources for model training, all components of our study are cost-free and free from proprietary licenses. This openness enables researchers and policymakers to build upon our framework or conduct similar studies without financial constraints.

This article is structured as follows: Section~\ref{sec:background} provides an overview of brick kilns, their impact on air pollution, alignment with UN SDGs, relevant Indian government policies, and object detection models. Section~\ref{sec:related_work} reviews related studies on object detection from aerial imagery, the application of computer vision for UN SDGs, and satellite-based brick kiln detection. In Section~\ref{sec:big_data}, we outline the big data resources used for automated compliance monitoring and population exposure estimation. Section~\ref{sec:brick_kiln_dataset} details the process of creating a geo-located brick kiln dataset using object detection on satellite imagery. Section~\ref{sec:compliance} presents results from automated compliance monitoring. Section~\ref{sec:air_pollution_and_HE} explores the experiments linking air pollution and health effects associated with brick kilns. Section~\ref{sec:discussion} discusses \rmText{potential policy improvements} socio-economic, data and policy implications, and finally, we discuss our study's limitations, propose future work, and provide a conclusion.

\ifanonymousmode
  \noindent\textbf{Reproducibility Statement:} our work is fully reproducible and we will release all the code and resource links to reproduce the results in this work \rmText{upon acceptance} \newText{on camera ready}.
\else
  \noindent\textbf{Reproducibility Statement:} Code and resources to reproduce our work can be found at our project page\footnote{\url{https://sustainability-lab.github.io/brick-kilns/}}
\fi



\section{Background}\label{sec:background}
In this section, we provide the background on brick kiln operations and technologies and their association with air pollution, sustainability, and Indian government policies.
\subsection{Brick Kilns}
A brick kiln is a facility where bricks are produced through shaping, drying, firing, and cooling processes. 
Brick-making requires a sufficient supply of soil, sand, water, and fuel. 
Soil used for brick-making requires moderate clay content, typically found in deposits along riverbanks, leading to the establishment of many kilns near rivers. The brick-making process involves four main steps~\cite{ILO1984}: 
\begin{enumerate}[label=\roman*)]
    \item \textbf{Shaping:} Bricks are molded into the desired shapes, either manually or using machines.
    \item \textbf{Drying:} The bricks are air-dried to allow shrinkage caused by water loss and to strengthen them before stacking for firing.
    \item \textbf{Firing:} Bricks are heated in kilns to achieve high mechanical strength and water resistance. This step releases polluting gases and particulates into the air through chimneys.
    \item \textbf{Exporting:} The finished bricks are removed in batches and transported from the kiln.
\end{enumerate}


\subsection{Continuous Fire Brick Kilns}
Brick kilns are primarily classified by their firing method into two categories: batch production kilns and continuous fire kilns. Continuous fire kilns are more energy-efficient as they reuse warm air from the combustion zone to dry unfired bricks, and the heat contained in the fired bricks is used to preheat the air.
In India, approximately 75\% of bricks are produced by continuous fire kilns~\cite{weyant2014emissions}, with FCBK and Zigzag kilns accounting for over 75\% of continuous fire kilns~\cite{tibrewal2023reconciliation}.
FCBKs feature a closed circular or oval circuit of bricks stacked in the annular space between the kiln's
outer and inner walls. They operate using natural draught generated by a central chimney but suffer from incomplete combustion, leading to significant pollutant emissions~\cite{ILO1984, Greentech2012}. Zigzag kilns, on the other hand, have a rectangular design with a chimney positioned at the center or side. The bricks are arranged to guide the fire along a zigzag path, enabling more efficient combustion and better fuel efficiency. The zigzag kilns can operate under natural draught or draught created by an induced draught fan. A majority of the zigzag kilns in India use induced draught fans. Figure~\ref{fig:data_samples} illustrates satellite views of Circular Fixed Chimney Bull's Trench Kilns (CFCBK), FCBKs, and Zigzag kilns.

\subsection{Brick Kilns and Sustainability}
\subsubsection{Air Pollution}
In 2017, India’s total brick production was estimated at $233 \pm 15$ billion bricks, consuming $990 \pm 125$ \si{\peta\joule} yr$^{-1}$ of energy~\cite{ILO1984}. The primary fuels for brick production are coal and biomass, which release polluting gases and particulates, including particulate matter (PM), SO$_2$, CO, and CO$_2$~\cite{rajarathnam2014}. Brick kilns contribute an estimated 8-14\% of India’s air pollution~\cite{worldbank2020}. Advanced technologies, such as Zigzag kilns, use improved firing methods, reducing PM and CO emissions by approximately 60–70\%~\cite{rajarathnam2014}. Approximately 45\% of India’s brick production occurs in five states: Uttar Pradesh, Bihar, West Bengal, Haryana, and Punjab, all located in the Indo-Gangetic plain—a region known for its dense population and severe air pollution~\cite{tibrewal2023reconciliation}. For this reason, our study focuses on these five states.

\subsubsection{United Nations Sustainable Development Goals}\label{sec:un_sdg}
Brick production relies heavily on labor for shaping bricks, fueling kilns, and maintaining continuous operations~\cite{ILO1984}. In India, the industry employs approximately 15 million workers, including children~\cite{rajarathnam2014, boyd2018slavery}. This raises concerns related to forced labor, modern slavery, and human trafficking, which are central issues addressed by UN Sustainable Development Goal (SDG) 8.7~\cite{unGoalDepartment}. Developing a comprehensive inventory of brick kiln locations can facilitate targeted surveys to identify potential human rights violations, provide assistance to workers, and establish clear guidelines for kiln owners to promote ethical practices. Furthermore, our research intersects with the following SDGs:
\begin{itemize}
    \item \textbf{SDG 3.4 \& 3.9:} Addressing premature deaths and illnesses from air pollution.
    \item \textbf{SDG 9.4:} Mitigating CO$_2$ emissions from industries.
    \item \textbf{SDG 11.6:} Monitoring PM$_{2.5}$ and PM$_{10}$ emissions in cities.
    \item \textbf{SDG 12.c:} Reducing fossil-fuel subsidies.
    \item \textbf{SDG 13.2:} Tracking greenhouse gas emissions.
\end{itemize}
\begin{figure}[h]
    \centering
    \begin{subfigure}{0.15\textwidth}
        \centering
        \includegraphics[width=\textwidth]{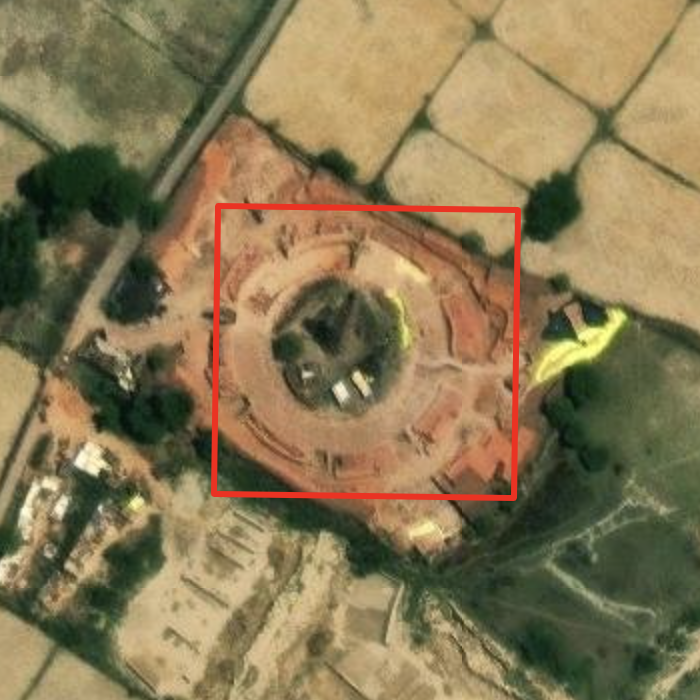}
        \caption{CFCBK - Esri}
    \end{subfigure} \hfill
    \begin{subfigure}{0.15\textwidth}
        \centering
        \includegraphics[width=\textwidth]{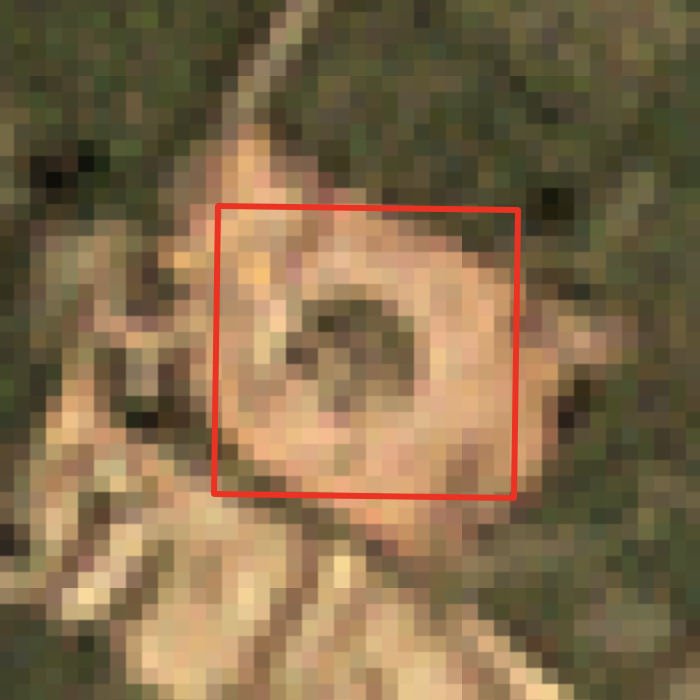}
        \caption{CFCBK - Planet}
    \end{subfigure} \hfill
    \begin{subfigure}{0.15\textwidth}
        \centering
        \includegraphics[width=\textwidth]{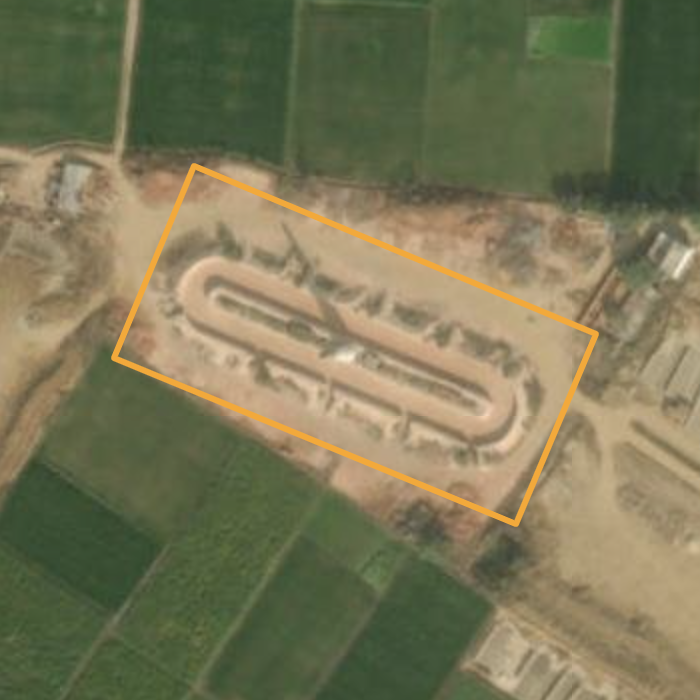}
        \caption{FCBK - Esri}
    \end{subfigure} \hfill
    \begin{subfigure}{0.15\textwidth}
        \centering
        \includegraphics[width=\textwidth]{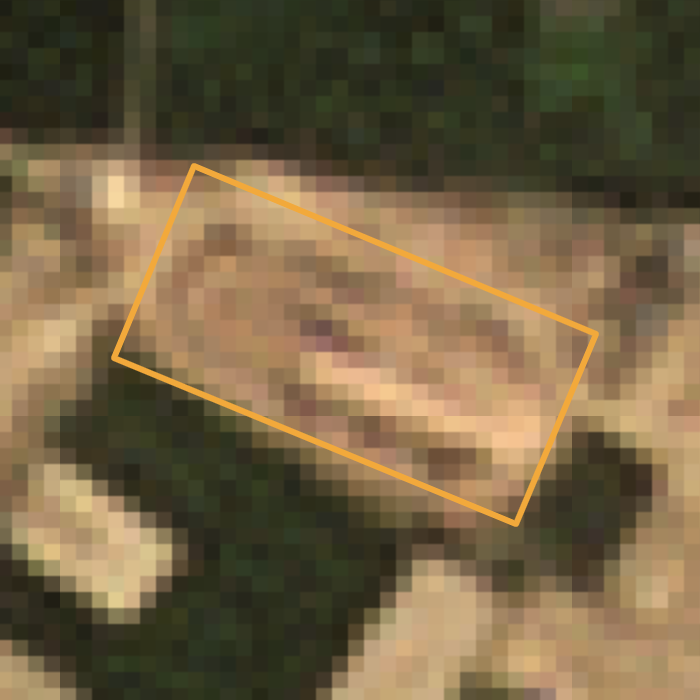}
        \caption{FCBK - Planet}
    \end{subfigure} \hfill
    \begin{subfigure}{0.15\textwidth}
        \centering
        \includegraphics[width=\textwidth]{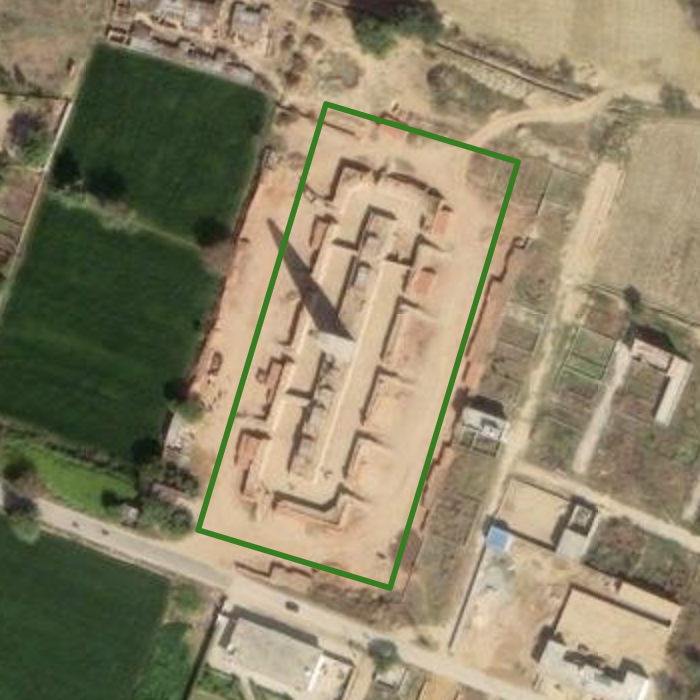}
        \caption{Zigzag - Esri}
    \end{subfigure} \hfill
    \begin{subfigure}{0.15\textwidth}
        \centering
        \includegraphics[width=\textwidth]{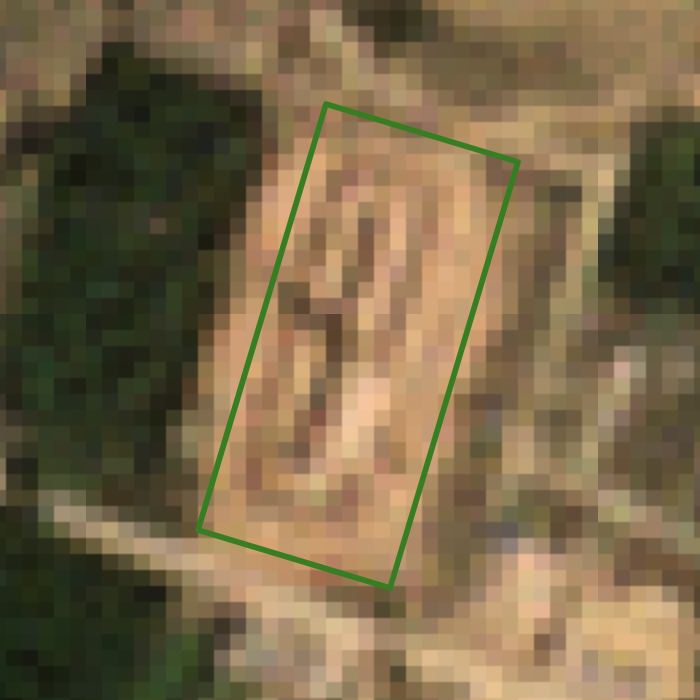}
        \caption{Zigzag - Planet}
    \end{subfigure}
    \caption{
    Satellite view of brick kilns with bounding boxes. CFCBK is Circular Fixed Chimney Bull’s Trench Kiln, and FCBK is Fixed Chimney Bull’s Trench Kiln. We use Esri's high-resolution basemap (zoom level 17-19, 1.19-0.3 m per pixel) to create geo-referenced bounding boxes and hand-validate the predicted bounding boxes. We use Planet's moderate-resolution imagery (zoom level 15, 4.77 m per pixel) to train object detection models. Planet Imagery \textcopyright\ 2024 Planet Labs Inc. Esri imagery \textcopyright\ Esri, TomTom, Garmin, Foursquare, METI/NASA, USGS.}
    \label{fig:data_samples}
\end{figure}

\subsection{Government Policies}
Several government policies outline guidelines for regulating brick kilns. The National Clean Air Program (NCAP)~\cite{cpcbNCAP}, launched in 2019, aims to reduce air pollution levels to comply with National Ambient Air Quality Standards (NAAQS). By 2024, 131 cities were identified as non-attainment cities for failing to meet these standards. NCAP has directed states, the Central Pollution Control Board (CPCB), and State Pollution Control Boards (SPCBs) to expedite the adoption of cleaner technologies, such as Zigzag firing technology, in brick kilns~\cite{cpcbNCAP}. In 2017, brick kilns operating without permits or failing to comply with environmental guidelines were shut down in Delhi and the National Capital Region (NCR)~\cite{cpcbNCAP}. 

The Environment (Protection) Amendment Rules, 2022~\cite{BrickKilnsRules2022}, mandated the conversion of all brick kilns within 10 km of non-attainment cities to cleaner technologies, including Zigzag technology, vertical shaft kilns, or piped natural gas, by 2024. \rmText{The rules also specified:
    Brick kilns must be located at least 800 meters from residential areas and fruit orchards.  
    A minimum distance of 1 km must be maintained between two brick kilns.}

States can set additional siting criteria to regulate brick kilns. However, monitoring compliance with these policies remains a challenge due to the limitations of manual surveys, which are resource-intensive. Automated Geographic Information System (GIS) analysis offers a scalable alternative, enabling near-instantaneous compliance monitoring. By minimizing reliance on time-consuming manual methods, GIS tools can enhance efficiency and support targeted inspections for effective policy enforcement.

\newSubsection{Brick Kiln Siting Rules}
\newText{
Industrial activities can significantly impact the environment through cumulative emissions, raw material extraction, and industrial waste. Therefore, industries must be sited with a balance between economic and environmental considerations. Several India states have established brick kiln siting regulations. One such example is the Brick Kiln Siting Rules, 2012 of Uttar Pradesh~\cite{uttar_pradesh_brick_kiln_rules_2012}.  
Air pollution from brick kilns, particularly the presence of sulfur dioxide (SO$_2$) and hydrogen fluoride (HF), negatively impacts the flowering of fruit trees (e.g., mangoes), leading to reduced fruit yield~\cite{ahmad2012hydrogen}. To mitigate this, the rules specify a minimum distance from fruit orchards.  
To reduce the impact of air pollution on human health, the rules also mandate a minimum distance from residential areas, hospitals, and schools. 
The primary source of brick earth, the raw material for brick-making, comes from mining topsoil from agricultural fields surrounding the kiln, typically up to a 2 meters deep. This practice can alter natural rainwater flow and lead to land erosion. To address this, the kiln siting rules specify minimum distances from railway tracks, state highways, and national highways.  
To prevent the formation of large brick kiln clusters, which can cause a significant cumulative environmental impacts, the Uttar Pradesh kiln siting rules establish a minimum distance between adjacent kilns. The following distance-based criteria have been established by the central government for all states~\cite{BrickKilnsRules2022}:
\begin{enumerate}
    \item Brick kilns must be located at least 800 meters from residential areas and fruit orchards.  
    \item A minimum distance of 1 km must be maintained between two brick kilns.  
\end{enumerate}
}

\subsection{Object Detection Models}
Object detection, a core task in machine learning-based computer vision, involves identifying objects in an image by predicting their bounding boxes and classifying them into predefined categories. The two types of bounding boxes commonly used in object detection are:
\begin{itemize}
\item \textbf{Axis-Aligned Bounding Boxes (AA):} Rectangular boxes with edges parallel to the horizontal and vertical axes.
\item \textbf{Oriented Bounding Boxes (OBB):} Rectangular boxes aligned with the orientation of the objects in the image.
\end{itemize}
OBBs offer a more precise representation of object areas, especially for irregularly oriented objects. In the context of brick kilns, OBBs facilitate accurate spatial extent estimation, which is essential for assessing brick production capacity and calculating emission factors for air quality modeling. Therefore, we employ object detection models supporting OBB annotations to improve the accuracy and applicability of our analysis.

\section{Big Data Resources}\label{sec:big_data}
In this section, we describe the national and global datasets utilized in our work for brick kiln detection and automatic compliance monitoring.
\subsection{Satellite Imagery}
Selecting suitable satellite imagery requires balancing cost and computational efficiency. High-resolution commercial imagery, such as Maxar (30-50 cm) and Airbus products, offer detailed views but is often too expensive for research. Sentinel-2 imagery from the European Space Agency offers free 10-meter resolution with a 5-day revisit cycle, making it a cost-effective alternative.
Brick kilns, on average, cover 150 meters in latitude and longitude and thus span approximately 15×15 pixels in Sentinel-2 imagery. Detecting them at such a low resolution is challenging. Therefore, we utilized freely available moderate-resolution imagery (4.77 meters) under a research license, as detailed in subsequent sections.

\subsubsection{Planet Labs} 
We utilize satellite imagery from Planet Labs in this work~\cite{planetEducationResearch}. Planet offers multiple satellite products, including 3-meter resolution daily imagery, 4.77-meter monthly and quarterly mosaics\footnote{\url{https://desktop.arcgis.com/en/arcmap/latest/manage-data/raster-and-images/what-is-a-mosaic.htm}}.
Imagery is provided through Planet’s Research and Education Program~\cite{planetEducationResearch} and can be freely downloaded via the Mosaics API~\cite{PlanetAP93:online}.
\textbf{Our study area spans five Indian states, covering over 520,000 km² (15.8\% of India) area.} The large $4096\times4096$ sized images are divided into overlapping $640\times640$ sized crops with a 64-pixel overlap ($\approx$300 meters), ensuring that brick kilns do not get cut at the image edges. We selected imagery from the first quarter of 2024 to align with the peak operational season and favorable weather conditions for cloud-free imagery. Although Planet imagery offers better resolution than Sentinel-2, it remains insufficient for identifying kiln technologies. To overcome this, we annotated high-resolution imagery and transferred the labels to Planet imagery, as detailed in the next section.

\subsubsection{Esri Wayback Imagery} 
Esri, a leading provider of GIS technology~\cite{arcgis}, delivers high-resolution basemaps (up to 30 cm) through a Web Map Service (WMS). These timestamped basemaps are typically available for one or two dates per month. We utilized Esri imagery from February 8, 2024, to align with the first-quarter Planet mosaics of 2024. We could not use high-resolution Esri imagery to train the object detection models because the WMS endpoint is designed for on-demand access rather than bulk downloads.


\subsection{OpenStreetMap} OpenStreetMap (OSM)~\cite{openstreetmap}, a collaborative, community-driven initiative, provides a comprehensive repository of geographic datasets contributed by volunteers worldwide. These datasets contain detailed information on infrastructure and natural features, facilitating extensive spatial analysis. Using the Geofabrik platform~\cite{geofabrik2024}, we extracted geometry data for features such as railway tracks, highways, rivers, schools, and habitation. This data is utilized for distance-based compliance monitoring. Table~\ref{tab:osm_data} provides a detailed summary of the extracted features and the filters applied during the extraction process. \newText{The data is obtained by downloading zip files provided by the Geofabrik platform and extracting the data based on specific filters. Researchers can replicate our data collection process using the filters outlined in Table~\ref{tab:osm_data}, ensuring transparency and reproducibility in our methodology.}


\subsection{Government Data}
We obtained geolocation data for 165,000 hospitals and nursing homes from the Indian Government’s data portal~\cite{NINHealthDataGov}. This dataset supports distance-based compliance monitoring of brick kilns, as discussed in Section~\ref{sec:compliance}.


\subsection{Population Data} We used the LandScan global population dataset~\cite{lebakula2024landscan}, produced by Oak Ridge National Laboratory, at a resolution of 0.0083 degrees. \newText{This dataset is updated once every year.} We downloaded the 2023 version of the dataset. We used this dataset to analyze population exposure due to brick kilns' air pollution, as detailed in Section~\ref{sec:population}. \newText{To determine the population within $x$ km of brick kilns, we summed all points falling within $x$ km radius of a brick kiln and ensured each point was counted only once to avoid duplication.} This approach allows for accurate estimation of population exposure while maintaining data integrity.

\begin{table}[h]
    \centering
    \caption{Datasets extracted from OpenStreetMap and filters applied to extract them.}
    \label{tab:osm_data}
    \begin{tabular}{llp{5 cm}}
       \toprule
       Shapefile & Data & Filters \\
       \midrule
       \multirow[m]{3}{*}{\texttt{gis\_osm\_landuse\_a\_free\_1.shp}} & Habitation & \texttt{fclass = ``residential''} \\
    & Orchards & \texttt{fclass = ``orchard''} \\
    & Nature reserves & \texttt{fclass = ``nature\_reserve''} \\
    \hdashline
\multirow[m]{3}{*}{\texttt{gis\_osm\_buildings\_a\_free\_1.shp}} & Schools & \texttt{type = ``school''} \\
& Hospitals & \texttt{type = ``hospital''} \\
& Religious places & \texttt{type = ``temple'' or ``mosque'' or ``church''} \\
\hdashline
\multirow[m]{3}{*}{\texttt{gis\_osm\_roads\_free\_1.shp}} & National \& Express highways & \texttt{ref starts with ``NH'' or ``NE''} \\
& State highways & \texttt{ref starts with ``SH''} \\
& District highways & \texttt{ref starts with ``MDR''} \\
\hdashline
\texttt{gis\_osm\_water\_a\_free\_1.shp} & Wetland & \texttt{fclass = ``wetland''}\\
\hdashline
\texttt{gis\_osm\_waterways\_free\_1.shp} & Rivers & \texttt{fclass = ``river''}\\
\hdashline
\texttt{gis\_osm\_railways\_free\_1.shp} & Railway tracks & No filter \\
       \bottomrule
    \end{tabular}
\end{table}




\section{Brick Kiln Dataset}\label{sec:brick_kiln_dataset}
In this section, we elaborate on the process of initial data preparation, model selection, brick kiln detection and external validation with surveys.
\subsection{Initial Data Preparation}
\label{sec:initial_regions}

\begin{figure}[h]
    \centering
    \begin{subfigure}[b]{0.45\textwidth}
        \centering
        \includegraphics[width=\linewidth]{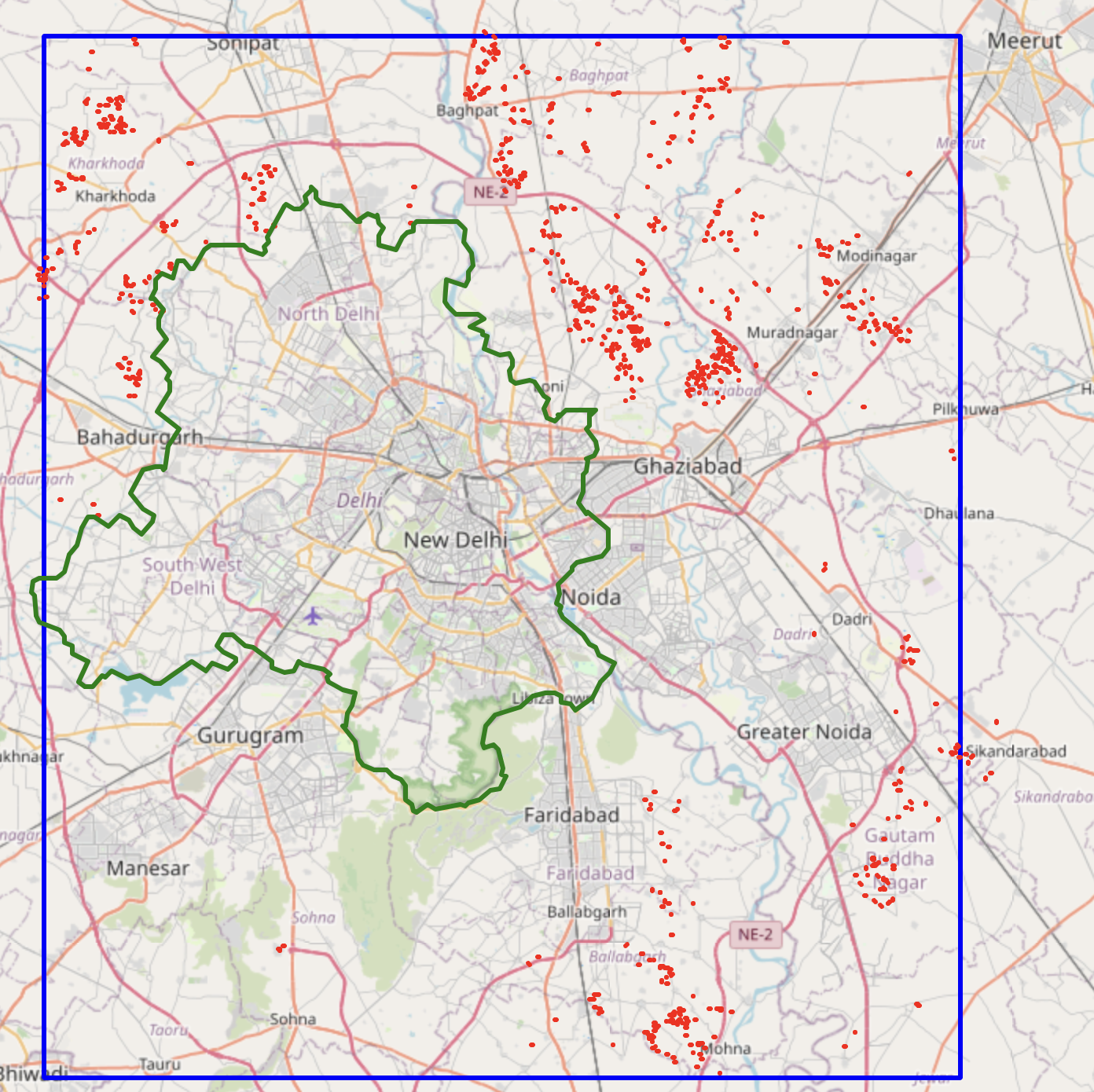} 
        \caption{Delhi Airshed}
        \label{fig:initial_regions_delhi}
    \end{subfigure}
    \hfill
    \begin{subfigure}[b]{0.45\textwidth}
        \centering
        \includegraphics[width=\linewidth]{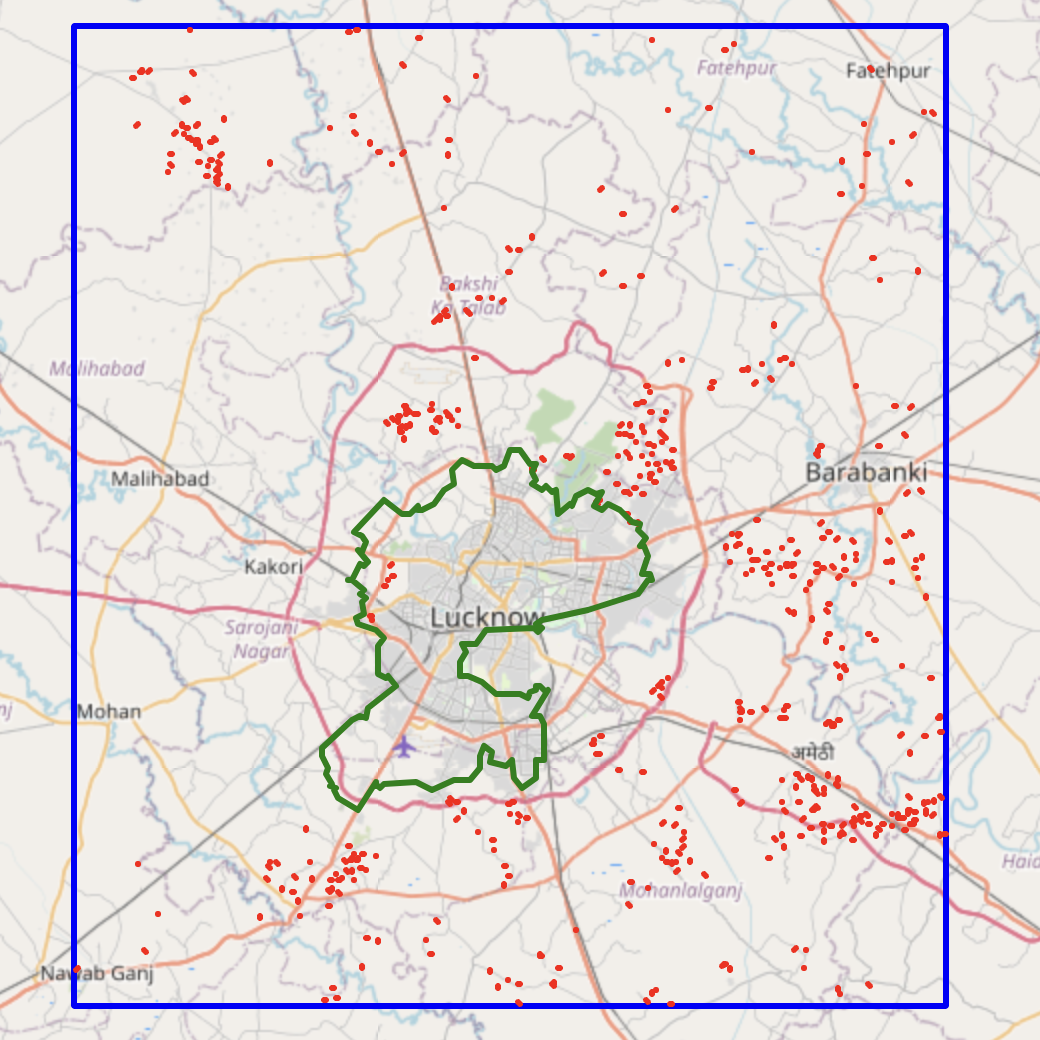} 
        \caption{Lucknow Airshed}
        \label{fig:initial_regions_lucknow}
    \end{subfigure}
    \hfill
    \begin{subfigure}[b]{0.45\textwidth}
        \centering
        \includegraphics[width=\linewidth]{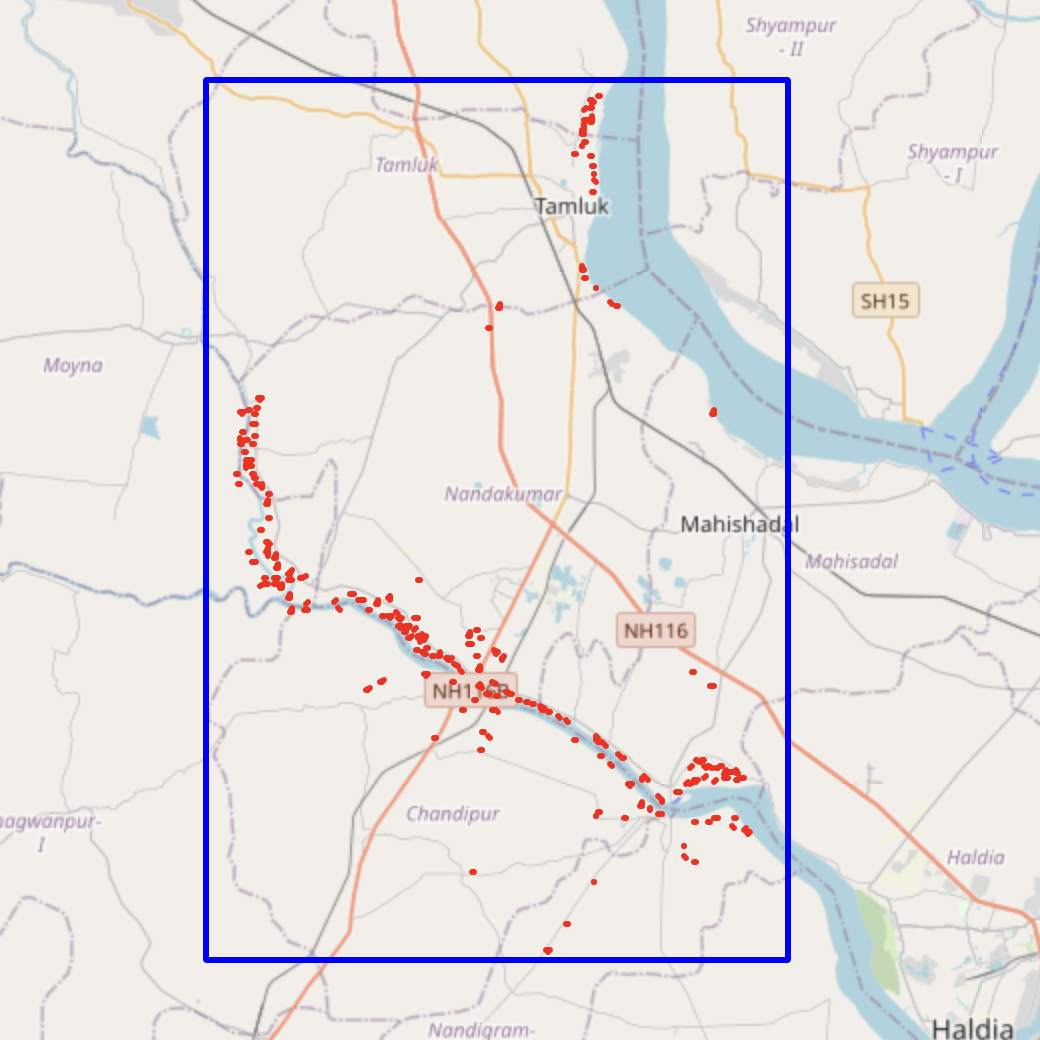} 
        \caption{West Bengal Small}
        \label{fig:initial_regions_wb}
    \end{subfigure}
    \hfill
    \begin{subfigure}[b]{0.45\textwidth}
        \centering
        \includegraphics[width=\linewidth]{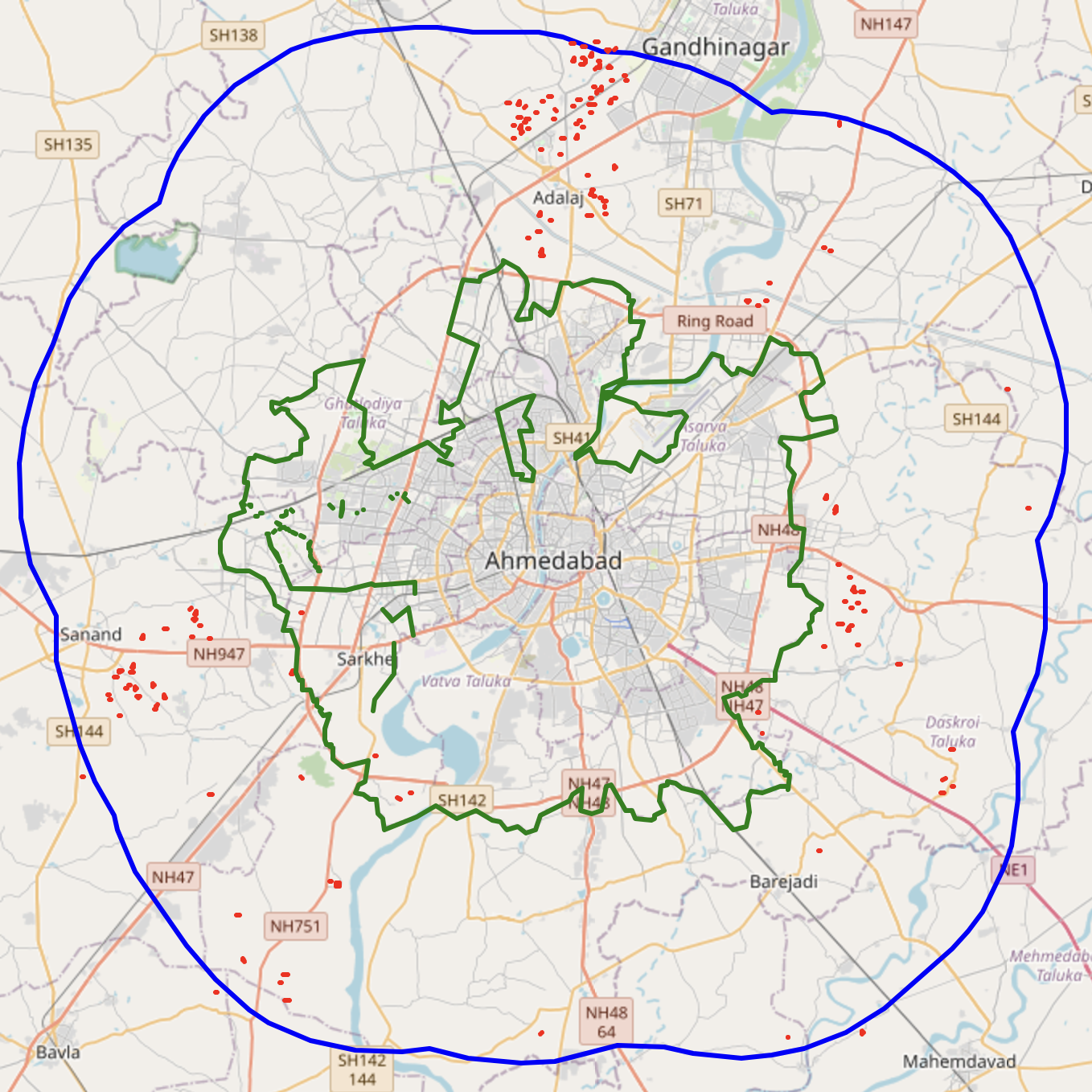} 
        \caption{Ahmedabad 10 km buffer}
        \label{fig:initial_regions_amd}
    \end{subfigure}
    \hfill
    \caption{Visualization of geographic regions for initial data annotation. The blue boundaries are the regions for annotation. The green boundaries denote the cities. The red dots denote the locations of brick kilns.}
    \label{fig:initial_regions}
\end{figure}

An initial labeled dataset is required to train the object detection models. We select the regions for initial labeling with the following criteria: i) highly polluted; ii) highly populated (highly impacted by brick kilns); iii) high likelihood of kilns (as suggested by an air quality expert); and iv) non-attainment cities.
\begin{enumerate}
\item \textbf{Delhi Airshed, India:} Delhi, India’s capital, is highly populated and consistently ranks among the world’s most polluted cities~\cite{iqair2024}. 
While no brick kilns are located within Delhi due to strict regulations, the surrounding region has a dense concentration of kilns. We selected an $80\times80$ grid over Delhi with a 0.01-degree resolution, as defined by Guttikunda et al.~\cite{GUTTIKUNDA2023119712}. The selected region is illustrated in Figure~\ref{fig:initial_regions_delhi}.
\item \textbf{Lucknow Airshed, Uttar Pradesh, India:} Lucknow, the capital of Uttar Pradesh (India’s most populous state), is classified as a non-attainment city under NCAP. A $60\times60$ airshed grid with 0.01-degree resolution was chosen for annotation, following the definition by Guttikunda et al.~\cite{GUTTIKUNDA2023119712}. The selected region is illustrated in Figure~\ref{fig:initial_regions_lucknow}.
\item \textbf{West Bengal Small Airshed, West Bengal:} A small airshed near the Haldi River in West Bengal was selected due to the high density of brick kilns in the area. A 639 km$^2$ region surrounding the identified kilns was annotated as illustrated in Figure~\ref{fig:initial_regions_wb}.
\item \textbf{Ahmedabad Buffer Region:} Ahmedabad, a non-attainment city, was studied in collaboration with the Gujarat Pollution Control Board (GPCB). Following the Environment (Protection) Amendment Rules, 2022~\cite{BrickKilnsRules2022}, we geo-located brick kilns within a 10 km buffer of Ahmedabad city. The kilns and selected region is shown in Figure~\ref{fig:initial_regions_amd}.
\end{enumerate}

As shown in Table~\ref{tab:initial_data_preperation}, we manually scanned over 15,018 km$^2$ area and identified 1,621 brick kilns across the four regions, completing the annotations in 125 hours. Table~\ref{tab:initial_data_preperation} provides region-wise details of area, annotation time, and class-wise distribution. The selected regions are illustrated in Figure~\ref{fig:initial_regions}.

\begin{table}[h]
    \centering
\begin{tabular}{lrrrrrr}
\toprule
Airshed & Area (km$^2$) & CFCBK & FCBK & Zigzag & Total Kilns & Annotation time (hours) \\
\midrule
Delhi Airshed~\cite{GUTTIKUNDA2023119712} & 6937 & 2 & 35 & 746 & 783 & 58 \\
Lucknow Airshed~\cite{GUTTIKUNDA2023119712} & 3962 & 26 & 225 & 241 & 492 & 33 \\
West Bengal Small & 639 & 0 & 89 & 110 & 199 & 5 \\
Ahmedabad 10 km buffer & 3480 & 38 & 108 & 1 & 147 & 29 \\
\midrule
Total & 15018 & 66 & 457 & 1098 & 1621 & 125 \\
\bottomrule
\end{tabular}
    \caption{Statistics of initial dataset. The `Delhi Airshed' mostly has Zigzag kilns because it contains the National Capital Region (NCR) and receives the most attention from policymakers. The `Lucknow Airshed' and `West Bengal Small' have a relatively balanced distribution of FCBK and Zigzag. `Ahmedabad 10 km buffer' region has a large number of FCBKs. All regions have less number of CFCBKs because CFCBK is an old technology, and most kilns have been upgraded to either FCBK or Zigzag over time.}
    \label{tab:initial_data_preperation}
\end{table}

\subsection{Annotation Process}\label{sec:annotation-process}
\begin{figure}[h]
    \centering
    \begin{subfigure}{0.45\textwidth}
        \centering
        \includegraphics[width=\textwidth]{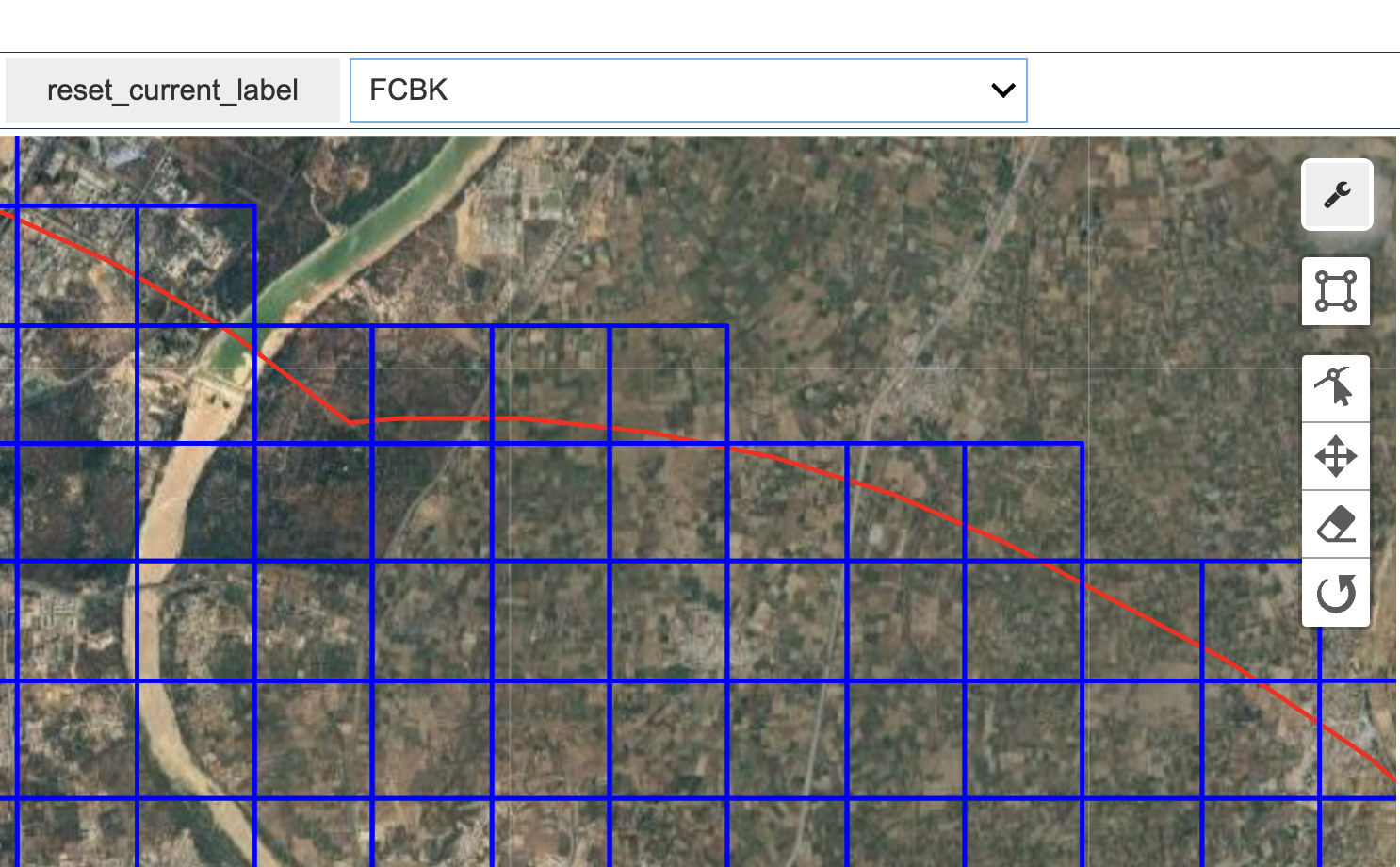}
        \caption{Grid for annotation}
        \label{fig:annotation_a}
    \end{subfigure} \hfill
    \begin{subfigure}{0.45\textwidth}
        \centering
        \includegraphics[width=\textwidth]{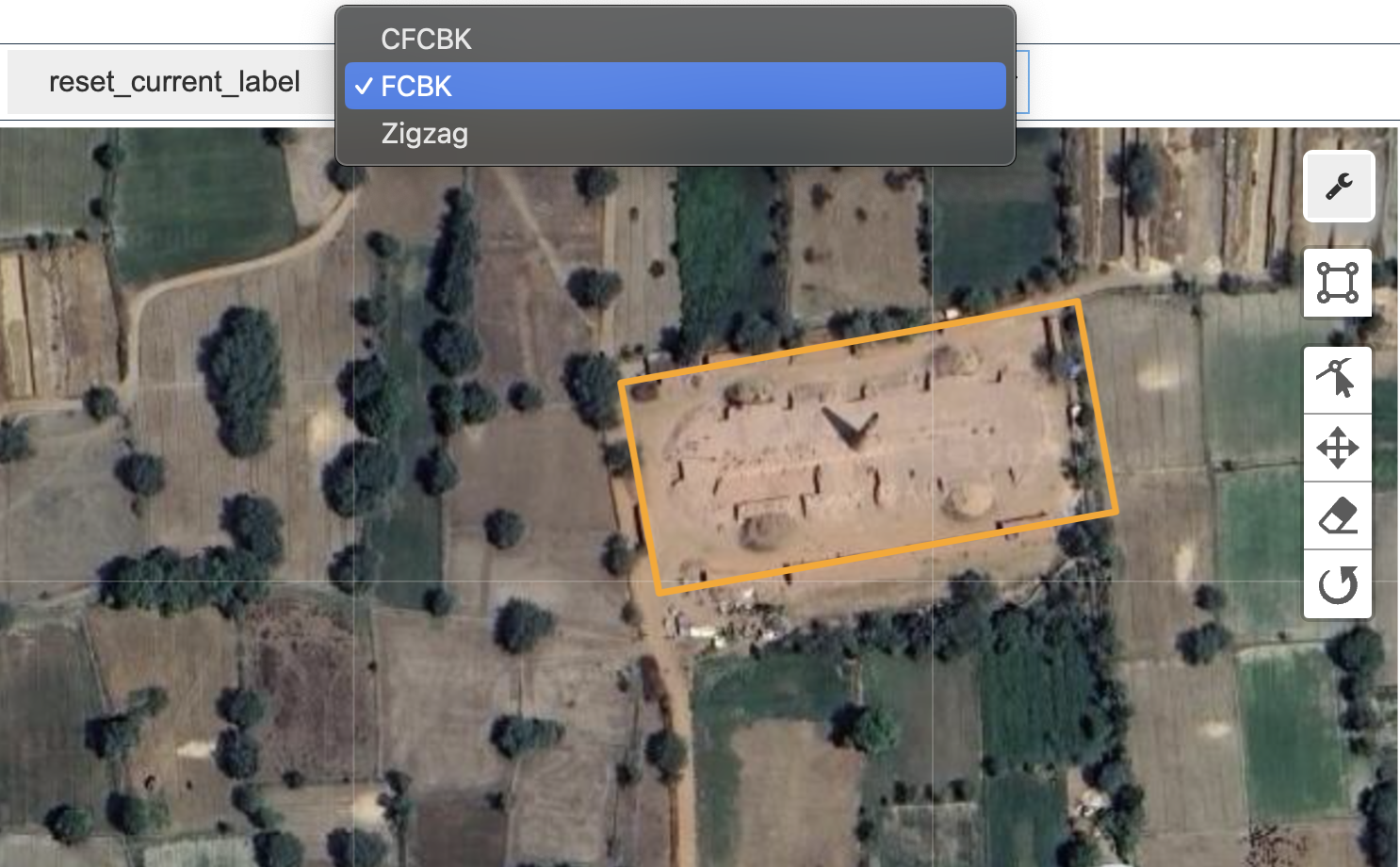}
        \caption{Labeling a kiln}
        \label{fig:annotation_b}
    \end{subfigure}
    \caption{Illustration of the labeling process. We first create a grid over the labeling region with 1 km$^2$ grid cells, as shown in (a). Blue boxes are the grids, and the red line is the boundary of the region of interest. Then, we zoom into a grid cell, put an OBB around kilns, and select the type of the kiln as shown in (b). Right menu buttons from the second button onward are used to draw, edit, move, and rotate an OBB in that order.}
    \label{fig:annotation}
\end{figure}

The annotation process involves labeling brick kilns with oriented bounding boxes (OBBs) and assigning a category to represent the kiln's technology.
We developed a custom labeling interface using the Leafmap library~\cite{Wu2021}, as shown in Figure~\ref{fig:annotation}, which utilizes the Esri Wayback Satellite Imagery WMS layer. The region of interest was divided into a set of grids (Figure~\ref{fig:annotation_a}), and each grid cell was manually inspected for labeling. \newText{If a grid cell is too small, the annotation process becomes excessively time-consuming, whereas a grid cell too large may lead to missed kilns during manual inspection. A grid size of 1 km$^2$ was found to provide a good tradeoff between speed and accuracy of annotation.} Within each cell, we draw OBBs around the identified kilns (Figure~\ref{fig:annotation_b}). \newText{The annotations created via our interface are geo-referenced, meaning that the coordinates of the bounding boxes are geo-coordinates and can thus be overlaid on any geo-referenced imagery.}
After completing the annotations, the labeled data were overlaid onto cropped, geo-referenced Planet images. Both Esri and Planet imagery are in the Web Mercator projection (EPSG:3857), and thus the OBBs remain undistorted after the reprojection from Esri to Planet imagery, as demonstrated in Figure~\ref{fig:data_samples}.

\subsection{Model Selection}
We utilize YOLO-OBB models implemented by Ultralytics~\cite{ultralytics_yolo_2023} for our experiments. These models deliver performance comparable to state-of-the-art OBB detection frameworks. YOLO's algorithm enables faster inference by detecting objects in a single forward pass, making it well-suited for scaling predictions across large geographic areas. We evaluate all OBB variants (`YOLOv8' and `YOLO11') across different sizes (`n': nano, `s': small, `m': medium, `l': large, `x': extra-large). The dataset described in Table~\ref{tab:initial_data_preperation} is combined and split into stratified subsets with an 80:10:10 ratio for training, validation, and testing. Stratification ensures a proportional representation of all classes in each subset. The training set (`train') is used to train the models, while the validation set (`validation') guides performance monitoring and early stopping. The test set (`test') evaluates final model performance. We measure model accuracy using the `Weighted mAP50' (mean Average Precision at 50\% Intersection over Union) metric, which adjusts the standard mAP50 by weighting class-specific average precisions (APs) inversely to their test dataset counts. Table~\ref{tab:map} summarizes the performance of various models on the test dataset. Among the evaluated models, \bestModel\ achieves the highest performance and is therefore used for brick kiln detection in subsequent experiments.

\begin{table}[h]
    \centering
    \caption{Performance of various models on the initial dataset. \texttt{\bestModel} model is yielding the best `Weighted mAP' among all models, and thus, we use it for brick kiln detection.}
    \label{tab:map}
   \begin{tabular}{lrrrr}
\toprule
Model & CFCBK & FCBK & Zigzag & Weighted mAP \\
\midrule
\texttt{yolov8l-obb} & 0.61 & 0.58 & 0.83 & 0.62 \\
\texttt{yolov8x-obb} & 0.63 & 0.55 & 0.82 & 0.63 \\
\texttt{yolo11x-obb} & 0.66 & 0.57 & 0.80 & 0.66 \\
\texttt{yolo11l-obb} & 0.68 & 0.51 & 0.76 & 0.66 \\
\texttt{yolov8m-obb} & 0.68 & 0.54 & 0.79 & 0.66 \\
\texttt{yolo11m-obb} & 0.73 & 0.61 & 0.83 & 0.71 \\
\bottomrule
\end{tabular}
\end{table}

\newSubsection{Out of the Region Performance}\label{sec:out_of_region} \newText{Deep learning models are not guaranteed to perform well when the test data differs significantly from the training data distribution. To evaluate the model's robustness and estimate exclusion errors (i.e., undetected kilns), we conducted a `Leave One Region Out' experiment. This involved using four initially labeled regions (described in Section~\ref{sec:initial_regions}) and iteratively designating one region for testing while training the best-performing model (as identified in Table~\ref{tab:map}) on the remaining regions. The results, presented in Table~\ref{tab:loro}, reveal that the model's mAP scores significantly drop in this setup. The Precision and Recall columns represent inclusion and exclusion errors, respectively (higher is better). Notably, Recall is consistently lower than Precision, indicating that exclusion errors are more prevalent than inclusion errors. This underscores the limited robustness of detection models when applied to regions distant from the training data. Although addressing this challenge is beyond the current paper's scope, we mitigated it during the data-building process by iteratively incorporating labeled data from all regions to enhance model robustness. These findings highlight the importance of diverse and representative training data for improving model generalization. Future work could explore advanced techniques, such as domain adaptation to further address this limitation.}

\newcommand{\colortabular}[2][\iftoggle{cameraReady}{black}{blue}]{%
    \begingroup
    \color{#1} 
    #2
    \endgroup
}

\begin{table}[h]
    \centering
    \caption{\newText{Results from the `Leave One Region Out' experiment. In each row, `Test Region' denotes the region excluded from training, while the model was trained on the remaining three regions. In addition to mean Average Precision (mAP), the number of True Positives (TP), False Positives (FP), and False Negatives (FN) are also reported. The metrics are significantly lower compared to Table~\ref{tab:map}, indicating that models perform poorly on out-of-region data when used out of the box. This highlights the importance of region-specific training data. However, in the final data-building process, we address this limitation by iteratively incorporating data from all regions. This iterative approach helps ensure more robust generalization across regions.}}
    \label{tab:loro}
    \colortabular{
   \begin{tabular}{lrrrrrrrrr}
\toprule
Test Region & CFCBK & FCBK & Zigzag & Weighted mAP & TP & FP & FN & P & R\\
\midrule
Delhi Airshed & 0.00 & 0.11 & 0.32 & 0.04 & 317 & 421 & 632 & 0.43 & 0.33\\
Lucknow Airshed & 0.32 & 0.01 & 0.54 & 0.32 & 221 & 206 & 275 & 0.52 & 0.45\\
West Bengal Small & 0.00 & 0.02 & 0.28 & 0.16 & 64 & 83 & 142 & 0.44 & 0.31\\
Ahmedabad 10 km buffer & 0.26 & 0.06 & 0.00 & 0.18 & 18 & 47 & 131 & 0.28 & 0.12\\
\bottomrule
\end{tabular}
}
\end{table}

\subsection{Iterative Dataset Building Process}
To expand our dataset, we applied the model trained on the initial data to detect brick kilns in Uttar Pradesh (UP), the most populous state in India and a significant contributor to brick kiln activity.
We employed a `hand-validation' process to ensure the accuracy of the model's predictions. The steps involved in hand-validation are as follows:
\begin{enumerate}
    \item If a bounding box is entirely incorrect (does not correspond to a brick kiln), we mark it as a false detection and discard it.
    \item If a bounding box partially overlaps a brick kiln, we adjust it to align accurately with the kiln.
    \item If the detected brick kiln is misclassified, we correct the category label.
\end{enumerate}

We used an IoU (Intersection over Union) threshold of 0.33 for Non-Maximum Suppression (NMS) to eliminate duplicate detections. This threshold balances the need to suppress overlapping boxes while retaining accurate detections. For classification confidence, we set a minimum threshold of 0.25, to gain more recall at the expense of precision.

Out of \UPMZeroPP\ detected kilns in Uttar Pradesh, \UPMZeroTP\ were found to be correct, yielding a precision of \UPMZeroPrecision\%. While the model's precision may appear low, it substantially reduces manual annotation effort. Annotating the entire Uttar Pradesh region manually would require over 2000 hours, assuming an average of 15 seconds per annotation. By contrast, the hand-validation of \UPMZeroPP\ detections requires approximately 112 hours, representing a significant efficiency improvement.

In the next iteration, we combine the initial dataset with \UPMZeroTP\ newly identified and hand-validated kilns and fine-tune the model using a validation split. The updated model is then applied to detect brick kilns across five states: Uttar Pradesh, Bihar, West Bengal, Haryana, and Punjab.
The precision of the model in this iteration is \UPFinalP\% for Uttar Pradesh, \BiharFinalP\% for Bihar, \WBFinalP\% for West Bengal, \HaryanaFinalP\% for Haryana, and \PunjabFinalP\% for Punjab. These precision values are derived by dividing the number of correctly identified kilns by the total number of detections in each state. Uttar Pradesh achieves the highest precision, as the updated training dataset includes a significant number of examples from this region, improving the model's accuracy there. Figure~\ref{fig:pred_samples} illustrates sample detections, including bounding boxes around each brick kiln, categorized by kiln type. After completing this iterative process, \textbf{we develop a comprehensive dataset containing a total of $\DTLfetch{datasetCounts}{State}{Total}{Total}$ hand-validated brick kilns with oriented bounding boxes}. Table~\ref{tab:india_bk} presents the state-wise distribution of hand-validated brick kilns, while Figure~\ref{fig:total_dataset} maps the geographic distribution of each kiln technology. From our analysis, we observe the following patterns:
\begin{itemize}
    \item CFCBKs: Predominantly found in Uttar Pradesh.
    \item FCBKs: Present in all five states but most common in Uttar Pradesh.
    \item Zigzag Kilns: Concentrated near the Delhi-NCR region, Bihar, and the eastern part of West Bengal.
\end{itemize}
This analysis, enriched with geographic and technological classifications, is valuable for further research and policy planning.

\begin{figure}[h]
    \centering
    \foreach \name in {CFCBK_106.png, CFCBK_12.png, CFCBK_131.png, CFCBK_39.png, CFCBK_5.png, CFCBK_50.png, 
    FCBK_0.png, FCBK_1.png, FCBK_10.png, FCBK_13.png, FCBK_14.png, FCBK_17.png, 
    Zigzag_11.png, Zigzag_16.png, Zigzag_18.png, Zigzag_2.png, Zigzag_20.png, Zigzag_22.png 
    }
    {
        \begin{subfigure}{0.15\textwidth}
            \centering
            \includegraphics[width=\textwidth]{figures/kiln_samples/\name}
        \end{subfigure} \hspace{0.004\textwidth}
    }
    \caption{A few samples of predicted brick kilns of each brick kiln category from our model. The first, second, and third rows show samples of CFCBK, FCBK, and Zigzag in that order. Imagery \textcopyright\ 2024 Planet Labs Inc.}
    \label{fig:pred_samples}
\end{figure}



\begin{figure}[h]
    \centering
    \begin{subfigure}[b]{0.33\textwidth}
        \centering
        \includegraphics[width=\textwidth]{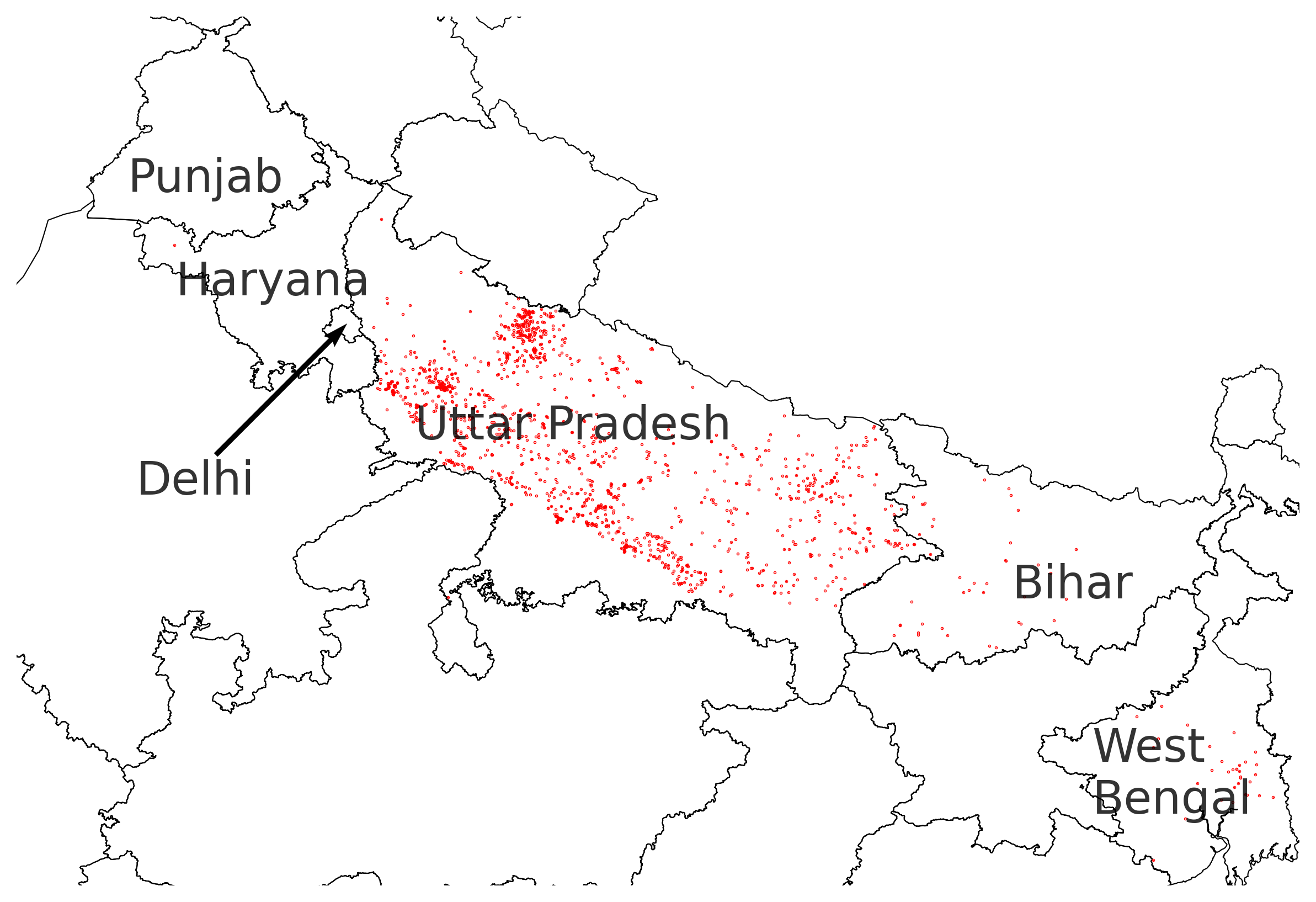}
        \caption{CFCBKs}
    \end{subfigure}
    \hfill
    \begin{subfigure}[b]{0.33\textwidth}
        \centering
        \includegraphics[width=\textwidth]{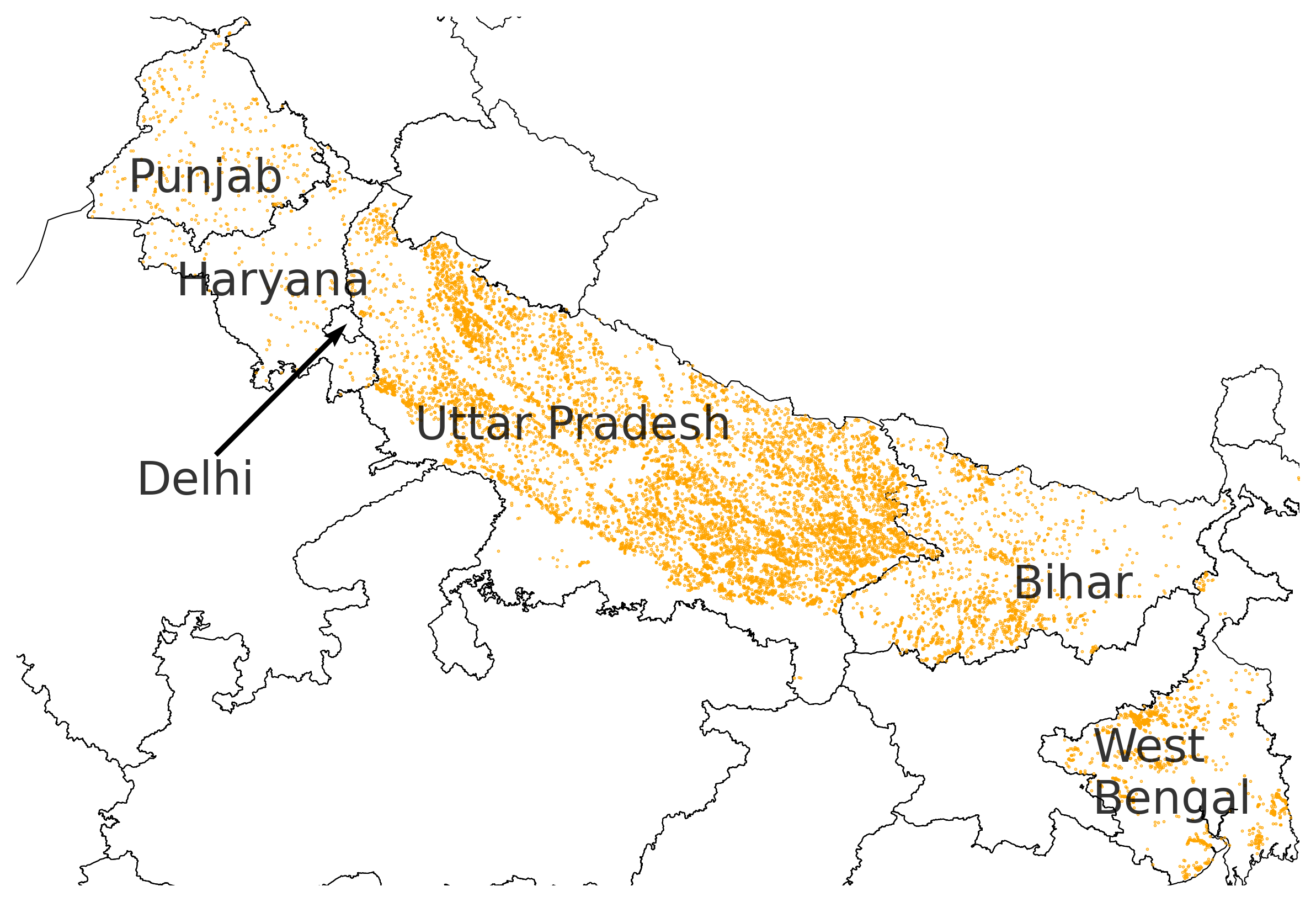}
        \caption{FCBKs}
    \end{subfigure}
    \hfill
    \begin{subfigure}[b]{0.33\textwidth}
        \centering
        \includegraphics[width=\textwidth]{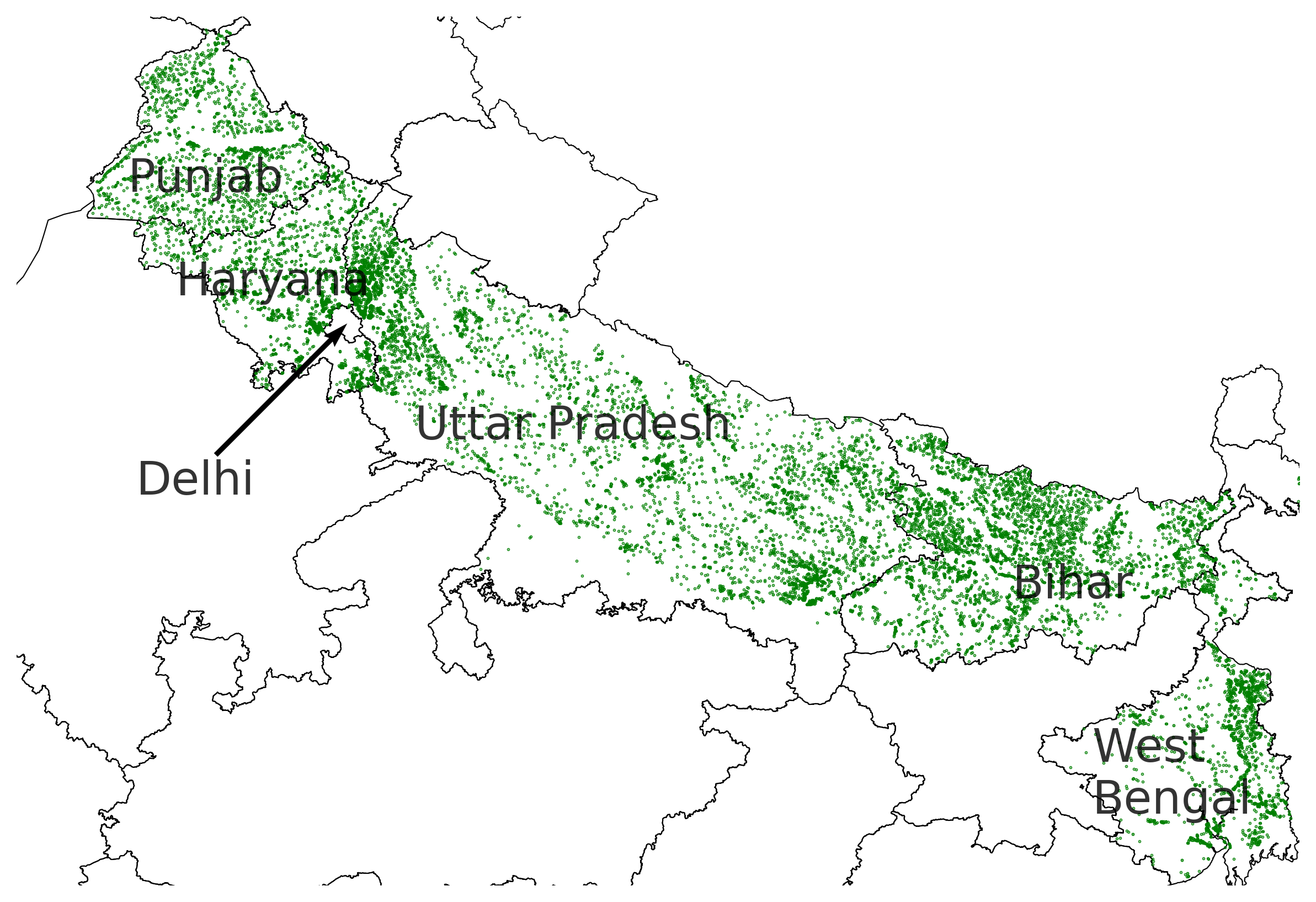}
        \caption{Zigzag Kilns}
    \end{subfigure}
    \caption{
    Brick kiln locations across five states of the Indo-Gangetic Plain, India. CFCBKs, which use a relatively old technology, are most prevalent in Uttar Pradesh. FCBKs are present in all 5 states, but their presence is strong in eastern Uttar Pradesh. Zigzag kilns are also present in all states and are densely located in Bihar and the area close to Delhi National Capital Region (Delhi-NCR).}
    \label{fig:total_dataset}
\end{figure}

\begin{table}[h]
    \centering

\caption{Brick kilns dataset across 5 states of Indo-Gangetic plain, India. We extract this data by covering 520,000 km$^2$ area with 448 million population.}
\label{tab:india_bk}
\begin{tabular}{lrrrr}
\toprule
State & CFCBK & FCBK & Zigzag & Total \\
\midrule
\DTLfetch{datasetCounts}{State}{Uttar Pradesh}{State} & \DTLfetch{datasetCounts}{State}{Uttar Pradesh}{CFCBK} & \DTLfetch{datasetCounts}{State}{Uttar Pradesh}{FCBK} & \DTLfetch{datasetCounts}{State}{Uttar Pradesh}{Zigzag} & \DTLfetch{datasetCounts}{State}{Uttar Pradesh}{Total} \\
\DTLfetch{datasetCounts}{State}{Bihar}{State} & \DTLfetch{datasetCounts}{State}{Bihar}{CFCBK} & \DTLfetch{datasetCounts}{State}{Bihar}{FCBK} & \DTLfetch{datasetCounts}{State}{Bihar}{Zigzag} & \DTLfetch{datasetCounts}{State}{Bihar}{Total} \\
\DTLfetch{datasetCounts}{State}{West Bengal}{State} & \DTLfetch{datasetCounts}{State}{West Bengal}{CFCBK} & \DTLfetch{datasetCounts}{State}{West Bengal}{FCBK} & \DTLfetch{datasetCounts}{State}{West Bengal}{Zigzag} & \DTLfetch{datasetCounts}{State}{West Bengal}{Total} \\
\DTLfetch{datasetCounts}{State}{Haryana}{State} & \DTLfetch{datasetCounts}{State}{Haryana}{CFCBK} & \DTLfetch{datasetCounts}{State}{Haryana}{FCBK} & \DTLfetch{datasetCounts}{State}{Haryana}{Zigzag} & \DTLfetch{datasetCounts}{State}{Haryana}{Total} \\
\DTLfetch{datasetCounts}{State}{Punjab}{State} & \DTLfetch{datasetCounts}{State}{Punjab}{CFCBK} & \DTLfetch{datasetCounts}{State}{Punjab}{FCBK} & \DTLfetch{datasetCounts}{State}{Punjab}{Zigzag} & \DTLfetch{datasetCounts}{State}{Punjab}{Total} \\
\midrule
\DTLfetch{datasetCounts}{State}{Total}{State} & \DTLfetch{datasetCounts}{State}{Total}{CFCBK} & \DTLfetch{datasetCounts}{State}{Total}{FCBK} & \DTLfetch{datasetCounts}{State}{Total}{Zigzag} & \DTLfetch{datasetCounts}{State}{Total}{Total} \\
\bottomrule
\end{tabular}
\end{table}


\subsection{External Validation}

\begin{figure}[h]
    \centering
    \begin{subfigure}[b]{0.33\textwidth}
        \centering
        \includegraphics[]{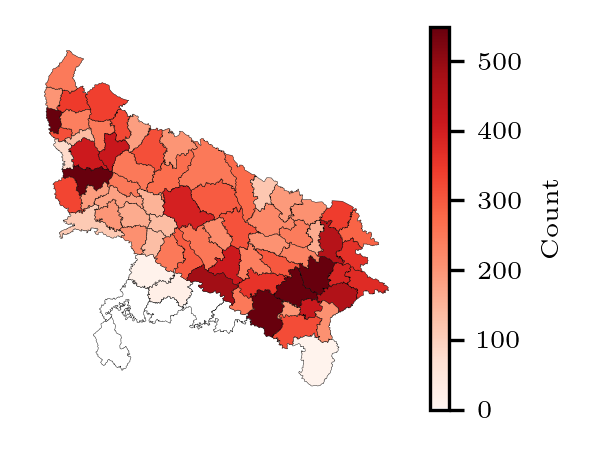}
        \caption{}
    \end{subfigure}
    \hfill
    \begin{subfigure}[b]{0.33\textwidth}
        \centering
        \includegraphics[]{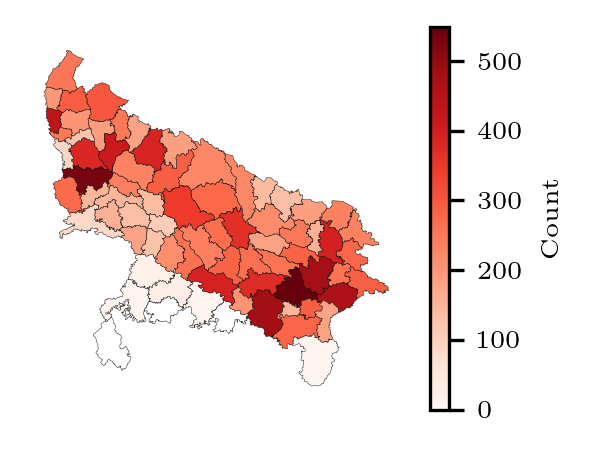}
        \caption{}
    \end{subfigure}
    \hfill
    \begin{subfigure}[b]{0.33\textwidth}
        \centering
        \includegraphics[]{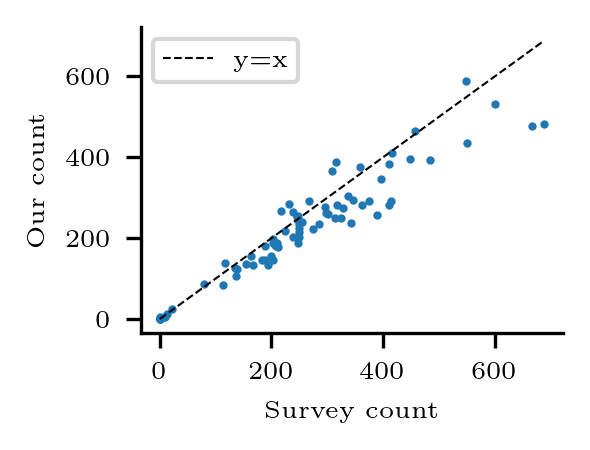}
        \caption{}
        \label{fig:fig3}
    \end{subfigure}
    \caption{District-wise brick kiln counts in Uttar Pradesh as per (a) UPPCB 2023 survey (\UPSurveyCounts\ kilns)~\cite{UPPCB2023} and (b) our hand-validated data (\UPOurCounts\ kilns). A comparison between our counts and the survey counts is shown in (c). Our counts have a Pearson correlation coefficient of 0.94 with the survey counts.
    }
    \label{fig:choropleth}
\end{figure}

\begin{figure}[h]
    \begin{subfigure}{0.33\textwidth}
    \centering
        \includegraphics[]{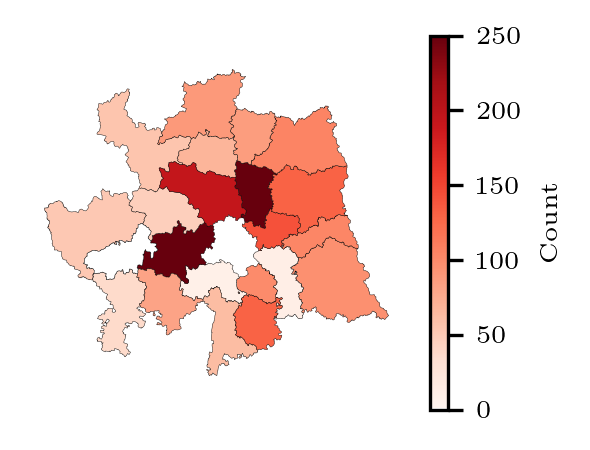}
        \caption{}
    \end{subfigure}\hfill
    \begin{subfigure}{0.33\textwidth}
    \centering
        \includegraphics[]{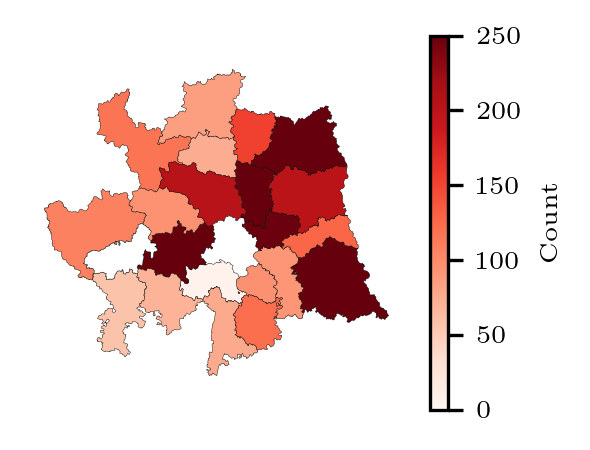}
        \caption{}
    \end{subfigure}
    \hfill
    \begin{subfigure}{0.33\textwidth}
    \centering
        \includegraphics[]{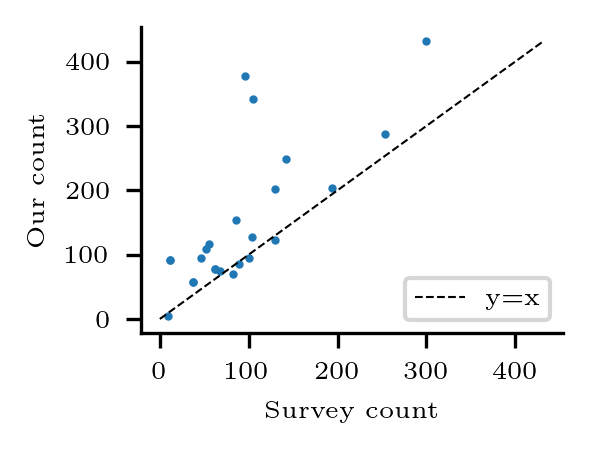}
        \caption{}
        \label{fig:delhi_ncr_choropleth_c}
    \end{subfigure}
    \caption{\newText{District-wise brick kiln counts in Delhi-NCR as per (a) CPCB 2022 survey and (b) our hand-validated data. White colored districts indicate missing survey data. A comparison between our counts and the survey counts is shown in (c). Our counts have a Pearson correlation coefficient of 0.76 with the survey counts. Our counts are higher compared to the surveyed counts which indicates establishment of new brick kilns between the survey time (2022) and our detection time (Q1, 2024).}}
    \label{fig:delhi_ncr_choropleth}
\end{figure}

\begin{figure}[h]
    \begin{subfigure}{0.33\textwidth}
    \centering
        \includegraphics[]{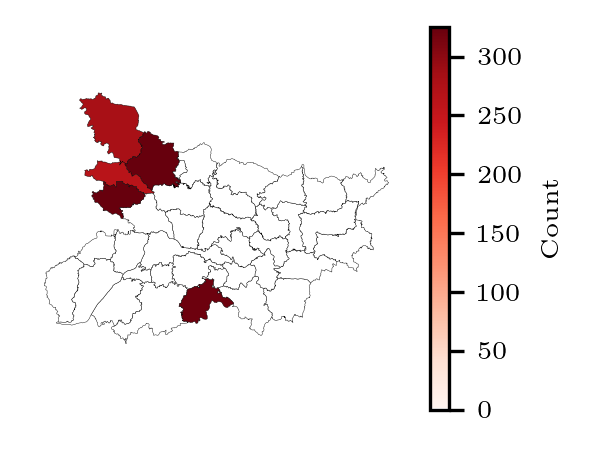}
        \caption{}
    \end{subfigure}\hfill
    \begin{subfigure}{0.33\textwidth}
    \centering
        \includegraphics[]{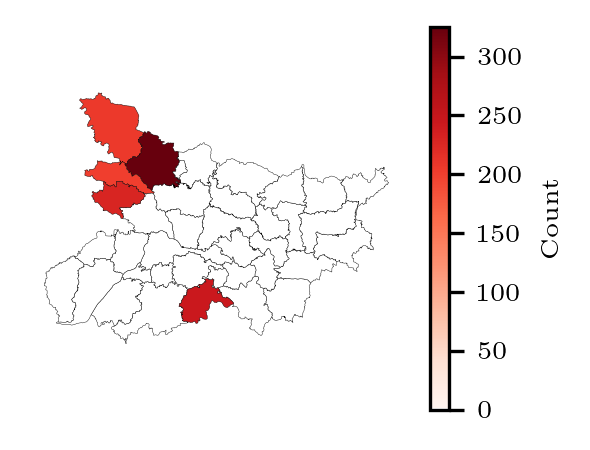}
        \caption{}
    \end{subfigure}
    \hfill
    \begin{subfigure}{0.33\textwidth}
    \centering
        \includegraphics[]{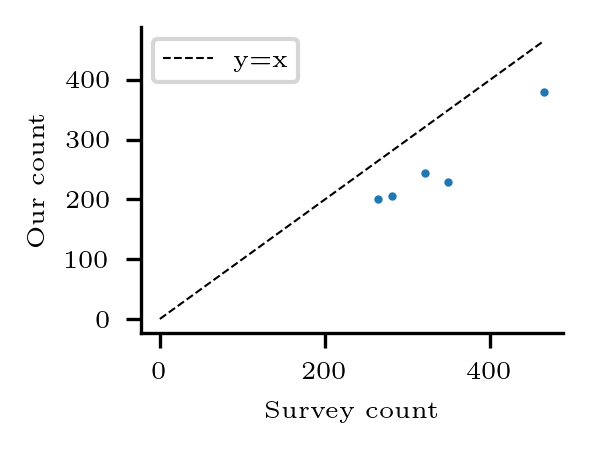}
        \caption{}
    \end{subfigure}
    \caption{\newText{District-wise brick kiln counts in Bihar as per (a) UNDP GeoAI field survey ~\cite{UNDP2023} and (b) our hand-validated data. White colored districts indicate missing survey data. A comparison between our counts and the survey counts is shown in (c). Our counts are lower compared to survey counts which could either indicate higher exclusion errors in Bihar or many kilns might have shut down permanently between survey time (2022) and our detection time (Q1, 2024). Our algorithms detect closed kilns as well but some kiln locations may not have visible traces of previously existing kilns.}}
    \label{fig:bihar_choropleth}
\end{figure}

\begin{figure}[h]
    \begin{subfigure}{0.33\textwidth}
    \centering
        \includegraphics[]{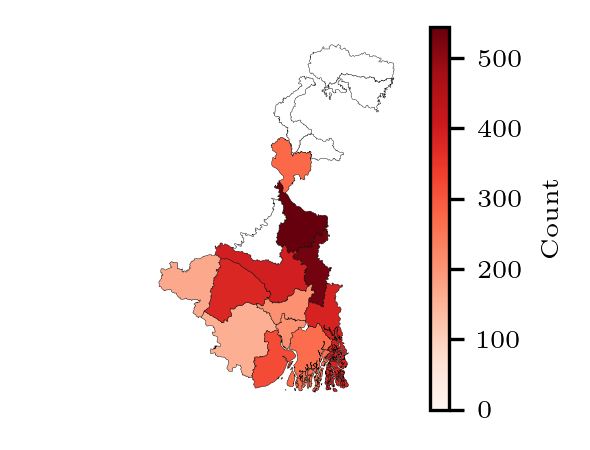}
        \caption{}
    \end{subfigure}\hfill
    \begin{subfigure}{0.33\textwidth}
    \centering
        \includegraphics[]{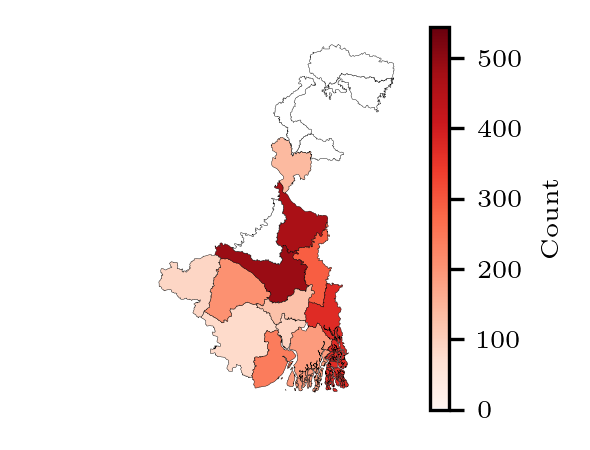}
        \caption{}
    \end{subfigure}
    \hfill
    \begin{subfigure}{0.33\textwidth}
    \centering
        \includegraphics[]{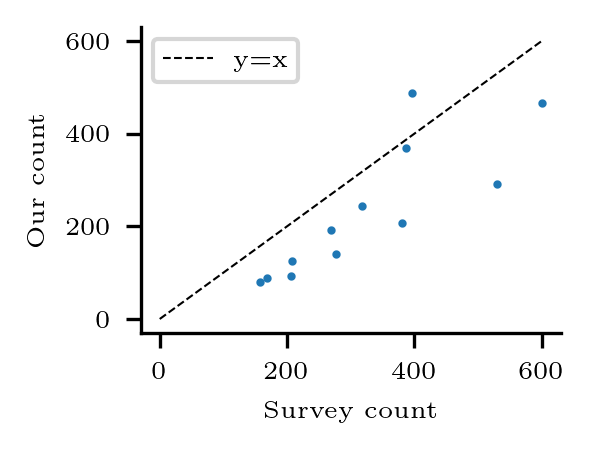}
        \caption{}
    \end{subfigure}
    \caption{\newText{District-wise brick kiln counts in West Bengal as per (a) WBSPCB 2022 survey ~\cite{bengal_brick_field_2022} and (b) our hand-validated data. White region indicates missing survey data. A comparison between our counts and the survey counts is shown in (c). Our counts have a Pearson correlation coefficient of 0.84 with the survey counts.}}
    \label{fig:wb_choropleth}
\end{figure}

\begin{table}[h]
 \begin{minipage}{0.52\textwidth}
 \centering
    \caption{\newText{District-wise brick kiln counts in Delhi-NCR. Our counts exceed the survey counts, indicating establishment of new kilns in the region between the survey time (2022) and our detection time (Q1, 2024).}}
    \label{tab:delhi_ncr_validation}
    \colortabular{
    \begin{tabular}{lrr}
        \toprule
        \textbf{District} & \textbf{Survey counts} & \textbf{Our counts} \\
        \midrule
        Baghpat & 300 & 432 \\
        Bhiwani & 52 & 108 \\
        Bulandshahr & 96 & 377 \\
        Faridabad & 100 & 95 \\
        GautamBudhNagar & 12 & 91 \\
        Ghaziabad & 142 & 248 \\
        Gurugram & 9 & 5 \\
        Hapur & 104 & 127 \\
        Jhajjar & 253 & 287 \\
        Jind & 55 & 117 \\
        Karnal & 89 & 85 \\
        Mahendergarh & 37 & 57 \\
        Meerut & 130 & 202 \\
        Mewat & 62 & 77 \\
        Muzaffarnagar & 105 & 341 \\
        Palwal & 130 & 122 \\
        Panipat & 68 & 74 \\
        Rewari & 82 & 70 \\
        Rohtak & 46 & 94 \\
        Shamli & 86 & 153 \\
        Sonipat & 194 & 204 \\
        \midrule
        \textbf{Total} & \textbf{2152} & \textbf{3366} \\
        \bottomrule
    \end{tabular}
    }
  \end{minipage}
  \hfill
    \begin{minipage}{0.45\textwidth}
    \centering
        \begin{minipage}{\textwidth}
        \caption{\newText{District-wise brick kiln counts in West Bengal. Our detection counts are lower because of higher exclusion errors due to domain shift and the likelihood that some kilns became non-operational between the survey year (2022) and our detection time (Q1, 2024).}}
        \label{tab:westbengal_validation}
        \colortabular{
        \resizebox{\textwidth}{!}{
        \begin{tabular}{lrr}
        \toprule
        \textbf{District} & \textbf{Survey counts} & \textbf{Our counts} \\
        \midrule
        Bankura & 380 & 207 \\
        Barddhaman & 396 & 488 \\
        Howrah & 206 & 92 \\
        Hugli & 208 & 126 \\
        Maldah & 276 & 140 \\
        Murshidabad & 600 & 467 \\
        Nadia & 530 & 291 \\
        North 24 Parganas & 387 & 370 \\
        Pashchim Medinipur & 158 & 80 \\
        Purba Medinipur & 317 & 244 \\
        Puruliya & 168 & 89 \\
        South 24 Parganas & 269 & 191 \\
        \midrule
        \textbf{Total} & \textbf{3895} & \textbf{2785} \\
        \bottomrule
    \end{tabular}}
    }
    \end{minipage}
    \begin{minipage}{\textwidth}
    \caption{\newText{District-wise brick kiln counts in Bihar. The table shows lower counts due to domain shift and the likelihood that some kilns became non-operational between the survey year (2022) and our detection time (Q1, 2024).}}
    \label{tab:bihar_validation}
    \colortabular{
    \resizebox{\textwidth}{!}{
    \begin{tabular}{lrr}
        \toprule
        \textbf{District} & \textbf{Survey counts} & \textbf{Our counts} \\
        \midrule
        Gopalganj & 264 & 201 \\
        Nawada & 321 & 244 \\
        Pashchim Champaran & 281 & 206 \\
        Purba Champaran & 465 & 380 \\
        Siwan & 349 & 229 \\
        \midrule
        \textbf{Total} & \textbf{1680} & \textbf{1260} \\
        \bottomrule
    \end{tabular}
    }}
\end{minipage}
    \end{minipage}
\end{table}

In 2023, the Uttar Pradesh Pollution Control Board (UPPCB) conducted a comprehensive field survey and prepared a report for judicial proceedings~\cite{uppcbbrickkiln}. This report provides district-wise counts of operational brick kilns, revealing a total of 19,671 brick kilns in Uttar Pradesh as of 2022. Though it provides the addresses of brick kilns, we do not reverse geo-locate them to avoid false mapping. 

Figure~\ref{fig:choropleth} presents a district-wise comparison of brick kiln counts derived from the UPPCB survey and our dataset. Notably, our study is the first to validate a machine-learning-derived brick kiln dataset using independently collected survey data. 
A correlation analysis, shown in Figure~\ref{fig:fig3}, demonstrates strong alignment between the two datasets, with a Pearson correlation coefficient of 0.94. The error analysis reveals a mean error of 41.8, a median error of 33.0, and a standard deviation of 40.7 across districts. These results underscore the robustness of our approach in estimating brick kiln distributions and counts at a district level.

\newText{CPCB submitted a compliance affidavit for Delhi-NCR brick kilns to the Supreme court of India in 2022\footnote{Diary No. 20331/2021, 18213/2021}. We derived the counts of brick kilns in each district within Delhi-NCR and compared them against our detections in Table~\ref{tab:delhi_ncr_validation} and Figure~\ref{fig:delhi_ncr_choropleth}. Figure~\ref{fig:delhi_ncr_choropleth_c} shows that our detections are higher compared to the counts derived from the affidavit. It indicates that between the survey time (around 2022) and our detection time (first quarter, 2024), many new brick kilns are established in the region. Similarly we derived the counts from a few districts of Bihar from the UNDP GeoAI report and reported the comparison with our counts in Figure~\ref{fig:bihar_choropleth} and Table~\ref{tab:bihar_validation}. In another case involving West Bengal Pollution Control Board\footnote{Application No. 60/2020/EZ}, a status report of conversion to Zigzag technology was submitted in the court. We derived the counts for the districts of West Bengal from the same UNDP report and presented the comparison with our counts in Figure~\ref{fig:wb_choropleth} and Table~\ref{tab:westbengal_validation}. In Bihar and West Bengal, our counts are lower compared to the surveyed counts in most districts, which could either indicate that our exclusion errors are higher or number of kilns have reduced in the region. Overall, we observe a notable alignment of our detected counts with the reported counts in the surveys, affidavits and reports.} \\


\section{Automatic Compliance Monitoring}\label{sec:compliance}

In this section, we present compliance rules and results from automatic compliance monitoring based on distance-based criteria.

\subsection{Distance-based Compliance Monitoring}
The Environment (Protection) Amendment Rules, 2022~\cite{BrickKilnsRules2022}, outline compliance requirements for brick kiln operations across all states of India. Among these, two key distance-based siting rules are: i) brick kilns must maintain a minimum distance of 800 meters from habitation and fruit orchards; ii) brick kilns must be at least 1 kilometer apart from one another. States are permitted to establish additional or stricter siting criteria. For our analysis, if both national and state-specific criteria exist, we prioritize state-specific rules. 

\begin{figure}[h]
    \centering
    \begin{subfigure}[b]{0.33\textwidth}
        \centering
        \includegraphics[]{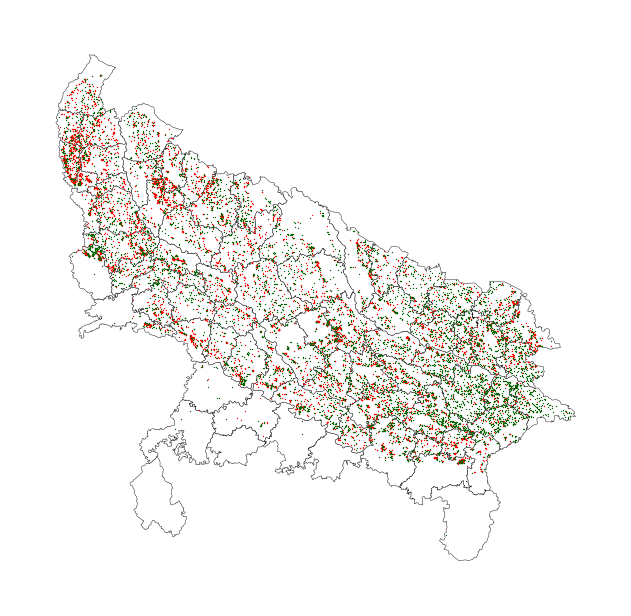} 
        \caption{}
        \label{fig:hospital}
    \end{subfigure}
    \hfill
    \begin{subfigure}[b]{0.33\textwidth}
        \centering
        \includegraphics[]{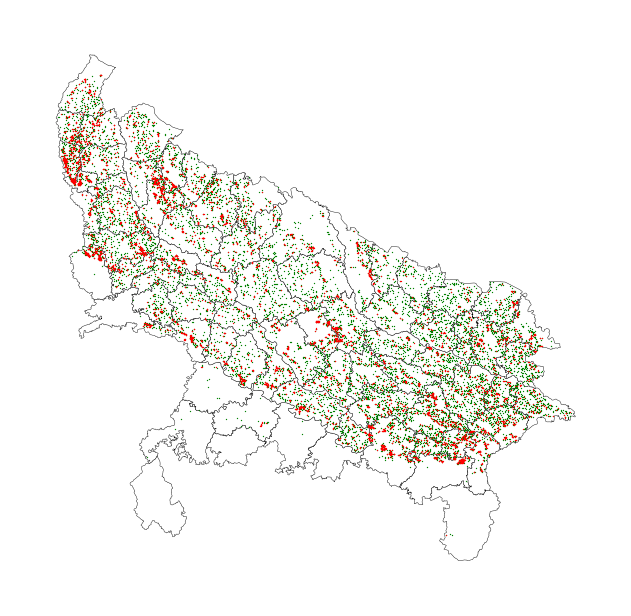} 
        \caption{}
        \label{fig:interbrickkilns}
    \end{subfigure}
    \hfill
    \begin{subfigure}[b]{0.33\textwidth}
        \centering
        \includegraphics[]{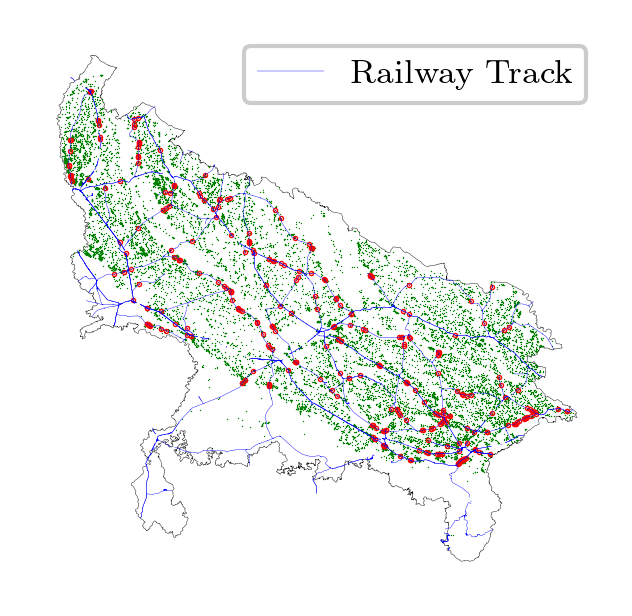} 
        \caption{}
        \label{fig:river}
    \end{subfigure}
    \caption{(a) Compliance with hospitals; (b) Compliance with inter kiln distance policy; (c) Compliance with railway tracks.}
    \label{fig:compliance_visual}
\end{figure}

We examined the compliance criteria for Uttar Pradesh~\cite{uttar_pradesh_brick_kiln_rules_2012}, Bihar, West Bengal, Haryana, and Punjab. Using the big data resources described in Section~\ref{sec:big_data}, we automatically computed distances between points of interest (e.g., habitation, hospitals, other kilns) and brick kilns. A summary of the results is provided in Table~\ref{tab:compliance_analysis}, with the corresponding distance thresholds detailed in Table~\ref{tab:compliance_distances}. We illustrate the locations of violating brick kilns in Figure~\ref{fig:compliance_visual}.
Based on our analysis, the following patterns emerge:
\begin{itemize}
    \item The most frequent violations occur with the `Inter-kiln distance' policy, indicating that brick kilns are predominantly established in clusters.
    \item The second highest violations are associated with the `hospitals' category, which specifies minimum distances from healthcare facilities.
    \item Uttar Pradesh has the most violations as well as the highest percentage of violations.
    \item Punjab and Haryana exhibit fewer violations, partly due to less stringent state-specific siting criteria for some categories.
\end{itemize}

The clustering of kilns and proximity to habitation are recurrent issues across all studied states. These findings underscore the need for better policy implementations and the utility of automatic compliance monitoring in identifying policy violations systematically and at scale.

\begin{table}[h]
    \centering
    \caption{Distance thresholds (in meters) for compliance criteria in each state as per various state-level siting criteria and central government rules.}
    \label{tab:compliance_distances}
    \begin{tabular}{lrrrrr}
\toprule
Criterion & Uttar Pradesh & Bihar & West Bengal & Haryana & Punjab \\
\midrule
Inter kiln & 800 & 1000 & 300 & 1000 & 1000 \\
Habitation & 1000 & 800 & 800 & 800 & 500 \\
National highway & 300 & 300 & 200 & - & - \\
River & - & 500 & 200 & - & - \\
State highway & 300 & 200 & 200 & - & 100 \\
District highway & 100 & - & - & - & - \\
Railway & 200 & 200 & 200 & - & - \\
Nature reserve & 5000 & - & 5000 & 1000 & - \\
Orchard & 800 & 800 & 800 & 800 & 800 \\
Wetland & - & 500 & - & - & - \\
School & 1000 & 800 & 1000 & 1000 & - \\
Religious places & 1000 & - & 1000 & - & - \\
\bottomrule
\end{tabular}
\end{table}

\begin{table}[h]
    \centering
    \caption{Automatic compliance detection of brick kilns in various states with respect to state-wise and central policies. `-' suggests that a policy is not defined for that state-criterion pair. The `Non compliant' row shows the total counts of brick kilns which violate at least one rule. More than 70\% of brick kilns violate at least one compliance rule.}
    \label{tab:compliance_analysis}
    \begin{tabular}{llrrrrrr}
\toprule
 &  & \multicolumn{5}{c}{State} \\
 &  & Uttar Pradesh & Bihar & West Bengal & Haryana & Punjab & Total \\
\midrule
\multirow[c]{13}{*}{Criterion} & Inter kiln & 6469 & 2343 & 669 & 1035 & 604 & 11120 \\
 & Hospital & 6741 & 656 & 1271 & 176 & - & 8844 \\
 & Habitation & 5765 & 1471 & 362 & 131 & 327 & 8056 \\
 & National highway & 1495 & 579 & 128 & - & - & 2202 \\
 & River & - & 1029 & 338 & - & - & 1367 \\
 & State highway & 720 & 150 & 96 & - & 40 & 1006 \\
 & District highway & 695 & - & - & - & - & 695 \\
 & Railway & 358 & 122 & 84 & - & - & 564 \\
 & Nature reserve & 328 & - & 21 & 0 & - & 349 \\
 & Orchard & 120 & 8 & 4 & 0 & 11 & 143 \\
 & Wetland & - & 70 & - & - & - & 70 \\
 & School & 19 & 15 & 26 & 4 & - & 64 \\
 & Religious places & 6 & - & 3 & - & - & 9 \\
 \midrule
 & Non compliant & 13296 & 3997 & 2081 & 1162 & 866 & 21402 \\
 & Brick Kiln count & 17335 & 6048 & 3105 & 2079 & 2071 & 30638 \\
 & Percentage violations & 77 & 66 & 67 & 56 & 42 & 70 \\
\bottomrule
\end{tabular}
\end{table}

\subsection{Technology-based Compliance Monitoring}\label{sec:tech_compliance}
As per the guidelines issued by the Central Pollution Control Board (CPCB)~\cite{cpcb}, new brick kilns in non-attainment cities are required to adopt induced-draft zigzag technology. To track the technology used in brick kilns over time, we leverage high-resolution historical imagery from Esri, which allows us to identify and classify kiln technologies on a year-by-year basis. This analysis can be approached in the following ways:
\begin{itemize}
    \item \textbf{Automatic method:} We can apply a detection model to identify kilns across different time periods, capturing both active and potentially closed kilns. This method requires fine-tuning our model on past imagery to optimize detection accuracy.
    \item \textbf{Manual method:} Alternatively, we manually backtrack the locations of kilns and document their transition or establishment timestamps. This approach misses kilns that may have existed earlier but are now closed. We acknowledge this limitation and discuss recommendations for improving future methodologies in Section~\ref{sec:lim_and_future_work}.
\end{itemize}

For this work, we opt for the manual method to accurately determine the year of establishment or conversion for each kiln. We employ a binary search strategy~\cite{Lin2019BinarySA} on the kilns detected by our model and hand-validated, ensuring a high level of precision in identifying key transitions over time.

\begin{figure}[h]
    \centering
    \begin{subfigure}{0.45\textwidth}
        \includegraphics[width=\linewidth]{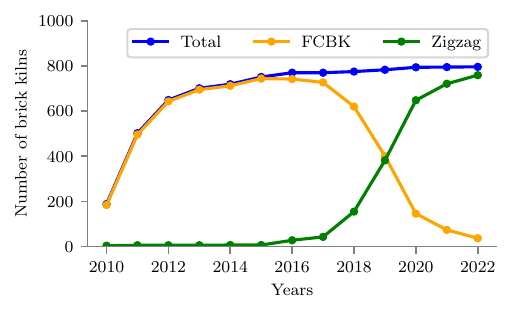}
        \caption{}
        \label{fig:delhi}
    \end{subfigure}
    \hfill
    \begin{subfigure}{0.45\textwidth}
        \includegraphics[width=\linewidth]{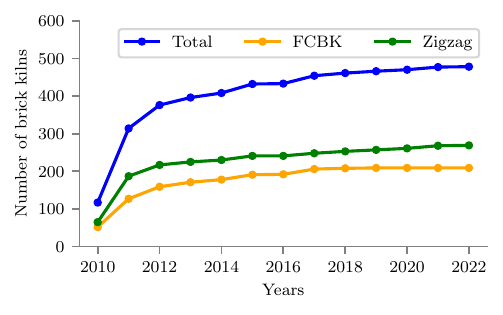}
        \caption{}
        \label{fig:lucknow}
    \end{subfigure}
    \caption{
    Brick kiln technology evolution over 12 years (2010-2022) for two NCAP non-attainment city airsheds: (a) Delhi Airshed~\cite{GUTTIKUNDA2023119712}, (b) Lucknow Airshed~\cite{GUTTIKUNDA2023119712}. Around 2017-21 in Delhi, there was a drastic change in technology from FCBK to Zigzag kilns, which could be attributed to strong policy enforcement in the region. In Lucknow, FCBK and Zigzag kiln growth has stabilized over the years, but no drastic technology shift has been observed. Lucknow started adapting to the Zigzag technology much earlier than Delhi.
    }
    \label{fig:conversion}
\end{figure}

We start by examining imagery from 2016, the midpoint of the 2010–2022 timeframe. If kilns were detected in 2016, we then check 2013 (the midpoint between 2010 and 2016) to refine the timeline. We continue narrowing the search in this manner, progressively focusing on shorter periods until we pinpoint the exact year of kiln establishment or technology transition. If kilns were not detected in the selected year, we adjust the timeframe accordingly, moving forward or backward as needed.
Using this approach, we classify the kiln technology for each year, beginning in 2010. The resulting counts of kilns by technology over time are shown in Figure~\ref{fig:conversion}. 
Notably, in the Delhi airshed (Figure~\ref{fig:delhi}), we observe a sharp transition from Fixed Chimney Bull’s Kilns (FCBK) to Zigzag kilns around 2016, aligning with the stricter enforcement of policies. In Figure~\ref{fig:lucknow}, we observe that Lucknow airshed, including another non-attainment city, Lucknow, has adopted Zigzag technology kilns much earlier than Delhi, but the growth of FCBK and Zigzag has been similar and declining over time. Due to the policy regulations over time, we do not observe a decline in the counts of FCBKs over time in Lucknow airshed. This analysis, enabled by high-resolution historical satellite imagery and advanced detection algorithms, not only reveals technological shifts in kiln practices across regions but also underscores the significant role of data-driven monitoring in tracking the adoption of cleaner technologies.
    

\section{Air Pollution and Health Effects}\label{sec:air_pollution_and_HE}
\subsection{Emissions}

In this section, we estimate air pollutant emissions from brick kilns. We derive emission rates (g/kg) from previous studies~\cite{rajarathnam2014, weyant2014emissions}, as listed in Table~\ref{tab:emission_factors}. Tibrewal et al.~\cite{tibrewal2023reconciliation} provide annual brick production data (kg/year). Brick kilns typically operate for six months each year, from 15th November to 15th May in Bihar and West Bengal, and from 15th December to 15th June in Uttar Pradesh, Punjab, and Haryana, based on expert inputs. We calculate daily production by dividing the annual production by 180 days.
We estimate emissions in tonnes per day for each state by combining emission rates (g/kg) for each technology with the mass of bricks produced (kg). We allocate the total production among different technologies based on our detections. We performed the calculations as follows (for Uttar Pradesh):

\begin{align*}
    \text{CFCBK + FCBK} &= 1450 + 9933 = 11443 \text{ kilns}\\
    \text{Zigzag kilns} &= 5952 \text{ kilns}\\
    \text{Total kilns} &= 17335 \text{ kilns}\\
    \text{PM$_{2.5}$ emissions} &= \left[\left(0.18 \text{ g/kg} \times 11443 \text{ kilns} + 0.09 \text{ g/kg} \times 5952 \text{ kilns}\right) / 17335 \text{ kilns}\right] \times (794816.67 \times 1000) \text{ kg} \\
    &= 118.51 \text{ tonnes}
\end{align*}

The final emission estimates are shown in Table~\ref{tab:emissions}. This analysis enables policymakers to assess the state-level contribution of brick kilns to various pollutants. Integrating these estimates with precise brick kiln locations can support the development of fine-grained emission inventories and facilitate detailed source apportionment studies as we discuss in the next section.

\begin{table}[h]
    \begin{minipage}{0.35\textwidth}
    \centering
    \caption{Emission rates for kiln technologies in g/kg of fired brick product from previous work~\cite{rajarathnam2014, weyant2014emissions}.}
        \label{tab:emission_factors}
        \begin{tabular}{lrr}
            \toprule
            Pollutant & CFCBK/FCBK & Zigzag \\
            \midrule
            PM$_{2.5}$  & 0.18 & 0.09   \\
            SO$_2$    & 0.52 & 0.15   \\
            CO & 3.63 & 1.19 \\
            CO$_2$ & 179.00 & 107.50 \\
            \bottomrule
        \end{tabular}
    \end{minipage}
    \hfill
     \begin{minipage}{0.64\textwidth}
    \centering
    \caption{Production of brick kilns and Emissions (tonnes per day) in various states}
        \label{tab:emissions}
        \begin{tabular}{lrrrrr}
\toprule
State & Mass & PM$_{2.5}$ & SO$_2$ & CO & CO$_2$ \\
\midrule
Uttar Pradesh & 794816.67 & 118.51 & 312.33 & 2219.30 & 122759.72 \\
Bihar & 401661.11 & 45.86 & 100.15 & 741.14 & 50890.09 \\
West Bengal & 321627.78 & 39.56 & 91.86 & 670.36 & 43003.29 \\
Haryana & 124283.33 & 11.89 & 21.54 & 167.01 & 13920.39 \\
Punjab & 99488.89 & 10.27 & 20.34 & 154.14 & 11742.67 \\
\midrule
Total & 1741877.78 & 226.08 & 546.23 & 3951.95 & 242316.16 \\
\bottomrule
\end{tabular}
    \end{minipage}
\end{table}

\subsection{Source Apportionment}
Emissions can be estimated from various sources, but they do not translate to air pollution without accounting for meteorology and complex physicochemical reactions in the atmosphere. PM$_{2.5}$ and PM$_{10}$ are major pollutants linked to respiratory diseases in humans~\cite{thangavel2022recent}. 
Using guidance from an air quality expert, we estimated PM$_{2.5}$ and PM$_{10}$ concentrations in the Delhi Airshed by employing the Chemical Transport Model (CTM), CAMx~\cite{camx}. CAMx utilizes meteorological inputs generated by the Weather Research and Forecasting (WRF) model~\cite{Skamarock2019}. We ran the model over the Delhi Airshed on an 80 km × 80 grid, covering the months of March and April 2024.  

Our findings reveal that brick kilns contributed 8\% of the total PM$_{2.5}$ in the Delhi Airshed during this period. This analysis provides a foundation for creating decision support systems to assess the impact of transitioning brick kiln technologies, such as converting to cleaner Zigzag technology or alternative fuels, on air quality. Such systems can guide policymakers in designing targeted interventions to reduce emissions and mitigate public health risks.

\subsection{Population Exposure}\label{sec:population}
In Section~\ref{sec:compliance} and Table~\ref{tab:compliance_analysis}, we observed that many brick kilns are found close to human habitation.
According to an analysis of the Air Quality Life Index conducted by the Energy Policy Institute at the University of Chicago (EPIC)~\cite{aqli2024}, an average citizen living in the Indo-Gangetic plain is likely to lose seven-and-half years of life. 
We use the global population data `LandScan' from Oak Ridge National Laboratory~\cite{lebakula2024landscan} to estimate the population living near the brick kilns. As we show in Table~\ref{tab:population_brick_kilns}, 30.66 million people live within 800 meters of brick kilns, which is a non-complying range as per the central government policy~\cite{BrickKilnsRules2022}.
\begin{table}[h]
\centering
\caption{Population (in millions) living within 0.8 km, 2 km, and 5 km of brick kilns across five states in the Indo-Gangetic Plain. According to central government policy, brick kilns must be at least 800 meters away from human habitation. However, more than 30 million people live within this range of brick kilns~\cite{BrickKilnsRules2022}. Our analysis highlights the need for targeted studies to assess the health impacts of this proximity and evaluate interventions, such as relocating kilns or adopting cleaner technologies, to mitigate risks.}
\label{tab:population_brick_kilns}
\begin{tabular}{lrrr}
\toprule
State & < 0.8 km & < 2 km & < 5 km \\
\midrule
Uttar Pradesh & 13.81 & 63.32 & 168.83 \\
Bihar         & 9.43  & 44.22 & 98.41   \\
West Bengal   & 4.35  & 18.54 & 50.47  \\
Punjab        & 1.95  & 10.03 & 25.64   \\
Haryana       & 1.12  & 6.34  & 19.36   \\
\midrule
Total & 30.66 & 142.45 & 362.71  \\
\bottomrule
\end{tabular}
\end{table}


\section{Discussion}\label{sec:discussion}
\newText{In this section, we discuss the broader implications of our findings, focusing on socio-economic, environmental, and policy aspects. We explore how this research can aid decision-making in the brick kiln sector without disrupting the livelihoods of workers. Additionally, we examine how our data can enhance air quality modeling by improving emission inventories, the importance of our work with respect to policies on transitioning kilns to Zigzag technology as per MoEFCC guidelines, and the challenges of isolating brick kiln pollution from satellite data.}

\newSubsection{Socio-Economic Considerations}
Our study demonstrates the potential of machine learning and satellite imagery to monitor compliance and assess the environmental impact of brick kilns at scale. By identifying violations and estimating contributions to air pollution, our work provides actionable insights for policymakers. However, such insights must be contextualized within the socio-economic realities of the brick kiln industry to ensure that interventions are equitable and sustainable. The brick kiln sector is not just a contributor to air pollution but also a critical source of livelihoods for approximately 15 million workers in India. Many of these workers come from underprivileged backgrounds, with limited alternative employment opportunities. Regulatory interventions, if implemented abruptly, risk adversely affecting these vulnerable populations. For instance, stringent enforcement of compliance rules or sudden shutdowns of non-compliant kilns could lead to loss of employment for thousands of workers.

To address this, we suggest the following changes.
\begin{itemize}
    \item We advocate for gradual implementation timelines and targeted support for kiln owners and workers. This includes skill development programs to enable smoother transitions to advanced kiln technologies like Zigzag kilns, which reduce emissions and improve working conditions for the workers.
    \item In this work, we find that various states have uneven policies defined for the brick kilns. In our opinion, policies must be informed by evidence-based studies like ours, which integrate spatial data, emission estimates, and population exposure. This ensures that interventions are effective without being unduly punitive.
    \item The health and safety of workers in brick kilns are often overlooked in policy discussions. Cleaner technologies can potentially mitigate occupational hazards, such as respiratory illnesses caused by prolonged exposure to particulate matter and harmful gases. Complementary measures, such as access to healthcare and education, are essential for long-term social upliftment of the workers.
\end{itemize} 

Our work provides an opportunity to develop sustainable and inclusive policies for the brick kiln sector. By addressing both environmental and human dimensions, this approach aligns with the goals of cleaner air, improved public health, and social equity.

\newSubsection{Data Implications:} \newText{Access to accurate and up-to-date location data for brick kilns is crucial for predicting air pollution. Our method can be utilized to update kiln locations every month or season. This information can support the air quality modeling community by improving emission inventories, considering changes in kiln technology that affect emissions and dust resuspension from land use. Currently, estimating the daily and seasonal production of a brick kiln is extremely difficult, making it challenging to determine its emissions. However, with further analysis on a higher resolution imagery, this study can be extended to record the designed daily production capacity of each kiln and the number of days or weeks it operates in a year, leading to more accurate emission estimates. Production capacity can be estimated by automatically computing the trench width of kilns. Production cycle and operating status can be estimated by detecting the open fuel holes on top of the brick kilns. Over time, these holes keep moving to process the next batch of the bricks, which can give us an estimate of production in a certain duration. This approach not only enhances the precision of emission modeling but also provides valuable insights for policymakers to regulate industrial activities more effectively. Furthermore, the integration of such data could facilitate the development of targeted mitigation strategies to reduce air pollution from brick kilns.}

\newSubsection{Policy Implications} \newText{Understanding the extent of air pollution caused by brick kilns is important both in terms of location and time. Spatially, it is necessary to determine how far and wide brick kilns are spread within and around an urban area. Temporally, it is essential to know when these kilns are active and how they impact local air quality. Additionally, evaluating the compliance rates of operational brick kilns is crucial, not only by examining past data but also by tracking them in near real-time while considering the rapid pace of urbanization. The Ministry of Environment, Forests, and Climate Change (MoEFCC) has set a revised deadline requiring all brick kilns to switch to zigzag kiln technology by February 2026. Checking the compliance rate at the district level will help state regulators identify districts that are falling behind and take additional steps to improve compliance in these areas. Monitoring compliance over time (Fig~\ref{fig:conversion}) also provides important insights into how long it takes for a new technology to spread. This information is valuable not only for researchers studying how small industries adopt new technologies but also for policymakers in setting realistic timelines for achieving full adoption of a new technology requirement. Furthermore, continuous monitoring of compliance will enable policymakers to assess whether additional incentives or enforcement measures are needed to ensure sustained adherence to environmental standards. Such efforts are particularly critical in regions experiencing rapid urbanization, where the environmental and health impacts of industrial activities are magnified.}

\newSubsection{Challenges in Isolating Brick Kiln Pollution from Satellite Data} \newText{Brick kilns primarily emit CO$_2$, CO, SO$_2$, and NO$_x$. Satellite products, such as Sentinel-5p~\cite{ESA_Sentinel5P}, provide global coverage of air quality and greenhouse gases. Recent studies~\cite{trenchev2023huge, meier2023advancing} have demonstrated the effectiveness of Sentinel-5p in identifying emissions from power plants and coal mines. However, isolating brick kiln emissions from satellite data presents significant challenges due to the following factors: \begin{itemize} \item \textbf{Low spatial resolution:} The spatial resolution of Sentinel-5p is 7 x 7 km$^2$, whereas a typical brick kiln measures approximately 200 x 200 m. This coarse resolution makes it difficult to detect the opening or closure of individual brick kilns using emission data. Furthermore, the large footprint of satellite measurements often includes multiple emission sources, complicating the attribution of emissions to specific brick kilns. \item \textbf{Low intensity emissions:} Brick kiln emissions are significantly weaker compared to those from power plants. As shown in Table~\ref{tab:emissions}, a single brick kiln emits only 0.12 tonnes of CO per day. In contrast, a coal-fired power plant in India emits 48 tonnes of CO per day~\citep{guttikunda2014atmospheric}, which is 400 times higher than the emissions from a brick kiln. The combination of low-resolution data and low-intensity emissions makes it particularly challenging to isolate brick kiln contributions from satellite observations. This limitation highlights the need for alternative or complementary approaches, such as higher-resolution satellite systems or ground-based measurements, to better quantify brick kiln pollution. \end{itemize}}

\newText{However, this does not indicate that brick kiln pollution is significantly lower than that of power plants, due to the fact that the number of brick kilns is much higher compared to the number of power plants. For example, as of 2025, Uttar Pradesh has 19 major coal-based power
plants\footnote{\url{https://en.wikipedia.org/wiki/Category:Coal-fired_power_stations_in_Uttar_Pradesh}} and around 20,000 brick kilns (approximately 1000x), which indicates that the contribution of brick kilns to CO generation is more than twice that of power plants in Uttar Pradesh.
}

\section{Related Work}\label{sec:related_work}
In this section, we discuss related work in the following areas: object detection from aerial imagery, the application of satellite imagery for addressing UN Sustainable Development Goals (SDGs), impact of brick kilns on air pollution and health, brick kiln detection and air quality modeling.

\subsection{Object Detection from Aerial Imagery}
Object detection in aerial imagery is challenging due to (i) the sparse and non-uniform distribution of targets, which complicates efficient detection, and (ii) the small size of focal objects, such as pedestrians, which often blend with their surroundings. This contrasts with datasets like MS COCO~\cite{lin2014microsoft}, where objects are larger and more distinguishable. Benchmark datasets like DOTA~\cite{xia2018dota} (18 categories, 11,268 images, and 1,793,658 instances) and HRSC2016~\cite{liu2017high} (1,070 high-resolution images with 2,976 ship instances) have been instrumental in advancing object detection for aerial imagery.
Recent methods~\cite{xie2021oriented, han2021redet, yang2022arbitrary} have leveraged oriented bounding boxes (OBBs) to improve detection accuracy for irregularly oriented objects. Our work also involves object detection from satellite imagery, and we have created total \DTLfetch{datasetCounts}{State}{Total}{Total} geo-referenced OBBs as part of this work.

\subsection{Computer Vision on Satellite Imagery for UN SDGs}
The United Nations’ 17 SDGs have been a focal point for research since their introduction in 2016~\cite{unGoalDepartment}. Satellite imagery and computer vision techniques have been effectively employed to address various SDGs. For instance, Burke et al.~\cite{burke2021using} applied computer vision to large-scale satellite data for agriculture, economics, and informal settlement detection. Xi et al.~\cite{xi2023satellite} developed a dataset supporting 25 SDG indicators, such as population, nighttime light, poverty, health, and living environment, to aid urban policymaking and SDG monitoring. 
Kumar et al.~\cite{ayush2020generating} used object detection on satellite imagery to measure poverty, aligning with SDGs by informing resource allocation and livelihood improvements. Yeh et al.~\cite{yeh2020using} predicted economic well-being across 20,000 African villages using multi-spectral imagery and deep learning. We have discussed the intersection of our work with UN SDGs in Section~\ref{sec:un_sdg}.

\subsection{\newText{Impact of Brick Kilns on Air Pollution and Health}} \newText{In this subsection, we highlight studies related to the impact of brick kilns on air pollution and health. \citet{guttikunda2009impact, guttikunda2008estimating} studied the impact of brick kiln air pollution on health in Dhaka, Bangladesh. Their research revealed that air pollution from brick kilns contributes to nearly 5000 premature deaths annually in Dhaka city. \citet{guttikunda2013particulate} estimated that brick kilns account for 23 to 30\% of air pollution in various clusters within the Greater Dhaka region, Bangladesh. Furthermore, \citet{guttikunda2014nature} demonstrated that transitioning to Zigzag and Hoffman technologies can reduce PM emissions by 40\% and 90\%, respectively, compared to the traditional Fixed Chimney Bull's Trench Kiln (FCBK) technology. \citet{le2010integrated} conducted a study in Northern Vietnam, estimating emission factors of brick kilns and using the Industrial Source Complex Short-Term (ISCST3) dispersion model to assess PM, CO, and SO$_2$ concentrations. Their findings indicated that SO$_2$ emissions exceeded National Ambient Air Quality Standards over 63 km$^2$ of a 100 km$^2$ modeled region during 2006-07. Similarly, \citet{skinder2014brick} monitored emissions from a cluster of kilns in Jammu \& Kashmir, finding that all measured pollutants, including oxides of sulphur (SOx), oxides of nitrogen (NOx), Respirable Suspended Particulate Matter (RSPM), and Non-Respirable Suspended Particulate Matter (NRSPM), exceeded National Ambient Air Quality Standards (NAAQS) during operational phases. \citet{rajarathnam2014} derived emission rates from various brick kiln technologies, estimating annual PM emissions of 0.94 million tonnes. In Delhi, \citet{guttikunda2013health} estimated that brick kilns contribute 4-17\% to air pollution, attributing 300 to 2700 premature deaths and 0.24-1.02 million asthma attacks annually to brick kiln-generated air pollution. \citet{raza2021impact} assessed the lung health of brick kiln workers in Kasur district, Pakistan, finding that 78\% of workers exhibited abnormal lung function. A comprehensive review by \citet{nicolaou2024brick} of 104 studies revealed consistent adverse health outcomes among brick kiln workers, including reduced lung function, increased respiratory symptoms, and higher rates of musculoskeletal complaints compared to unexposed populations. These findings underscore the critical need for stricter regulations and cleaner technologies to mitigate the significant health and environmental impacts of brick kiln emissions.}

\subsection{\newText{Brick Kiln Detection}} In this subsection, we provide an overview of previous work focused on brick kiln detection, highlighting advancements in both manual efforts and detection algorithms. \newText{\citet{boyd2018slavery} conducted a comprehensive manual survey using Google Earth, covering 32,000 km$^2$ and identifying 1,142 brick kilns. Building on these findings, \citet{li2019aging} developed an algorithmic approach to analyze kiln growth, revealing that most kilns are less than 10 years old, indicating a significant recent expansion in their numbers.} \newText{\citet{nazir2019tiny} introduced the Tiny-Inception-ResNet-v2 model, trained on 782 Maxar images at zoom level 20, and evaluated it on 700 images at zoom level 17, successfully identifying 102 brick kilns.} \citet{nazir2020kiln} advanced this work by developing Kiln-Net, a gated neural network, which was trained and evaluated over a 3,300 km$^2$ area using 1,300 kiln images and 521 instances. Their model detected 3,392 brick kilns, demonstrating improved accuracy in kiln identification. \newText{\citet{misra2020mapping} employed transfer learning with VGG-16 on Sentinel-2 imagery, detecting 1,564 brick kilns in the Delhi NCR region. This represented a 1.5-fold increase over manual detection methods reported by \citet{guttikunda2013gis} in the same area.} \citet{lee2021scalable} developed a fast detection method based on classification, mapping all brick kilns in Bangladesh and classifying them into FCBK and Zigzag types for the first time. Their study identified a total of 6,978 kilns. \newText{\citet{dewitt2021remote} conducted a geospatial analysis of brick kilns in Kabul, Afghanistan, analyzing changes over four time periods: 1965, 2004, 2011, and 2018. Their findings revealed a doubling of kilns from 303 in 1965 to 625 in 2018.} \newText{\citet{tahir2021brick} evaluated the performance of the YOLOv3 model for brick kiln detection in a region of Bangladesh.} \newText{\citet{paul2022brick} assessed the performance of three CNN-based detection models using Maxar WorldView-2 imagery for brick kiln detection.} \newText{\citet{imaduddin2023detection} applied pixel-level classification to Sentinel-2 imagery for brick kiln detection in the Delhi-NCR region, though their approach did not focus on identifying individual kilns.} \newText{\citet{hamdani2024brick} combined pixel classification of Sentinel-2 imagery with YOLOv8 on Google Maps imagery to classify kilns into FCBK and Zigzag types, detecting 11,000 brick kilns in the Indo-Gangetic Plain of Pakistan.} \newText{\citet{mondal2024scalable} applied CNN-based models to detect brick kilns across 28 highly populated districts in five Indian states, identifying 7,477 kilns with 86\% precision.} Our study represents the first large-scale effort to geo-locate and classify brick kilns by technology type (CFCBK, FCBK, Zigzag) across five states in the Indo-Gangetic Plain, covering an expansive area of 520,000 km$^2$ using medium-resolution, free-for-research imagery. We identified \DTLfetch{datasetCounts}{State}{Total}{Total} kilns, validated district-wise counts against field surveys, and analyzed the temporal evolution of kiln technologies. 

\begin{table}[h]
    \centering
        \caption{ 
\newText{
Comparison of previous studies on brick kiln detection and our work. Ours is the first work to use a freely available medium resolution imagery to detect and classify a high number of brick kilns over a large area.
}}
    \resizebox{\textwidth}{!}{%
    \colortabular{
\begin{tabular}{lllrrr}
    \toprule
    \textbf{Paper ID} & \textbf{Satellite Imagery} & \textbf{Classification} &\textbf{Area (km$^2$)} & \textbf{Annotated Kilns} & \textbf{Detected + Verified Kilns}\\ 
    \midrule
    \citet{boyd2018slavery} & Google Earth & No & 32,000 & 1142 & 0 \\
    \citet{nazir2019tiny} & Maxar & No & - & 102 & 0 \\
    \citet{nazir2020kiln} & Maxar & No & 3,300 & 521 & 3392 \\ 
    \citet{misra2020mapping} & Sentinel-2 & No & 6,400 & 282 & 1564 \\
    \citet{lee2021scalable} & Maxar & FCBK, Zigzag & 148,460 & - & 6978 \\ 
    \citet{boyd2021informing} & Airbus & FCBK & 1,551,997 & 1142 & 66455 \\ 
    \citet{paul2022brick} & Maxar & FCBK & - & 641 & - \\
    \citet{hamdani2024brick} & Sentinel-2, Google Maps & FCBK, Zigzag & 518,000 & 670 & 11000 \\
    \citet{mondal2024scalable} & Google Maps & No & 276,000 & 1042 & 7477 \\
    \midrule 
    \textbf{Our Study} & \textbf{Planet Labs} & \textbf{CFCBK, FCBK, Zigzag} & \textbf{520,775} & \textbf{1,621} & \textbf{\DTLfetch{datasetCounts}{State}{Total}{Total}} \\
    \bottomrule
\end{tabular}
        }
    }
    \label{tab:brick_kiln}
\end{table}

\newSubsection{Air Quality Modeling} \newText{Brick kiln detection helps update the emission inventory, which is a critical component used in air quality modeling. In this section, we highlight notable research work in air quality modeling. Agarwal et al.~\cite{PollutionMapper} utilized NOAA's HYSPLIT model and NASA's GEOS-CF dataset to track particle movements and identify air pollution sources in Delhi. Our work can enhance the accuracy of the emission inventory used in the GEOS-CF modeling framework, thereby improving the precision of~\citet{PollutionMapper}. \citet{GUTTIKUNDA2019124} conducted gridded chemical transport modeling of air pollution using the CAMx model~\cite{camx}, incorporating emission inventory and meteorology data modeled with WRF~\cite{Skamarock2019WRF}. Similar studies have employed models such as ISCST3~\cite{le2010integrated}, AERMOD~\cite{kanabkaew2015assessment}, and CALPUFF~\cite{xie2021atmospheric}. Another research direction~\cite{cheng2018neural, Patel_Purohit_Patel_Sahni_Batra_2022, hu2023graph, liang2023airformer, wang2025airradar} focuses on modeling air quality data using ground-based sensor observations and additional predictors, including meteorological data, Points of Interest (PoIs), and road networks. 
}

\section{Limitations and Future Work}\label{sec:lim_and_future_work}
In this section, we outline the limitations of our study and propose directions for future research.

\begin{enumerate}
\item As we establish in Figure~\ref{fig:choropleth}, our coverage closely resembles the manual survey conducted by the Uttar Pradesh State Pollution Control Board; however, identifying whether a kiln is operational at a given time remains challenging with RGB imagery. During operation, kilns exhibit elevated temperatures, presenting an opportunity to use thermal sensors from satellites for status detection. Current satellite thermal imagery, with resolutions around 100 meters, lacks the precision needed for such applications. Future advancements in satellite technologies may make operational status detection feasible.
    \item In Section~\ref{sec:tech_compliance} and Figure~\ref{fig:conversion}, we manually analyzed the growth of brick kilns over time for two non-attainment cities. Extending this effort to the state level is resource-intensive and requires manual input. Fine-tuning our detection models on historical satellite imagery could enable automated assessment of kiln technology evolution over time.
    \item We have demonstrated a reasonably accurate precision and practically useful recall of our studies. However, moderate-resolution of the imagery poses a challenge in identifying kiln technologies accurately. One may utilize a high-resolution imagery to improve the accuracy of technology identification.
    \item \newText{We explored the out-of-region performance of models in Section~\ref{sec:out_of_region} and found that models perform poorly in out-of-region settings. This limitation underscores the need for more robust generalization capabilities in current architectures. Future work can leverage domain adaptation techniques to address this challenge, potentially reducing exclusion errors and improving model applicability across diverse scenarios. Additionally, exploring adaptive methods that account for regional differences could enhance the reliability of models in real-world applications.}
\end{enumerate}

\section{Conclusion}
In this work, we geo-locate and hand-validate a large-scale dataset of \DTLfetch{datasetCounts}{State}{Total}{Total}\ brick kilns along with their technology, across five states of Indo-Gangetic plain. We then identify the brick kilns that violate central and state-level compliance policies. We provide emissions estimations, the percentage contribution of brick kilns, and population exposure to brick kiln air pollution. Our work establishes a pipeline to study and analyze a large-scale unorganized sector such as brick kilns and find insights directly consumable by policymakers, researchers, and air quality experts. We also discuss the problems with current policies and how they can negatively impact the lives of the brick kiln workers. We hope our work is utilized by policymakers to form better policies towards a positive impact on the brick sector of India and millions of lives.


\begin{acks}
We sincerely thank Dr. Sameer Maithel, a respected researcher in the brick kilns area, for his valuable review and constructive feedback on our work. His insights have been instrumental in strengthening our paper. We would like to acknowledge the funding support from Ministry of Education (Grant no: AI-powered Sustainable Cities/AICoE/2024/AIS/21000).
\end{acks}

\bibliographystyle{ACM-Reference-Format}
\bibliography{main}


\begin{thebibliography}{86}


\ifx \showCODEN    \undefined \def \showCODEN     #1{\unskip}     \fi
\ifx \showDOI      \undefined \def \showDOI       #1{#1}\fi
\ifx \showISBNx    \undefined \def \showISBNx     #1{\unskip}     \fi
\ifx \showISBNxiii \undefined \def \showISBNxiii  #1{\unskip}     \fi
\ifx \showISSN     \undefined \def \showISSN      #1{\unskip}     \fi
\ifx \showLCCN     \undefined \def \showLCCN      #1{\unskip}     \fi
\ifx \shownote     \undefined \def \shownote      #1{#1}          \fi
\ifx \showarticletitle \undefined \def \showarticletitle #1{#1}   \fi
\ifx \showURL      \undefined \def \showURL       {\relax}        \fi
\providecommand\bibfield[2]{#2}
\providecommand\bibinfo[2]{#2}
\providecommand\natexlab[1]{#1}
\providecommand\showeprint[2][]{arXiv:#2}

\bibitem[Bri({[n.\,d.]})]%
        {BrickKilnsRules2022}
 \bibinfo{year}{[n.\,d.]}\natexlab{}.
\newblock \bibinfo{title}{Environment (Protection) Amendment Rules, 2022}.
\newblock \bibinfo{howpublished}{\url{https://cpcb.nic.in/uploads/Industry-Specific-Standards/Effluent/74-brick_kiln.pdf}}.
\newblock
\newblock
\shownote{[Accessed 09-11-2024]}.


\bibitem[esr({[n.\,d.]})]%
        {esriWaybackServer}
 \bibinfo{year}{[n.\,d.]}\natexlab{}.
\newblock \bibinfo{title}{Esri Wayback Imagery --- esri.com}.
\newblock \bibinfo{howpublished}{\url{https://www.esri.com/arcgis-blog/products/arcgis-living-atlas/imagery/wayback-server-connection-in-pro/}}.
\newblock
\newblock
\shownote{[Accessed 10-11-2024]}.


\bibitem[unG({[n.\,d.]})]%
        {unGoalDepartment}
 \bibinfo{year}{[n.\,d.]}\natexlab{}.
\newblock \bibinfo{title}{{G}oal 8 | {D}epartment of {E}conomic and {S}ocial {A}ffairs --- sdgs.un.org}.
\newblock \bibinfo{howpublished}{\url{https://sdgs.un.org/goals/goal8\#targets_and_indicators}}.
\newblock
\newblock
\shownote{[Accessed 08-11-2024]}.


\bibitem[cpc({[n.\,d.]})]%
        {cpcbNCAP}
 \bibinfo{year}{[n.\,d.]}\natexlab{}.
\newblock \bibinfo{title}{{N}{C}{A}{P} --- prana.cpcb.gov.in}.
\newblock \bibinfo{howpublished}{\url{https://prana.cpcb.gov.in}}.
\newblock
\newblock
\shownote{[Accessed 09-11-2024]}.


\bibitem[NIN({[n.\,d.]})]%
        {NINHealthDataGov}
 \bibinfo{year}{[n.\,d.]}\natexlab{}.
\newblock \bibinfo{title}{NIN health facilities geo code and additional parameters}.
\newblock \bibinfo{howpublished}{\url{https://www.data.gov.in/resource/nin-health-faclities-geo-code-and-additional-parameters-updated-till-last-month}}.
\newblock
\newblock
\shownote{[Accessed 26-11-2024]}.


\bibitem[Pla({[n.\,d.]})]%
        {PlanetAP93:online}
 \bibinfo{year}{[n.\,d.]}\natexlab{}.
\newblock \bibinfo{title}{Planet API Reference}.
\newblock \bibinfo{howpublished}{\url{https://developers.planet.com/docs/basemaps/reference/}}.
\newblock
\newblock
\shownote{(Accessed on 11/08/2024)}.


\bibitem[Agarwal et~al\mbox{.}(2024)]%
        {PollutionMapper}
\bibfield{author}{\bibinfo{person}{Dhruv Agarwal}, \bibinfo{person}{Srinivasan Iyengar}, {and} \bibinfo{person}{Pankaj Kumar}.} \bibinfo{year}{2024}\natexlab{}.
\newblock \showarticletitle{PollutionMapper: Identifying Global Air Pollution Sources}.
\newblock \bibinfo{journal}{\emph{ACM J. Comput. Sustain. Soc.}} \bibinfo{volume}{2}, \bibinfo{number}{1}, Article \bibinfo{articleno}{7} (\bibinfo{date}{Jan.} \bibinfo{year}{2024}), \bibinfo{numpages}{23}~pages.
\newblock
\urldef\tempurl%
\url{https://doi.org/10.1145/3617129}
\showDOI{\tempurl}


\bibitem[Agency(2017)]%
        {ESA_Sentinel5P}
\bibfield{author}{\bibinfo{person}{European~Space Agency}.} \bibinfo{year}{2017}\natexlab{}.
\newblock \bibinfo{title}{Sentinel-5P}.
\newblock
\newblock
\urldef\tempurl%
\url{https://sentinel.esa.int/web/sentinel/missions/sentinel-5p}
\showURL{%
\tempurl}
\newblock
\shownote{Accessed: 2025-04-01}.


\bibitem[Ahmad et~al\mbox{.}(2012)]%
        {ahmad2012hydrogen}
\bibfield{author}{\bibinfo{person}{Muhammad~Nauman Ahmad}, \bibinfo{person}{Leon~JL van~den Berg}, \bibinfo{person}{Hamid~Ullah Shah}, \bibinfo{person}{Tariq Masood}, \bibinfo{person}{Patrick B{\"u}ker}, \bibinfo{person}{Lisa Emberson}, {and} \bibinfo{person}{Mike Ashmore}.} \bibinfo{year}{2012}\natexlab{}.
\newblock \showarticletitle{Hydrogen fluoride damage to vegetation from peri-urban brick kilns in Asia: a growing but unrecognised problem?}
\newblock \bibinfo{journal}{\emph{Environmental pollution}}  \bibinfo{volume}{162} (\bibinfo{year}{2012}), \bibinfo{pages}{319--324}.
\newblock


\bibitem[Ayush et~al\mbox{.}(2020)]%
        {ayush2020generating}
\bibfield{author}{\bibinfo{person}{Kumar Ayush}, \bibinfo{person}{Burak Uzkent}, \bibinfo{person}{Marshall Burke}, \bibinfo{person}{David Lobell}, {and} \bibinfo{person}{Stefano Ermon}.} \bibinfo{year}{2020}\natexlab{}.
\newblock \showarticletitle{Generating interpretable poverty maps using object detection in satellite images}.
\newblock \bibinfo{journal}{\emph{arXiv preprint arXiv:2002.01612}} (\bibinfo{year}{2020}).
\newblock


\bibitem[Board(2019)]%
        {cpcb}
\bibfield{author}{\bibinfo{person}{Central Pollution~Control Board}.} \bibinfo{year}{2019}\natexlab{}.
\newblock \bibinfo{title}{{Central Pollution Control Board}}.
\newblock \bibinfo{howpublished}{\url{https://cpcb.nic.in/National-Air-Quality-Index/}}.
\newblock
\newblock
\shownote{[Online; accessed 02-December-2024]}.


\bibitem[Boyd et~al\mbox{.}(2018)]%
        {boyd2018slavery}
\bibfield{author}{\bibinfo{person}{Doreen~S Boyd}, \bibinfo{person}{Bethany Jackson}, \bibinfo{person}{Jessica Wardlaw}, \bibinfo{person}{Giles~M Foody}, \bibinfo{person}{Stuart Marsh}, {and} \bibinfo{person}{Kevin Bales}.} \bibinfo{year}{2018}\natexlab{}.
\newblock \showarticletitle{Slavery from space: Demonstrating the role for satellite remote sensing to inform evidence-based action related to UN SDG number 8}.
\newblock \bibinfo{journal}{\emph{ISPRS journal of photogrammetry and remote sensing}}  \bibinfo{volume}{142} (\bibinfo{year}{2018}), \bibinfo{pages}{380--388}.
\newblock


\bibitem[Boyd et~al\mbox{.}(2021)]%
        {boyd2021informing}
\bibfield{author}{\bibinfo{person}{Doreen~S Boyd}, \bibinfo{person}{Bertrand Perrat}, \bibinfo{person}{Xiaodong Li}, \bibinfo{person}{Bethany Jackson}, \bibinfo{person}{Todd Landman}, \bibinfo{person}{Feng Ling}, \bibinfo{person}{Kevin Bales}, \bibinfo{person}{Austin Choi-Fitzpatrick}, \bibinfo{person}{James Goulding}, \bibinfo{person}{Stuart Marsh}, {et~al\mbox{.}}} \bibinfo{year}{2021}\natexlab{}.
\newblock \showarticletitle{Informing action for United Nations SDG target 8.7 and interdependent SDGs: Examining modern slavery from space}.
\newblock \bibinfo{journal}{\emph{Humanities and Social Sciences Communications}} \bibinfo{volume}{8}, \bibinfo{number}{1} (\bibinfo{year}{2021}).
\newblock


\bibitem[Burke et~al\mbox{.}(2021)]%
        {burke2021using}
\bibfield{author}{\bibinfo{person}{Marshall Burke}, \bibinfo{person}{Anne Driscoll}, \bibinfo{person}{David~B Lobell}, {and} \bibinfo{person}{Stefano Ermon}.} \bibinfo{year}{2021}\natexlab{}.
\newblock \showarticletitle{Using satellite imagery to understand and promote sustainable development}.
\newblock \bibinfo{journal}{\emph{Science}} \bibinfo{volume}{371}, \bibinfo{number}{6535} (\bibinfo{year}{2021}), \bibinfo{pages}{eabe8628}.
\newblock


\bibitem[camx(2019)]%
        {camx}
\bibfield{author}{\bibinfo{person}{camx}.} \bibinfo{year}{2019}\natexlab{}.
\newblock \bibinfo{title}{{camx}}.
\newblock \bibinfo{howpublished}{\url{https://www.camx.com/}}.
\newblock
\newblock
\shownote{[Online; accessed 02-December-2024]}.


\bibitem[Cheng et~al\mbox{.}(2018)]%
        {cheng2018neural}
\bibfield{author}{\bibinfo{person}{Weiyu Cheng}, \bibinfo{person}{Yanyan Shen}, \bibinfo{person}{Yanmin Zhu}, {and} \bibinfo{person}{Linpeng Huang}.} \bibinfo{year}{2018}\natexlab{}.
\newblock \showarticletitle{A neural attention model for urban air quality inference: Learning the weights of monitoring stations}. In \bibinfo{booktitle}{\emph{Proceedings of the AAAI conference on artificial intelligence}}, Vol.~\bibinfo{volume}{32}.
\newblock


\bibitem[DeWitt et~al\mbox{.}(2021)]%
        {dewitt2021remote}
\bibfield{author}{\bibinfo{person}{Jessica~D DeWitt}, \bibinfo{person}{Peter~G Chirico}, \bibinfo{person}{Marissa~A Alessi}, {and} \bibinfo{person}{Kathleen~M Boston}.} \bibinfo{year}{2021}\natexlab{}.
\newblock \showarticletitle{Remote sensing inventory and geospatial analysis of brick kilns and clay quarrying in Kabul, Afghanistan}.
\newblock \bibinfo{journal}{\emph{Minerals}} \bibinfo{volume}{11}, \bibinfo{number}{3} (\bibinfo{year}{2021}), \bibinfo{pages}{296}.
\newblock


\bibitem[{Energy Policy Institute at the University of Chicago (EPIC)}(2024)]%
        {aqli2024}
\bibfield{author}{\bibinfo{person}{{Energy Policy Institute at the University of Chicago (EPIC)}}.} \bibinfo{year}{2024}\natexlab{}.
\newblock \bibinfo{title}{{About the Air Quality Life Index (AQLI)}}.
\newblock
\newblock
\urldef\tempurl%
\url{https://aqli.epic.uchicago.edu/about/}
\showURL{%
\tempurl}
\newblock
\shownote{Accessed: 2024-11-26}.


\bibitem[Esri(2024)]%
        {arcgis}
\bibfield{author}{\bibinfo{person}{Esri}.} \bibinfo{year}{2024}\natexlab{}.
\newblock \bibinfo{title}{ArcGIS Geographic Information System}.
\newblock
\newblock
\urldef\tempurl%
\url{https://www.esri.com/en-us/arcgis/about-arcgis/overview}
\showURL{%
\tempurl}
\newblock
\shownote{Accessed: 2024-11-27}.


\bibitem[Guttikunda(2008)]%
        {guttikunda2008estimating}
\bibfield{author}{\bibinfo{person}{Sarath Guttikunda}.} \bibinfo{year}{2008}\natexlab{}.
\newblock \showarticletitle{Estimating health impacts of urban air pollution}.
\newblock In \bibinfo{booktitle}{\emph{SIM-air Working Paper Series}}. Vol.~\bibinfo{volume}{6}. \bibinfo{publisher}{New Delhi, India}, \bibinfo{pages}{2008}.
\newblock


\bibitem[Guttikunda(2009)]%
        {guttikunda2009impact}
\bibfield{author}{\bibinfo{person}{Sarath Guttikunda}.} \bibinfo{year}{2009}\natexlab{}.
\newblock \showarticletitle{Impact analysis of brick kilns on the air quality in Dhaka, Bangladesh}.
\newblock \bibinfo{journal}{\emph{SIM-air working paper series}} (\bibinfo{year}{2009}), \bibinfo{pages}{234}.
\newblock


\bibitem[Guttikunda et~al\mbox{.}(2023)]%
        {GUTTIKUNDA2023119712}
\bibfield{author}{\bibinfo{person}{Sarath Guttikunda}, \bibinfo{person}{Nishadh Ka}, \bibinfo{person}{Tanushree Ganguly}, {and} \bibinfo{person}{Puja Jawahar}.} \bibinfo{year}{2023}\natexlab{}.
\newblock \showarticletitle{Plugging the ambient air monitoring gaps in India's national clean air programme (NCAP) airsheds}.
\newblock \bibinfo{journal}{\emph{Atmospheric Environment}}  \bibinfo{volume}{301} (\bibinfo{year}{2023}), \bibinfo{pages}{119712}.
\newblock
\showISSN{1352-2310}
\urldef\tempurl%
\url{https://doi.org/10.1016/j.atmosenv.2023.119712}
\showDOI{\tempurl}


\bibitem[Guttikunda et~al\mbox{.}(2013)]%
        {guttikunda2013particulate}
\bibfield{author}{\bibinfo{person}{Sarath~K Guttikunda}, \bibinfo{person}{Bilkis~A Begum}, {and} \bibinfo{person}{Zia Wadud}.} \bibinfo{year}{2013}\natexlab{}.
\newblock \showarticletitle{Particulate pollution from brick kiln clusters in the Greater Dhaka region, Bangladesh}.
\newblock \bibinfo{journal}{\emph{Air Quality, Atmosphere \& Health}}  \bibinfo{volume}{6} (\bibinfo{year}{2013}), \bibinfo{pages}{357--365}.
\newblock


\bibitem[Guttikunda and Calori(2013)]%
        {guttikunda2013gis}
\bibfield{author}{\bibinfo{person}{Sarath~K Guttikunda} {and} \bibinfo{person}{Giuseppe Calori}.} \bibinfo{year}{2013}\natexlab{}.
\newblock \showarticletitle{A GIS based emissions inventory at 1 km$\times$ 1 km spatial resolution for air pollution analysis in Delhi, India}.
\newblock \bibinfo{journal}{\emph{Atmospheric Environment}}  \bibinfo{volume}{67} (\bibinfo{year}{2013}), \bibinfo{pages}{101--111}.
\newblock


\bibitem[Guttikunda and Goel(2013)]%
        {guttikunda2013health}
\bibfield{author}{\bibinfo{person}{Sarath~K Guttikunda} {and} \bibinfo{person}{Rahul Goel}.} \bibinfo{year}{2013}\natexlab{}.
\newblock \showarticletitle{Health impacts of particulate pollution in a megacity—Delhi, India}.
\newblock \bibinfo{journal}{\emph{Environmental Development}}  \bibinfo{volume}{6} (\bibinfo{year}{2013}), \bibinfo{pages}{8--20}.
\newblock


\bibitem[Guttikunda et~al\mbox{.}(2014)]%
        {guttikunda2014nature}
\bibfield{author}{\bibinfo{person}{Sarath~K Guttikunda}, \bibinfo{person}{Rahul Goel}, {and} \bibinfo{person}{Pallavi Pant}.} \bibinfo{year}{2014}\natexlab{}.
\newblock \showarticletitle{Nature of air pollution, emission sources, and management in the Indian cities}.
\newblock \bibinfo{journal}{\emph{Atmospheric environment}}  \bibinfo{volume}{95} (\bibinfo{year}{2014}), \bibinfo{pages}{501--510}.
\newblock


\bibitem[Guttikunda and Jawahar(2014)]%
        {guttikunda2014atmospheric}
\bibfield{author}{\bibinfo{person}{Sarath~K Guttikunda} {and} \bibinfo{person}{Puja Jawahar}.} \bibinfo{year}{2014}\natexlab{}.
\newblock \showarticletitle{Atmospheric emissions and pollution from the coal-fired thermal power plants in India}.
\newblock \bibinfo{journal}{\emph{Atmospheric Environment}}  \bibinfo{volume}{92} (\bibinfo{year}{2014}), \bibinfo{pages}{449--460}.
\newblock


\bibitem[Guttikunda et~al\mbox{.}(2019b)]%
        {guttikunda2019air}
\bibfield{author}{\bibinfo{person}{Sarath~K Guttikunda}, \bibinfo{person}{KA Nishadh}, \bibinfo{person}{Sudhir Gota}, \bibinfo{person}{Pratima Singh}, \bibinfo{person}{Arijit Chanda}, \bibinfo{person}{Puja Jawahar}, {and} \bibinfo{person}{Jai Asundi}.} \bibinfo{year}{2019}\natexlab{b}.
\newblock \showarticletitle{Air quality, emissions, and source contributions analysis for the Greater Bengaluru region of India}.
\newblock \bibinfo{journal}{\emph{Atmospheric Pollution Research}} \bibinfo{volume}{10}, \bibinfo{number}{3} (\bibinfo{year}{2019}), \bibinfo{pages}{941--953}.
\newblock


\bibitem[Guttikunda et~al\mbox{.}(2019a)]%
        {GUTTIKUNDA2019124}
\bibfield{author}{\bibinfo{person}{Sarath~K. Guttikunda}, \bibinfo{person}{K.A. Nishadh}, {and} \bibinfo{person}{Puja Jawahar}.} \bibinfo{year}{2019}\natexlab{a}.
\newblock \showarticletitle{Air pollution knowledge assessments (APnA) for 20 Indian cities}.
\newblock \bibinfo{journal}{\emph{Urban Climate}}  \bibinfo{volume}{27} (\bibinfo{year}{2019}), \bibinfo{pages}{124--141}.
\newblock
\showISSN{2212-0955}
\urldef\tempurl%
\url{https://doi.org/10.1016/j.uclim.2018.11.005}
\showDOI{\tempurl}


\bibitem[Hamdani et~al\mbox{.}(2024)]%
        {hamdani2024brick}
\bibfield{author}{\bibinfo{person}{Muhammad Suleman~Ali Hamdani}, \bibinfo{person}{Khizer Zakir}, \bibinfo{person}{Neetu Kushwaha}, \bibinfo{person}{Syeda~Eman Fatima}, {and} \bibinfo{person}{Hassan~Aftab Sheikh}.} \bibinfo{year}{2024}\natexlab{}.
\newblock \showarticletitle{Brick Kiln Dataset for Pakistan's IGP Region Using AI}.
\newblock \bibinfo{journal}{\emph{arXiv preprint arXiv:2412.00052}} (\bibinfo{year}{2024}).
\newblock


\bibitem[Han et~al\mbox{.}(2021)]%
        {han2021redet}
\bibfield{author}{\bibinfo{person}{Jiaming Han}, \bibinfo{person}{Jian Ding}, \bibinfo{person}{Nan Xue}, {and} \bibinfo{person}{Gui-Song Xia}.} \bibinfo{year}{2021}\natexlab{}.
\newblock \showarticletitle{Redet: A rotation-equivariant detector for aerial object detection}. In \bibinfo{booktitle}{\emph{Proceedings of the IEEE Conference on Computer Vision and Pattern Recognition}}. \bibinfo{pages}{2786--2795}.
\newblock


\bibitem[Haque et~al\mbox{.}(2022)]%
        {haque2022impact}
\bibfield{author}{\bibinfo{person}{Shama~E Haque}, \bibinfo{person}{Minhaz~M Shahriar}, \bibinfo{person}{Nazmun Nahar}, {and} \bibinfo{person}{Md~Sazzadul Haque}.} \bibinfo{year}{2022}\natexlab{}.
\newblock \showarticletitle{Impact of brick kiln emissions on soil quality: A case study of Ashulia brick kiln cluster, Bangladesh}.
\newblock \bibinfo{journal}{\emph{Environmental Challenges}}  \bibinfo{volume}{9} (\bibinfo{year}{2022}), \bibinfo{pages}{100640}.
\newblock


\bibitem[Hu et~al\mbox{.}(2023)]%
        {hu2023graph}
\bibfield{author}{\bibinfo{person}{Junfeng Hu}, \bibinfo{person}{Yuxuan Liang}, \bibinfo{person}{Zhencheng Fan}, \bibinfo{person}{Hongyang Chen}, \bibinfo{person}{Yu Zheng}, {and} \bibinfo{person}{Roger Zimmermann}.} \bibinfo{year}{2023}\natexlab{}.
\newblock \showarticletitle{Graph Neural Processes for Spatio-Temporal Extrapolation}. In \bibinfo{booktitle}{\emph{Proceedings of the 29th ACM SIGKDD Conference on Knowledge Discovery and Data Mining}}.
\newblock


\bibitem[Imaduddin et~al\mbox{.}(2023)]%
        {imaduddin2023detection}
\bibfield{author}{\bibinfo{person}{Syed Imaduddin}, \bibinfo{person}{Yusuf~Ahmed Khan}, \bibinfo{person}{Khushboo Mirza}, {and} \bibinfo{person}{BK Bhadra}.} \bibinfo{year}{2023}\natexlab{}.
\newblock \showarticletitle{Detection of Brick Kilns Using Multi-Spectral Bands of Sentinel-2 Imagery}. In \bibinfo{booktitle}{\emph{2023 International Conference on Artificial Intelligence and Smart Communication (AISC)}}. IEEE, \bibinfo{pages}{496--503}.
\newblock


\bibitem[{IQAir}({[n.\,d.]})]%
        {iqair2024}
\bibfield{author}{\bibinfo{person}{{IQAir}}.} \bibinfo{year}{[n.\,d.]}\natexlab{}.
\newblock \bibinfo{title}{{World's Most Polluted Cities - 2024}}.
\newblock \bibinfo{howpublished}{\url{https://www.iqair.com/in-en/world-most-polluted-cities}}.
\newblock
\newblock
\shownote{Accessed: 2024-11-26}.


\bibitem[Jocher et~al\mbox{.}(2023)]%
        {ultralytics_yolo_2023}
\bibfield{author}{\bibinfo{person}{Glenn Jocher}, \bibinfo{person}{Jing Qiu}, {and} \bibinfo{person}{Ayush Chaurasia}.} \bibinfo{year}{2023}\natexlab{}.
\newblock \bibinfo{booktitle}{\emph{Ultralytics YOLO}}.
\newblock Ultralytics.
\newblock
\urldef\tempurl%
\url{https://ultralytics.com}
\showURL{%
\tempurl}


\bibitem[Kanabkaew and Buasing(2015)]%
        {kanabkaew2015assessment}
\bibfield{author}{\bibinfo{person}{T Kanabkaew} {and} \bibinfo{person}{K Buasing}.} \bibinfo{year}{2015}\natexlab{}.
\newblock \showarticletitle{Assessment of air pollution concentrations from brick kilns using an atmospheric dispersion model}.
\newblock \bibinfo{journal}{\emph{WIT Trans. Ecol. Environ}}  \bibinfo{volume}{198} (\bibinfo{year}{2015}), \bibinfo{pages}{27--37}.
\newblock


\bibitem[Labs({[n.\,d.]})]%
        {planetEducationResearch}
\bibfield{author}{\bibinfo{person}{Planet Labs}.} \bibinfo{year}{[n.\,d.]}\natexlab{}.
\newblock \bibinfo{title}{Education and Research Program}.
\newblock \bibinfo{howpublished}{\url{https://www.planet.com/industries/education-and-research/}}.
\newblock
\newblock
\shownote{[Accessed 08-11-2024]}.


\bibitem[Le and Oanh(2010)]%
        {le2010integrated}
\bibfield{author}{\bibinfo{person}{Hoang~Anh Le} {and} \bibinfo{person}{Nguyen Thi~Kim Oanh}.} \bibinfo{year}{2010}\natexlab{}.
\newblock \showarticletitle{Integrated assessment of brick kiln emission impacts on air quality}.
\newblock \bibinfo{journal}{\emph{Environmental monitoring and assessment}}  \bibinfo{volume}{171} (\bibinfo{year}{2010}), \bibinfo{pages}{381--394}.
\newblock


\bibitem[Lebakula et~al\mbox{.}(2024)]%
        {lebakula2024landscan}
\bibfield{author}{\bibinfo{person}{V. Lebakula}, \bibinfo{person}{J. Epting}, \bibinfo{person}{J. Moehl}, \bibinfo{person}{C. Stipek}, \bibinfo{person}{D. Adams}, \bibinfo{person}{A. Reith}, \bibinfo{person}{J. Kaufman}, \bibinfo{person}{J. Gonzales}, \bibinfo{person}{B. Reynolds}, \bibinfo{person}{S. Basford}, \bibinfo{person}{A. Martin}, \bibinfo{person}{W. Buck}, \bibinfo{person}{A. Faxon}, \bibinfo{person}{A. Cunningham}, \bibinfo{person}{A. Roy}, \bibinfo{person}{Z. Barbose}, \bibinfo{person}{J. Massaro}, \bibinfo{person}{S. Walters}, \bibinfo{person}{C. Woody}, {and} \bibinfo{person}{M. Urban}.} \bibinfo{year}{2024}\natexlab{}.
\newblock \bibinfo{title}{LandScan Silver Edition [Data set]}.
\newblock
\newblock
\urldef\tempurl%
\url{https://doi.org/10.48690/1531770}
\showDOI{\tempurl}


\bibitem[Lee et~al\mbox{.}(2021)]%
        {lee2021scalable}
\bibfield{author}{\bibinfo{person}{Jihyeon Lee}, \bibinfo{person}{Nina~R Brooks}, \bibinfo{person}{Fahim Tajwar}, \bibinfo{person}{Marshall Burke}, \bibinfo{person}{Stefano Ermon}, \bibinfo{person}{David~B Lobell}, \bibinfo{person}{Debashish Biswas}, {and} \bibinfo{person}{Stephen~P Luby}.} \bibinfo{year}{2021}\natexlab{}.
\newblock \showarticletitle{Scalable deep learning to identify brick kilns and aid regulatory capacity}.
\newblock \bibinfo{journal}{\emph{Proceedings of the National Academy of Sciences}} \bibinfo{volume}{118}, \bibinfo{number}{17} (\bibinfo{year}{2021}), \bibinfo{pages}{e2018863118}.
\newblock


\bibitem[Li et~al\mbox{.}(2019)]%
        {li2019aging}
\bibfield{author}{\bibinfo{person}{Xiaodong Li}, \bibinfo{person}{Giles~M Foody}, \bibinfo{person}{Doreen~S Boyd}, {and} \bibinfo{person}{Feng Ling}.} \bibinfo{year}{2019}\natexlab{}.
\newblock \showarticletitle{Aging brick kilns in the asian brick belt using a long time series of Landsat sensor data to inform the study of modern day slavery}. In \bibinfo{booktitle}{\emph{IGARSS 2019-2019 IEEE International Geoscience and Remote Sensing Symposium}}. IEEE, \bibinfo{pages}{130--133}.
\newblock


\bibitem[Liang et~al\mbox{.}(2023)]%
        {liang2023airformer}
\bibfield{author}{\bibinfo{person}{Yuxuan Liang}, \bibinfo{person}{Yutong Xia}, \bibinfo{person}{Songyu Ke}, \bibinfo{person}{Yiwei Wang}, \bibinfo{person}{Qingsong Wen}, \bibinfo{person}{Junbo Zhang}, \bibinfo{person}{Yu Zheng}, {and} \bibinfo{person}{Roger Zimmermann}.} \bibinfo{year}{2023}\natexlab{}.
\newblock \showarticletitle{Airformer: Predicting nationwide air quality in china with transformers}. In \bibinfo{booktitle}{\emph{Proceedings of the AAAI conference on artificial intelligence}}, Vol.~\bibinfo{volume}{37}. \bibinfo{pages}{14329--14337}.
\newblock


\bibitem[Lin(2019)]%
        {Lin2019BinarySA}
\bibfield{author}{\bibinfo{person}{Anthony Lin}.} \bibinfo{year}{2019}\natexlab{}.
\newblock \showarticletitle{Binary search algorithm}.
\newblock \bibinfo{journal}{\emph{WikiJournal of Science}} (\bibinfo{year}{2019}).
\newblock
\urldef\tempurl%
\url{https://api.semanticscholar.org/CorpusID:62128693}
\showURL{%
\tempurl}


\bibitem[Lin et~al\mbox{.}(2014)]%
        {lin2014microsoft}
\bibfield{author}{\bibinfo{person}{Tsung-Yi Lin}, \bibinfo{person}{Michael Maire}, \bibinfo{person}{Serge Belongie}, \bibinfo{person}{James Hays}, \bibinfo{person}{Pietro Perona}, \bibinfo{person}{Deva Ramanan}, \bibinfo{person}{Piotr Doll{\'a}r}, {and} \bibinfo{person}{C~Lawrence Zitnick}.} \bibinfo{year}{2014}\natexlab{}.
\newblock \showarticletitle{Microsoft coco: Common objects in context}. In \bibinfo{booktitle}{\emph{Computer Vision--ECCV 2014: 13th European Conference, Zurich, Switzerland, September 6-12, 2014, Proceedings, Part V 13}}. Springer, \bibinfo{pages}{740--755}.
\newblock


\bibitem[Liu et~al\mbox{.}(2017)]%
        {liu2017high}
\bibfield{author}{\bibinfo{person}{Zikun Liu}, \bibinfo{person}{Liu Yuan}, \bibinfo{person}{Lubin Weng}, {and} \bibinfo{person}{Yiping Yang}.} \bibinfo{year}{2017}\natexlab{}.
\newblock \showarticletitle{A high resolution optical satellite image dataset for ship recognition and some new baselines}. In \bibinfo{booktitle}{\emph{International conference on pattern recognition applications and methods}}, Vol.~\bibinfo{volume}{2}. SciTePress, \bibinfo{pages}{324--331}.
\newblock


\bibitem[Ltd(2012)]%
        {Greentech2012}
\bibfield{author}{\bibinfo{person}{Greentech Knowledge Solutions~Pvt Ltd}.} \bibinfo{year}{2012}\natexlab{}.
\newblock \bibinfo{title}{Evaluating Energy Conservation Potential of Brick Production in India}.
\newblock
\newblock
\newblock
\shownote{Prepared for SAARC Energy Centre}.


\bibitem[Meier(2023)]%
        {meier2023advancing}
\bibfield{author}{\bibinfo{person}{Sandro Meier}.} \bibinfo{year}{2023}\natexlab{}.
\newblock \bibinfo{title}{Advancing the quantification of CO2 and NOx emissions from power plants using Sentinel-5P observations: A novel approach to address NOx chemistry in plumes}.
\newblock
\newblock


\bibitem[Misra et~al\mbox{.}(2020)]%
        {misra2020mapping}
\bibfield{author}{\bibinfo{person}{Prakhar Misra}, \bibinfo{person}{Ryoichi Imasu}, \bibinfo{person}{Sachiko Hayashida}, \bibinfo{person}{Ardhi~Adhary Arbain}, \bibinfo{person}{Ram Avtar}, {and} \bibinfo{person}{Wataru Takeuchi}.} \bibinfo{year}{2020}\natexlab{}.
\newblock \showarticletitle{Mapping brick kilns to support environmental impact studies around Delhi using Sentinel-2}.
\newblock \bibinfo{journal}{\emph{ISPRS International Journal of Geo-Information}} \bibinfo{volume}{9}, \bibinfo{number}{9} (\bibinfo{year}{2020}), \bibinfo{pages}{544}.
\newblock


\bibitem[Mondal et~al\mbox{.}(2024)]%
        {mondal2024scalable}
\bibfield{author}{\bibinfo{person}{Rishabh Mondal}, \bibinfo{person}{Zeel~B Patel}, \bibinfo{person}{Vannsh Jani}, {and} \bibinfo{person}{Nipun Batra}.} \bibinfo{year}{2024}\natexlab{}.
\newblock \showarticletitle{Scalable Methods for Brick Kiln Detection and Compliance Monitoring from Satellite Imagery: A Deployment Case Study in India}.
\newblock \bibinfo{journal}{\emph{arXiv preprint arXiv:2402.13796}} (\bibinfo{year}{2024}).
\newblock


\bibitem[Nazir et~al\mbox{.}(2019)]%
        {nazir2019tiny}
\bibfield{author}{\bibinfo{person}{Usman Nazir}, \bibinfo{person}{Numan Khurshid}, \bibinfo{person}{Muhammad Ahmed~Bhimra}, {and} \bibinfo{person}{Murtaza Taj}.} \bibinfo{year}{2019}\natexlab{}.
\newblock \showarticletitle{Tiny-Inception-ResNet-v2: Using deep learning for eliminating bonded labors of brick kilns in South Asia}. In \bibinfo{booktitle}{\emph{Proceedings of the IEEE/CVF Conference on Computer Vision and Pattern Recognition Workshops}}. \bibinfo{pages}{39--43}.
\newblock


\bibitem[Nazir et~al\mbox{.}(2020)]%
        {nazir2020kiln}
\bibfield{author}{\bibinfo{person}{Usman Nazir}, \bibinfo{person}{Usman~Khalid Mian}, \bibinfo{person}{Muhammad~Usman Sohail}, \bibinfo{person}{Murtaza Taj}, {and} \bibinfo{person}{Momin Uppal}.} \bibinfo{year}{2020}\natexlab{}.
\newblock \showarticletitle{Kiln-net: A gated neural network for detection of brick kilns in South Asia}.
\newblock \bibinfo{journal}{\emph{IEEE Journal of Selected Topics in Applied Earth Observations and Remote Sensing}}  \bibinfo{volume}{13} (\bibinfo{year}{2020}), \bibinfo{pages}{3251--3262}.
\newblock


\bibitem[Nicolaou et~al\mbox{.}(2024)]%
        {nicolaou2024brick}
\bibfield{author}{\bibinfo{person}{Laura Nicolaou}, \bibinfo{person}{Fiona Sylvies}, \bibinfo{person}{Isabel Veloso}, \bibinfo{person}{Katherine Lord}, \bibinfo{person}{Ram~K Chandyo}, \bibinfo{person}{Arun~K Sharma}, \bibinfo{person}{Laxman~P Shrestha}, \bibinfo{person}{David~L Parker}, \bibinfo{person}{Steven~M Thygerson}, \bibinfo{person}{Peter~F DeCarlo}, {et~al\mbox{.}}} \bibinfo{year}{2024}\natexlab{}.
\newblock \showarticletitle{Brick kiln pollution and its impact on health: A systematic review and meta-analysis}.
\newblock \bibinfo{journal}{\emph{Environmental research}} (\bibinfo{year}{2024}), \bibinfo{pages}{119220}.
\newblock


\bibitem[{OpenStreetMap contributors}(2024a)]%
        {openstreetmap}
\bibfield{author}{\bibinfo{person}{{OpenStreetMap contributors}}.} \bibinfo{year}{2024}\natexlab{a}.
\newblock \bibinfo{title}{OpenStreetMap}.
\newblock
\newblock
\urldef\tempurl%
\url{https://www.openstreetmap.org}
\showURL{%
\tempurl}
\newblock
\shownote{Accessed: 2024-11-27}.


\bibitem[{OpenStreetMap contributors}(2024b)]%
        {geofabrik2024}
\bibfield{author}{\bibinfo{person}{{OpenStreetMap contributors}}.} \bibinfo{year}{2024}\natexlab{b}.
\newblock \bibinfo{title}{{OpenStreetMap Data Downloaded from Geofabrik}}.
\newblock
\newblock
\urldef\tempurl%
\url{https://download.geofabrik.de/}
\showURL{%
\tempurl}
\newblock
\shownote{Accessed from Geofabrik [https://download.geofabrik.de/] on [date].}.


\bibitem[Organisation. and Programme.(1984)]%
        {ILO1984}
\bibfield{author}{\bibinfo{person}{International~Labour Organisation.} {and} \bibinfo{person}{United Nations~Development Programme.}} \bibinfo{year}{1984}\natexlab{}.
\newblock \bibinfo{booktitle}{\emph{Small-scale brickmaking.}}
\newblock \bibinfo{publisher}{ILO}, \bibinfo{address}{Geneva}.
\newblock
\showISBNx{9221035670}


\bibitem[Patel et~al\mbox{.}(2022)]%
        {Patel_Purohit_Patel_Sahni_Batra_2022}
\bibfield{author}{\bibinfo{person}{Zeel~B Patel}, \bibinfo{person}{Palak Purohit}, \bibinfo{person}{Harsh~M Patel}, \bibinfo{person}{Shivam Sahni}, {and} \bibinfo{person}{Nipun Batra}.} \bibinfo{year}{2022}\natexlab{}.
\newblock \showarticletitle{Accurate and Scalable Gaussian Processes for Fine-Grained Air Quality Inference}.
\newblock \bibinfo{journal}{\emph{Proceedings of the AAAI Conference on Artificial Intelligence}} \bibinfo{volume}{36}, \bibinfo{number}{11} (\bibinfo{date}{Jun.} \bibinfo{year}{2022}), \bibinfo{pages}{12080--12088}.
\newblock
\urldef\tempurl%
\url{https://doi.org/10.1609/aaai.v36i11.21467}
\showDOI{\tempurl}


\bibitem[Paul et~al\mbox{.}(2022)]%
        {paul2022brick}
\bibfield{author}{\bibinfo{person}{Arati Paul}, \bibinfo{person}{Soumya Bandyopadhyay}, {and} \bibinfo{person}{Uday Raj}.} \bibinfo{year}{2022}\natexlab{}.
\newblock \showarticletitle{Brick kiln detection in remote sensing imagery using deep neural network and change analysis}.
\newblock \bibinfo{journal}{\emph{Spatial Information Research}} \bibinfo{volume}{30}, \bibinfo{number}{5} (\bibinfo{year}{2022}), \bibinfo{pages}{607--616}.
\newblock


\bibitem[Programme(2023)]%
        {UNDP2023}
\bibfield{author}{\bibinfo{person}{United Nations~Development Programme}.} \bibinfo{year}{2023}\natexlab{}.
\newblock \bibinfo{title}{GeoAI for Brick Kilns in Bihar: Learnings and Recommendations}.
\newblock
\newblock
\urldef\tempurl%
\url{https://www.undp.org/sites/g/files/zskgke326/files/2023-12/geoai_for_brick_kilns_v8_web_pages_002_0.pdf}
\showURL{%
\tempurl}
\newblock
\shownote{Accessed: 31 March 2025}.


\bibitem[Rajarathnam et~al\mbox{.}(2014)]%
        {rajarathnam2014}
\bibfield{author}{\bibinfo{person}{Uma Rajarathnam}, \bibinfo{person}{Vasudev Athalye}, \bibinfo{person}{Santhosh Ragavan}, \bibinfo{person}{Sameer Maithel}, \bibinfo{person}{Dheeraj Lalchandani}, \bibinfo{person}{Sonal Kumar}, \bibinfo{person}{Ellen Baum}, \bibinfo{person}{Cheryl Weyant}, {and} \bibinfo{person}{Tami Bond}.} \bibinfo{year}{2014}\natexlab{}.
\newblock \showarticletitle{Assessment of air pollutant emissions from brick kilns}.
\newblock \bibinfo{journal}{\emph{Atmospheric Environment}}  \bibinfo{volume}{98} (\bibinfo{year}{2014}), \bibinfo{pages}{549--553}.
\newblock
\showISSN{1352-2310}
\urldef\tempurl%
\url{https://doi.org/10.1016/j.atmosenv.2014.08.075}
\showDOI{\tempurl}


\bibitem[Raza and Ali(2021)]%
        {raza2021impact}
\bibfield{author}{\bibinfo{person}{Ali Raza} {and} \bibinfo{person}{Zulfiqar Ali}.} \bibinfo{year}{2021}\natexlab{}.
\newblock \showarticletitle{Impact of air pollution generated by brick kilns on the pulmonary health of workers}.
\newblock \bibinfo{journal}{\emph{Journal of Health and Pollution}} \bibinfo{volume}{11}, \bibinfo{number}{31} (\bibinfo{year}{2021}), \bibinfo{pages}{210906}.
\newblock


\bibitem[Redmon et~al\mbox{.}(2016)]%
        {redmon2016you}
\bibfield{author}{\bibinfo{person}{Joseph Redmon}, \bibinfo{person}{Santosh Divvala}, \bibinfo{person}{Ross Girshick}, {and} \bibinfo{person}{Ali Farhadi}.} \bibinfo{year}{2016}\natexlab{}.
\newblock \showarticletitle{You only look once: Unified, real-time object detection}. In \bibinfo{booktitle}{\emph{Proceedings of the IEEE conference on computer vision and pattern recognition}}. \bibinfo{pages}{779--788}.
\newblock


\bibitem[Skamarock et~al\mbox{.}(2019a)]%
        {Skamarock2019}
\bibfield{author}{\bibinfo{person}{W.~C. Skamarock}, \bibinfo{person}{J.~B. Klemp}, \bibinfo{person}{J. Dudhia}, \bibinfo{person}{D.~O. Gill}, \bibinfo{person}{Z. Liu}, \bibinfo{person}{J. Berner}, \bibinfo{person}{W. Wang}, \bibinfo{person}{J.~G. Powers}, \bibinfo{person}{M.~G. Duda}, \bibinfo{person}{D.~M. Barker}, {and} \bibinfo{person}{X.-Y. Huang}.} \bibinfo{year}{2019}\natexlab{a}.
\newblock \bibinfo{booktitle}{\emph{A Description of the Advanced Research WRF Version 4}}.
\newblock \bibinfo{type}{NCAR Technical Note} NCAR/TN-556+STR. \bibinfo{institution}{National Center for Atmospheric Research (NCAR)}. \bibinfo{pages}{145} pages.
\newblock
\urldef\tempurl%
\url{https://doi.org/10.5065/1dfh-6p97}
\showDOI{\tempurl}


\bibitem[Skamarock et~al\mbox{.}(2019b)]%
        {Skamarock2019WRF}
\bibfield{author}{\bibinfo{person}{W.~C. Skamarock}, \bibinfo{person}{J.~B. Klemp}, \bibinfo{person}{J. Dudhia}, \bibinfo{person}{D.~O. Gill}, \bibinfo{person}{Z. Liu}, \bibinfo{person}{J. Berner}, \bibinfo{person}{W. Wang}, \bibinfo{person}{J.~G. Powers}, \bibinfo{person}{M.~G. Duda}, \bibinfo{person}{D.~M. Barker}, {and} \bibinfo{person}{X.-Y. Huang}.} \bibinfo{year}{2019}\natexlab{b}.
\newblock \bibinfo{booktitle}{\emph{A Description of the Advanced Research WRF Version 4}}.
\newblock \bibinfo{type}{NCAR Tech. Note} NCAR/TN-556+STR. \bibinfo{institution}{NCAR}. \bibinfo{pages}{145} pages.
\newblock
\urldef\tempurl%
\url{https://doi.org/10.5065/1dfh-6p97}
\showDOI{\tempurl}


\bibitem[Skinder et~al\mbox{.}(2014)]%
        {skinder2014brick}
\bibfield{author}{\bibinfo{person}{BM Skinder}, \bibinfo{person}{AK Pandit}, \bibinfo{person}{AQ Sheikh}, {and} \bibinfo{person}{BA Ganai}.} \bibinfo{year}{2014}\natexlab{}.
\newblock \showarticletitle{Brick kilns: cause of atmospheric pollution}.
\newblock \bibinfo{journal}{\emph{J Pollut Eff Cont}} \bibinfo{volume}{2}, \bibinfo{number}{112} (\bibinfo{year}{2014}), \bibinfo{pages}{3}.
\newblock


\bibitem[Szegedy et~al\mbox{.}(2017)]%
        {szegedy2017inception}
\bibfield{author}{\bibinfo{person}{Christian Szegedy}, \bibinfo{person}{Sergey Ioffe}, \bibinfo{person}{Vincent Vanhoucke}, {and} \bibinfo{person}{Alexander Alemi}.} \bibinfo{year}{2017}\natexlab{}.
\newblock \showarticletitle{Inception-v4, inception-resnet and the impact of residual connections on learning}. In \bibinfo{booktitle}{\emph{Proceedings of the AAAI conference on artificial intelligence}}, Vol.~\bibinfo{volume}{31}.
\newblock


\bibitem[Tahir et~al\mbox{.}(2021)]%
        {tahir2021brick}
\bibfield{author}{\bibinfo{person}{Rosheen Tahir}, \bibinfo{person}{Muhammad~Shaaf Imran}, \bibinfo{person}{Sidra Minhas}, \bibinfo{person}{Nosheen Sabahat}, \bibinfo{person}{Sardar Haider~Waseem Ilyas}, {and} \bibinfo{person}{Haider~Raza Gadi}.} \bibinfo{year}{2021}\natexlab{}.
\newblock \showarticletitle{Brick kiln detection and localization using deep learning techniques}. In \bibinfo{booktitle}{\emph{2021 International Conference on Artificial Intelligence (ICAI)}}. IEEE, \bibinfo{pages}{37--43}.
\newblock


\bibitem[Thangavel et~al\mbox{.}(2022)]%
        {thangavel2022recent}
\bibfield{author}{\bibinfo{person}{Prakash Thangavel}, \bibinfo{person}{Duckshin Park}, {and} \bibinfo{person}{Young-Chul Lee}.} \bibinfo{year}{2022}\natexlab{}.
\newblock \showarticletitle{Recent insights into particulate matter (PM2. 5)-mediated toxicity in humans: an overview}.
\newblock \bibinfo{journal}{\emph{International journal of environmental research and public health}} \bibinfo{volume}{19}, \bibinfo{number}{12} (\bibinfo{year}{2022}), \bibinfo{pages}{7511}.
\newblock


\bibitem[Tibrewal et~al\mbox{.}(2023)]%
        {tibrewal2023reconciliation}
\bibfield{author}{\bibinfo{person}{Kushal Tibrewal}, \bibinfo{person}{Chandra Venkataraman}, \bibinfo{person}{Harish Phuleria}, \bibinfo{person}{Veena Joshi}, \bibinfo{person}{Sameer Maithel}, \bibinfo{person}{Anand Damle}, \bibinfo{person}{Anurag Gupta}, \bibinfo{person}{Pradnya Lokhande}, \bibinfo{person}{Shahadev Rabha}, \bibinfo{person}{Binoy~K Saikia}, {et~al\mbox{.}}} \bibinfo{year}{2023}\natexlab{}.
\newblock \showarticletitle{Reconciliation of energy use disparities in brick production in India}.
\newblock \bibinfo{journal}{\emph{Nature Sustainability}} \bibinfo{volume}{6}, \bibinfo{number}{10} (\bibinfo{year}{2023}), \bibinfo{pages}{1248--1257}.
\newblock


\bibitem[Trenchev et~al\mbox{.}(2023)]%
        {trenchev2023huge}
\bibfield{author}{\bibinfo{person}{Plamen Trenchev}, \bibinfo{person}{Maria Dimitrova}, {and} \bibinfo{person}{Daniela Avetisyan}.} \bibinfo{year}{2023}\natexlab{}.
\newblock \showarticletitle{Huge CH4, NO2 and CO Emissions from Coal Mines in the Kuznetsk Basin (Russia) Detected by Sentinel-5P}.
\newblock \bibinfo{journal}{\emph{Remote Sensing}} \bibinfo{volume}{15}, \bibinfo{number}{6} (\bibinfo{year}{2023}), \bibinfo{pages}{1590}.
\newblock


\bibitem[UNEP(2019)]%
        {unep19}
\bibfield{author}{\bibinfo{person}{UNEP}.} \bibinfo{year}{2019}\natexlab{}.
\newblock \bibinfo{title}{Emissions Gap Report 2019}.
\newblock , \bibinfo{numpages}{11}~pages.
\newblock


\bibitem[{Uttar Pradesh Government}(2012)]%
        {uttar_pradesh_brick_kiln_rules_2012}
\bibfield{author}{\bibinfo{person}{{Uttar Pradesh Government}}.} \bibinfo{year}{2012}\natexlab{}.
\newblock \bibinfo{title}{The Uttar Pradesh Brick Kilns (Siting Criteria for Establishment) Rules, 2012}.
\newblock
\newblock
\urldef\tempurl%
\url{https://upload.indiacode.nic.in/showfile?actid=AC_UP_88_464_00007_00007_1616154959565&type=rule&filename=final_bricklin_rules_pdf.pdf}
\showURL{%
\tempurl}
\newblock
\shownote{Accessed: April 1, 2025}.


\bibitem[{Uttar Pradesh Pollution Control Board (UPPCB)}({[n.\,d.]})]%
        {uppcbbrickkiln}
\bibfield{author}{\bibinfo{person}{{Uttar Pradesh Pollution Control Board (UPPCB)}}.} \bibinfo{year}{[n.\,d.]}\natexlab{}.
\newblock \bibinfo{title}{{Guidelines for brick kiln Industry}}.
\newblock \bibinfo{howpublished}{\url{http://uppcb.com/pdf/brick_280519.pdf}}.
\newblock
\newblock
\shownote{Accessed: 2024-11-26}.


\bibitem[{Uttar Pradesh Pollution Control Board (UPPCB)}(2023)]%
        {UPPCB2023}
\bibfield{author}{\bibinfo{person}{{Uttar Pradesh Pollution Control Board (UPPCB)}}.} \bibinfo{year}{2023}\natexlab{}.
\newblock \bibinfo{title}{UPPCB Report on Brick Kilns - March 2023}.
\newblock
\newblock
\urldef\tempurl%
\url{http://www.indiaenvironmentportal.org.in/files/file/UPPCB-report-brick-kilns-March-2023.pdf}
\showURL{%
\tempurl}
\newblock
\shownote{Accessed: 2024-11-04}.


\bibitem[vs~West Bengal Pollution Control~Board(2022)]%
        {bengal_brick_field_2022}
\bibfield{author}{\bibinfo{person}{Bengal Brick Field Owners~Association vs West Bengal Pollution Control~Board}.} \bibinfo{year}{2022}\natexlab{}.
\newblock \bibinfo{title}{Bengal Brick Field Owners Association vs. West Bengal Pollution Control Board}.
\newblock
\newblock
\urldef\tempurl%
\url{https://indiankanoon.org/doc/104363701/}
\showURL{%
\tempurl}
\newblock
\shownote{Accessed: 31 March 2025}.


\bibitem[Wang et~al\mbox{.}(2025)]%
        {wang2025airradar}
\bibfield{author}{\bibinfo{person}{Qiongyan Wang}, \bibinfo{person}{Yutong Xia}, \bibinfo{person}{Siru ZHong}, \bibinfo{person}{Weichuang Li}, \bibinfo{person}{Yuankai Wu}, \bibinfo{person}{Shifen Cheng}, \bibinfo{person}{Junbo Zhang}, \bibinfo{person}{Yu Zheng}, {and} \bibinfo{person}{Yuxuan Liang}.} \bibinfo{year}{2025}\natexlab{}.
\newblock \showarticletitle{AirRadar: Inferring Nationwide Air Quality in China with Deep Neural Networks}.
\newblock \bibinfo{journal}{\emph{arXiv preprint arXiv:2501.13141}} (\bibinfo{year}{2025}).
\newblock


\bibitem[Weyant et~al\mbox{.}(2014)]%
        {weyant2014emissions}
\bibfield{author}{\bibinfo{person}{Cheryl Weyant}, \bibinfo{person}{Vasudev Athalye}, \bibinfo{person}{Santhosh Ragavan}, \bibinfo{person}{Uma Rajarathnam}, \bibinfo{person}{Dheeraj Lalchandani}, \bibinfo{person}{Sameer Maithel}, \bibinfo{person}{Ellen Baum}, {and} \bibinfo{person}{Tami~C Bond}.} \bibinfo{year}{2014}\natexlab{}.
\newblock \showarticletitle{Emissions from South Asian brick production}.
\newblock \bibinfo{journal}{\emph{Environmental science \& technology}} \bibinfo{volume}{48}, \bibinfo{number}{11} (\bibinfo{year}{2014}), \bibinfo{pages}{6477--6483}.
\newblock


\bibitem[{World Health Organization}(nd)]%
        {who_air_pollution}
\bibfield{author}{\bibinfo{person}{{World Health Organization}}.} \bibinfo{year}{n.d.}\natexlab{}.
\newblock \bibinfo{title}{Air Pollution in India}.
\newblock
\newblock
\urldef\tempurl%
\url{https://www.who.int/news-room/fact-sheets/detail/ambient-(outdoor)-air-quality-and-health}
\showURL{%
\tempurl}
\newblock
\shownote{Accessed: 2024-11-29}.


\bibitem[WorldBank(2020)]%
        {worldbank2020}
\bibfield{author}{\bibinfo{person}{WorldBank}.} \bibinfo{year}{2020}\natexlab{}.
\newblock \bibinfo{title}{DIRTY STACKS, HIGH STAKES: An Overview of Brick Sector in South Asia}.
\newblock
\newblock


\bibitem[Wu(2021)]%
        {Wu2021}
\bibfield{author}{\bibinfo{person}{Qiusheng Wu}.} \bibinfo{year}{2021}\natexlab{}.
\newblock \showarticletitle{Leafmap: A Python package for interactive mapping and geospatial analysis with minimal coding in a Jupyter environment}.
\newblock \bibinfo{journal}{\emph{Journal of Open Source Software}} \bibinfo{volume}{6}, \bibinfo{number}{63} (\bibinfo{year}{2021}), \bibinfo{pages}{3414}.
\newblock
\urldef\tempurl%
\url{https://doi.org/10.21105/joss.03414}
\showDOI{\tempurl}


\bibitem[Xi et~al\mbox{.}(2023)]%
        {xi2023satellite}
\bibfield{author}{\bibinfo{person}{Yanxin Xi}, \bibinfo{person}{Yu Liu}, \bibinfo{person}{Tong Li}, \bibinfo{person}{Jingtao Ding}, \bibinfo{person}{Yunke Zhang}, \bibinfo{person}{Sasu Tarkoma}, \bibinfo{person}{Yong Li}, {and} \bibinfo{person}{Pan Hui}.} \bibinfo{year}{2023}\natexlab{}.
\newblock \showarticletitle{A satellite imagery dataset for long-term sustainable development in united states cities}.
\newblock \bibinfo{journal}{\emph{Scientific data}} \bibinfo{volume}{10}, \bibinfo{number}{1} (\bibinfo{year}{2023}), \bibinfo{pages}{866}.
\newblock


\bibitem[Xia et~al\mbox{.}(2018)]%
        {xia2018dota}
\bibfield{author}{\bibinfo{person}{Gui-Song Xia}, \bibinfo{person}{Xiang Bai}, \bibinfo{person}{Jian Ding}, \bibinfo{person}{Zhen Zhu}, \bibinfo{person}{Serge Belongie}, \bibinfo{person}{Jiebo Luo}, \bibinfo{person}{Mihai Datcu}, \bibinfo{person}{Marcello Pelillo}, {and} \bibinfo{person}{Liangpei Zhang}.} \bibinfo{year}{2018}\natexlab{}.
\newblock \showarticletitle{DOTA: A large-scale dataset for object detection in aerial images}. In \bibinfo{booktitle}{\emph{Proceedings of the IEEE conference on computer vision and pattern recognition}}. \bibinfo{pages}{3974--3983}.
\newblock


\bibitem[Xie et~al\mbox{.}(2021b)]%
        {xie2021atmospheric}
\bibfield{author}{\bibinfo{person}{Liuzhen Xie}, \bibinfo{person}{Qixiang Xu}, {and} \bibinfo{person}{Ruidong He}.} \bibinfo{year}{2021}\natexlab{b}.
\newblock \showarticletitle{Atmospheric pollution impact assessment of brick and tile industry: a case study of Xinmi City in Zhengzhou, China}.
\newblock \bibinfo{journal}{\emph{Sustainability}} \bibinfo{volume}{13}, \bibinfo{number}{4} (\bibinfo{year}{2021}), \bibinfo{pages}{2414}.
\newblock


\bibitem[Xie et~al\mbox{.}(2021a)]%
        {xie2021oriented}
\bibfield{author}{\bibinfo{person}{Xingxing Xie}, \bibinfo{person}{Gong Cheng}, \bibinfo{person}{Jiabao Wang}, \bibinfo{person}{Xiwen Yao}, {and} \bibinfo{person}{Junwei Han}.} \bibinfo{year}{2021}\natexlab{a}.
\newblock \showarticletitle{Oriented R-CNN for object detection}. In \bibinfo{booktitle}{\emph{Proceedings of the IEEE/CVF international conference on computer vision}}. \bibinfo{pages}{3520--3529}.
\newblock


\bibitem[Yang and Yan(2022)]%
        {yang2022arbitrary}
\bibfield{author}{\bibinfo{person}{Xue Yang} {and} \bibinfo{person}{Junchi Yan}.} \bibinfo{year}{2022}\natexlab{}.
\newblock \showarticletitle{On the arbitrary-oriented object detection: Classification based approaches revisited}.
\newblock \bibinfo{journal}{\emph{International Journal of Computer Vision}} \bibinfo{volume}{130}, \bibinfo{number}{5} (\bibinfo{year}{2022}), \bibinfo{pages}{1340--1365}.
\newblock


\bibitem[Yeh et~al\mbox{.}(2020)]%
        {yeh2020using}
\bibfield{author}{\bibinfo{person}{Christopher Yeh}, \bibinfo{person}{Anthony Perez}, \bibinfo{person}{Anne Driscoll}, \bibinfo{person}{George Azzari}, \bibinfo{person}{Zhongyi Tang}, \bibinfo{person}{David Lobell}, \bibinfo{person}{Stefano Ermon}, {and} \bibinfo{person}{Marshall Burke}.} \bibinfo{year}{2020}\natexlab{}.
\newblock \showarticletitle{Using publicly available satellite imagery and deep learning to understand economic well-being in Africa}.
\newblock \bibinfo{journal}{\emph{Nature communications}} \bibinfo{volume}{11}, \bibinfo{number}{1} (\bibinfo{year}{2020}), \bibinfo{pages}{2583}.
\newblock


\end{thebibliography}



\end{document}